\documentclass[10pt,a4paper]{report}
\usepackage[utf8]{inputenc}
\usepackage{amsmath}
\usepackage{amsfonts}
\usepackage{amssymb}
\usepackage{array}			
\usepackage{bbm}
\usepackage{mathtools}
\usepackage{ragged2e}		

\usepackage[titletoc]{appendix}

\usepackage{verbatim}		
\usepackage{enumerate}		

\usepackage{titlesec}		
\usepackage[bottom=2.5cm, left=2.5cm, right=2.5cm]{geometry}

\usepackage{IEEEtrantools}	
\usepackage{amsthm}			

\usepackage[sorting=none,backend=bibtex]{biblatex}

\usepackage[usenames,dvipsnames]{xcolor}
\usepackage{tikz}
\usepackage{pgfplots}
\usepackage{float}
\usepackage{wrapfig}

\author{Matej Balog, University of Oxford \\ supervised by Prof Yee Whye Teh, Department of Statistics, University of Oxford }
\title{The Mondrian Process for Machine Learning}

\titleformat{\chapter}{\normalfont\huge\bfseries}{}{0pt}{\Huge}
\titlespacing*{\chapter}{0pt}{-50pt}{20pt}	

\usepackage{chngcntr}
\counterwithin{equation}{chapter}

\usepackage{multirow}

\usepackage{algorithm}
\usepackage[noend]{algpseudocode}
\renewcommand{\algorithmiccomment}[1]{\bgroup\hfill\(\triangleright\)~#1\egroup}

\algrenewcommand\algorithmicrequire{\textbf{Input:}}
\algrenewcommand\algorithmicensure{\textbf{Output:}}
\algrenewcommand{\algorithmiccomment}[1]{\hskip3em$\rightarrow$ #1}

\usepackage{enumitem}
\setlist{nolistsep}

\theoremstyle{plain}
\newtheorem{theorem}{Theorem}[chapter]
\newtheorem{corollary}[theorem]{Corollary}
\newtheorem{lemma}[theorem]{Lemma}
\newtheorem{proposition}[theorem]{Proposition}
\theoremstyle{remark}
\newtheorem*{remark}{Remark}
\theoremstyle{definition}
\newtheorem{definition}[theorem]{Definition}

\newtheorem{example}[theorem]{Example}

\usepackage{graphicx}
\usepackage{mdwlist}

\newcommand{\hs}{\hspace{5mm}}
\newcommand{\eps}{\varepsilon}

\DeclareMathOperator*{\argmin}{argmin}
\DeclareMathOperator*{\argmax}{argmax}

\renewcommand{\d}{\,\mathrm{d}}

\newcommand{\bs}{\boldsymbol}

\newcommand{\II}{\mathbb{I}}
\newcommand{\IN}{\mathbb{N}}
\newcommand{\IR}{\mathbb{R}}
\newcommand{\IQ}{\mathbb{Q}}

\newcommand{\mB}{\mathcal{B}}

\newcommand{\mD}{\mathcal{D}}

\newcommand{\mL}{\mathcal{L}}
\newcommand{\mN}{\mathcal{N}}
\newcommand{\mM}{\mathcal{M}}
\newcommand{\mO}{\mathcal{O}}

\newcommand{\mS}{\mathcal{S}}

\newcommand{\mU}{\mathcal{U}}

\renewcommand{\sc}{\textsc}

\newcommand{\IP}{\mathbb{P}}
\newcommand{\IE}{\mathbb{E}}

\newcommand{\Ber}{\operatorname{Ber}}

\newcommand{\Poisson}{\operatorname{Poisson}}
\newcommand{\Exp}{\operatorname{Exp}}
\newcommand{\Beta}{\operatorname{Beta}}

\bibliography{MondrianProcess}

\begin{document}

\begin{titlepage}
\begin{center}

\includegraphics[width=0.15\textwidth]{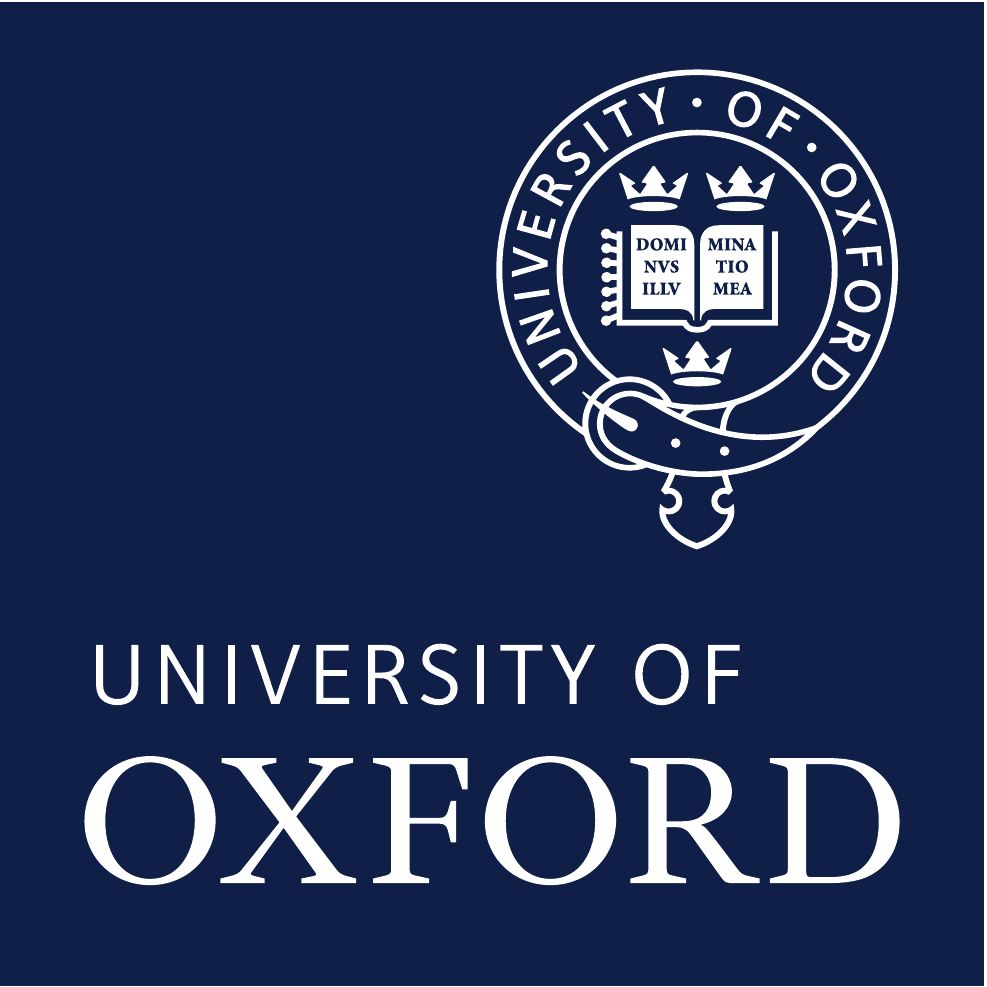}~\\[1cm]

\textsc{}\\[1.5cm]
\textsc{\Large Part C Computer Science Project Report}\\[0.5cm]
\rule{\linewidth}{0.5mm} \\[0.4cm]
{ \huge \bfseries The Mondrian Process in Machine Learning \\[0.4cm] }
\rule{\linewidth}{0.5mm} \\[1.5cm]

\noindent
\begin{minipage}[t]{0.4\textwidth}
\begin{flushleft} \large
\emph{Author:}\\
Matej \textsc{Balog} \\
Merton College \\
University of Oxford
\end{flushleft}
\end{minipage}%
\begin{minipage}[t]{0.4\textwidth}
\begin{flushright} \large
\emph{Supervisor:} \\
Professor Yee Whye \textsc{Teh} \\
Department of Statistics \\
University of Oxford
\end{flushright}
\end{minipage}

\vfill
{\large \today}
\end{center}
\end{titlepage}

\begin{abstract}
\vspace{1.5em}
\begin{center}
\begin{minipage}{0.7\textwidth}
This report is concerned with the Mondrian process \cite{roy2009mondrian} and its applications in machine learning. The Mondrian process is a guillotine-partition-valued stochastic process that possesses an elegant self-consistency property. The first part of the report uses simple concepts from applied probability to define the Mondrian process and explore its properties.

The Mondrian process has been used as the main building block of a clever online random forest classification algorithm that turns out to be equivalent to its batch counterpart.~\cite{LakshminarayananRoyTeh2014} We outline a slight adaptation of this algorithm to regression, as the remainder of the report uses regression as a case study of how Mondrian processes can be utilized in machine learning. In particular, the Mondrian process will be used to construct a fast approximation to the computationally expensive kernel ridge regression problem with a Laplace kernel.

The complexity of random guillotine partitions generated by a Mondrian process and hence the complexity of the resulting regression models is controlled by a lifetime hyperparameter. It turns out that these models can be efficiently trained and evaluated for all lifetimes in a given range at once, without needing to retrain them from scratch for each lifetime value. This leads to an efficient procedure for determining the right model complexity for a dataset at hand.

The limitation of having a single lifetime hyperparameter will motivate the final Mondrian grid model, in which each input dimension is endowed with its own lifetime parameter. In this model we preserve the property that its hyperparameters can be tweaked without needing to retrain the modified model from scratch.
\end{minipage}
\end{center}
\end{abstract}

\tableofcontents

\setcounter{chapter}{-1}
\chapter{Preliminaries}

\section{Notation}

Capital letters are used for counts: $N$ stands for a number of data points and $D$ for the dimensionality of the input space. When we consider Mondrian forests, $M$ will denote the number of Mondrian trees in the forest. In later chapters we will compute a randomized feature space and we will use $C$ to denote the number of its dimensions (number of features). Whenever possible, we will use matching lowercase letters as indices over corresponding ranges, i.e., we will use $n$ to index datapoints, $d$ to index input dimensions, $m$ to index Mondrian trees and $c$ to index random feature space dimensions.

Matrices and vectors are typeset in boldface (e.g., $\mathbf{A}$, $\bs{\theta}$) with the only exception of feature vectors. Throughout this report $\mathbf{X} \in \IR^{N \times D}$ stands for a data matrix (also called the \emph{design matrix}) whose $n$-th row $\mathsf{x}_n^T \in \IR^D$ is the feature vector of the $n$-th data point (in input space). The $(n, d)$-entry $x_{nd}$ of this matrix is the value of feature $d$ for datapoint $n$. Once we use a function $z$ to map input data points into a randomized feature space, we will have a feature matrix $\mathbf{Z} \in \IR^{N \times C}$ whose $n$-th row $\mathsf{z}_n^T \in \IR^C$ is the feature vector of the $n$-th data point in the new $C$-dimensional feature space.

By $\mathbf{e}_i$ we will denote the $i$-th standard basis vector, i.e. a binary vector with a single $1$ entry in position $i$. The dimensionality of this vector will be clear from context. The identity matrix of dimension $k \times k$ will be written $\mathbf{I}_k$.

The indicator function $\II(P)$, takes value $1$ when the predicate $P$ is true and the value $0$ otherwise.

\section{Mathematical preliminaries}

In this section we recall basic mathematical concepts from applied probability that we would like to use throughout the report without repeated elaboration.

\subsection*{Probability distributions}

\begin{definition} The \textbf{exponential distribution} with rate $\lambda > 0$, written $\Exp(\lambda)$, is the continuous probability distribution on $\IR$ with probability density function $p(x | \lambda) = \lambda e^{-\lambda x} \II(x \geq 0)$.
\end{definition}

Recall that the rate $\lambda$ of an exponential random variable is inversely proportional to its mean $\frac{1}{\lambda}$ (see Proposition~\ref{ExpDistributionExpectation}), meaning that variables with large rate will tend to take smaller values, and vice versa.

\begin{definition} For $D \geq 1$, the $D$-dimensional (non-degenerate) \textbf{Gaussian} (or \textbf{Normal}) distribution with mean $\bs{\mu} \in \IR^D$ and (positive definite) covariance $\bs{\Sigma} \in \IR^{D \times D}$ has density
\begin{equation*}
\mN(\mathbf{x} | \boldsymbol{\mu}, \boldsymbol{\Sigma} )
= \frac{1}{(2 \pi)^{\frac{D}{2}} | \boldsymbol{\Sigma} |^{\frac{1}{2}}}
  \exp\left( - \frac{1}{2} ( \mathbf{x} - \boldsymbol{\mu} )^T \boldsymbol{\Sigma}^{-1} ( \mathbf{x} - \boldsymbol{\mu} ) \right)
\end{equation*}
where $| \bs{\Sigma} |$ is the determinant of $\bs{\Sigma}$. In the case $D = 1$ we have $\bs{\Sigma} = | \bs{\Sigma} | = \sigma^2 \geq 0$ and we call this the variance; its inverse $p := \sigma^{-2}$ is then called the precision.
\end{definition}

\subsection*{Lack of memory and competing exponential clocks}

The exponential distribution plays a major role in the construction of the Mondrian process. This is because of its lack of memory property and the related concept of competing exponential clocks, which lead to elegant properties of the Mondrian process.

\begin{lemma}[Simple lack of memory property]
\label{SimpleLackOfMemoryProperty}
Let $Z \sim \Exp(\lambda)$. Then
\begin{equation*}
\forall{t \geq 0}
\hs
\left( (Z - t) \mid (Z > t) \right) \sim \Exp(\lambda)
\end{equation*}
In words, the residual lifetime $Z - t$ given survival $\{ Z > t \}$ is again $\Exp(\lambda)$ distributed.
\begin{proof} The more general Lemma~\ref{LackOfMemoryProperty} below is proved as Lemma~\ref{LackOfMemoryPropertyProof} in the appendix.
\end{proof}
\end{lemma}

It is interesting to note that the exponential distribution is the unique probability distribution supported on the positive reals with this property (see Proposition~\ref{ExponentialUniqueMemoryless} in the appendix). It turns out that the lack of memory property of the exponential distribution also holds at a random time, provided that it is independent of the exponential random variable considered.

\begin{lemma}[Lack of memory property]
\label{LackOfMemoryProperty}
Let $Z$ be an exponential random variable and $T$ an independent nonnegative random variable. Then $Z$ has the \emph{lack of memory property} at the random time $T$, i.e.
\begin{equation*}
\forall{u \geq 0}
\hs
\IP(Z - T > u | Z > T) = \IP(Z > u)
\end{equation*}
\begin{proof} Proof appears as Lemma~\ref{LackOfMemoryPropertyProof} in the appendix.
\end{proof}
\end{lemma}

\begin{definition} A set of $N$ \textbf{competing exponential clocks} is a collection of $N$ \emph{independent} exponential random variables $Z_1, \hdots, Z_N$ with respective rates $\lambda_1, \hdots, \lambda_N > 0$.
\end{definition}

We can think of these $N$ random variables as clocks, all started at the same time $0$, and each having an independent and exponentially distributed residual time until ringing. Natural questions to asks are: when will the first of the $N$ clocks ring and which clock is it going to be? Once the first clock rings, what is the joint distribution of residual times of the remaining $N - 1$ clocks? The lack of memory property of the exponential distribution leads to simple answers, entailed in the following theorem.

\begin{theorem}[Competing exponential clocks]
\label{CompetingExponentialClocks}
Say $Z_1, \hdots, Z_N$ are $N$ competing exponential clocks with respective rates $\lambda_1, \hdots, \lambda_N$. Then
\begin{itemize}
\item[•] the time until any of the clocks rings has $\Exp( \sum_n \lambda_n )$ distribution,
\item[•] the probability that the $n$-th clock is the first one to ring is $\frac{\lambda_n}{\sum_k \lambda_k}$,
\item[•] conditionally given the time and identity of the first clock to ring, the remaining $N - 1$ clocks remain independent and each has preserved its original $\Exp(\lambda_n)$ residual time distribution.
\end{itemize}
\begin{proof} Partial proofs are given in Appendix A.
\end{proof}
\end{theorem}

\subsection*{Statistical parameter estimation}

Say we have a probabilistic model parametrized by $\bs{\theta}$. The \textbf{likelihood} $\mL(\bs{\theta} | \mD)$ is the probability $p(\mD | \bs{\theta})$ of observed data $\mD$ under this model as a function of the parameters $\bs{\theta}$. The maximum likelihood estimate (\textbf{MLE}) of the parameters is
\begin{equation*}
\bs{\theta}^{\text{MLE}} := \argmax_{\bs{\theta}} p(\mD | \bs{\theta})
\end{equation*}
Suppose that before observing any data, we also have a prior belief about the value of the parameters $\bs{\theta}$, encoded as a \textbf{prior} probability distribution $p(\bs{\theta})$. Then the maximum a posteriori (\textbf{MAP}) estimate of $\bs{\theta}$ is the set of parameters that maximizes the \textbf{posterior} distribution $p(\bs{\theta} | \mD)$:
\begin{equation*}
\bs{\theta}^{\text{MAP}}
:= \argmax_{\bs{\theta}} p( \bs{\theta} | \mD )
= \argmax_{\bs{\theta}} \frac{p(\mD | \bs{\theta}) p( \bs{\theta} )}{p(\mD)}
= \argmax_{\bs{\theta}} p(\mD | \bs{\theta}) p( \bs{\theta} )
\end{equation*}
We say that a prior is \textbf{conjugate} for a likelihood, if the resulting posterior is a probability distribution from the same parametric family as the prior.

\begin{example}
\label{GaussianPredictionModel}
Suppose we want to model data $y \in \IR$ with a Gaussian likelihood $p(y | \mu) = \mN(y | \mu, \sigma_{\text{noise}}^2)$ where $\sigma_{\text{noise}}^2$ is fixed and the mean $\mu$ is unknown. Say we place a prior distribution $p(\mu) = \mN(\mu | \mu_{\text{prior}}, \sigma_{\text{prior}}^2)$ on $\mu$ to express our belief that $\mu$ is not far from $\mu_{\text{prior}}$. Then the prior is conjugate to the likelihood and the posterior distribution is again Gaussian. More concretely, if we gather $N$ independent observations $\mD = \{ y_1, \hdots, y_n \}$ then the posterior distribution of $\mu$ is
\begin{equation*}
p(\mu | \mD)
= \mN\left( \mu \mid \frac{p_{\text{prior}} \mu_{\text{prior}} + p_{\text{noise}} \sum_{n = 1}^N y_n}{p_{\text{prior}} + N p_{\text{noise}}}, ( p_{\text{prior}} + N p_{\text{noise}} )^{-1} \right)
\end{equation*}
where $p_{\text{prior}} = \sigma_{\text{prior}}^{-2}$ and $p_{\text{noise}} = \sigma_{\text{noise}}^{-2}$ are the prior and noise precisions (inverse variances), respectively.
\begin{proof} Appears as Proposition~\ref{GaussianPredictionModelProof} in the appendix.
\end{proof}
\end{example}

\chapter{Mondrian process}

In this chapter we define the Mondrian process \cite{roy2009mondrian} as a temporal stochastic process taking values in guillotine partitions of an axis-aligned box and then attempt to give intuitive explanations for some of its elegant properties. Let us start by agreeing on terminology.

\begin{definition} A \textbf{temporal stochastic process} taking values in a space $\mS$ is a collection $(M_t)_{t \geq 0}$ of $\mS$-valued random variables, indexed by a parameter $t \in [0, \infty)$ that we think of as time.
\end{definition}

\begin{definition} An \textbf{(axis-aligned) box} $\Theta$ in $\IR^D$ is a set of the form $\Theta = \Theta_1 \times \cdots \times \Theta_D \subseteq \IR^D$, where each $\Theta_d$ is a bounded interval $[a_d, b_d]$. We only work with axis-aligned boxes in this report, so we drop the "axis-aligned" qualification henceforth.
\end{definition}

\begin{definition} The \textbf{linear dimension} of a box $\Theta = [a_1, b_1] \times \cdots \times [a_D, b_D]$ in $\IR^D$, written $\text{LD}(\Theta)$, is the sum of its $D$ dimensions, i.e., $\text{LD}(\Theta) := \sum_{d = 1}^D (b_d - a_d)$.
\end{definition}

\begin{definition} Given a box $\Theta$ in $\IR^D$, a \textbf{guillotine partition} of $\Theta$ is a hierarchical partition of $\Theta$ obtained by recursively splitting boxes of the partition by some hyperplane orthogonal to one of the $D$ coordinate axes.
\end{definition}

Guillotine partitions can be thought of as \emph{$k$-d trees}, where each node corresponds to a box in $\IR^D$ and each non-leaf node $n$ has exactly two children, corresponding to the two boxes obtained after cutting the box associated with $n$ by a hyperplane that is orthogonal to one of the $D$ coordinate axes.

\subsection*{Mondrian process definition}

In this subsection we define the Mondrian process over a box $\Theta = [a_1, b_1] \times \cdots \times [a_D, b_D]$ as a temporal stochastic process taking values in guillotine partitions of $\Theta$. An intuitive way of thinking about the process is that it starts out at time $t = 0$ with the trivial partition of $\Theta$ (containing no cuts) and as time progresses, new cuts start to randomly appear, hierarchically splitting $\Theta$ into more refined partitions. The precise distribution that governs how the cuts appear is given by this recursive generative process:

\begin{algorithm}
\begin{algorithmic}[1]
\Procedure{Mondrian}{$\Theta$}
  \State \Return \Call{Mondrian-Started-At}{$\Theta$, $0$}
\EndProcedure
\end{algorithmic}
\begin{algorithmic}[1]
\Procedure{Mondrian-Started-At}{$\Theta, t_0$}
  \Comment{$\Theta = [a_1, b_1] \times \cdots \times [a_D, b_D]$}
  \State $T \sim \text{Exp}( \text{LD}(\Theta) )$
  \Comment{time until first cut appears}
  \State $d \sim \text{Discrete}(p_1, \hdots, p_D)$ where $p_d \propto (b_d - a_d)$
  \Comment{dimension of that cut}
  \State $x \sim \mU([a_d, b_d])$
  \Comment{location of that cut}
  \State $M^{<} \gets $ \Call{Mondrian-Started-At}{$\Theta^{<}, t_0 + T$} where $\Theta^{<} = \{ \mathbf{z} \in \Theta \mid z_d \leq x \}$
  \State $M^{>} \gets $ \Call{Mondrian-Started-At}{$\Theta^{>}, t_0 + T$} where $\Theta^{>} = \{ \mathbf{z} \in \Theta \mid z_d \geq x \}$
  \State \Return $(t_0, t_0 + T, d, x, M^{<}, M^{>})$
\EndProcedure
\end{algorithmic}
\end{algorithm}

The recursive procedure \sc{Mondrian-Started-At}($\Theta, t_0$) generates a Mondrian process on the box $\Theta$, started at time $t_0$. Let us analyze this procedure line by line:
\begin{itemize}
\item[•] Line $2$ generates the time it takes for the first cut in $\Theta$ to appear. The distribution is exponential with rate the linear dimension of $\Theta$. Note that by Proposition~\ref{ExpDistributionExpectation}, in larger boxes a cut is expected sooner than in smaller ones. The absolute time $t_0 + T$ of the generated cut is called its \textbf{birth time}.
\item[•] Lines $3$ and $4$ generate the dimension $d$ and location $x$ of the first cut, respectively. The former is generated proportionally to the dimensions of $\Theta$ and the latter is then chosen uniformly. The cutting hyperplane is orthogonal to the $d$-th coordinate axis and crosses it at the point $x$. Thus the cutting hyperplane "lives" in the $d$-th dimension.
\item[•] Lines $5$ and $6$ recursively generate \emph{independent} Mondrians $M^{<}$, $M^{>}$ on the two boxes $\Theta^{<}$, $\Theta^{>}$ obtained by cutting $\Theta$ at $x$ in dimension $d$. The start time of these Mondrians equals the birth time of the cut that gave rise to $\Theta^{<}$ and $\Theta^{>}$. Note that the spaces $\Theta^{<}$, $\Theta^{>}$ are indeed still boxes in $\IR^D$, so the recursive calls are valid.
\item[•] Line $7$ returns a node of the k-d tree representing the hierarchical partition. The node is a $6$-tuple of the form $(t_b, t_c, d, x, M^{<}, M^{>})$, where the entries represent respectively the birth time, the cut time, the cut dimension, the cut location and the two children of the \emph{node}. Note that the cut time of a node equals the birth time of the cut that splits it.
\end{itemize}

\begin{remark} Several remarks about this generative process are in order:
\begin{itemize}
\item[•] Lines 3-4 can be informally summarized as sampling the cut uniformly from the linear dimension.
\item[•] The distributions $\text{Exp}( \text{LD}(\Theta) )$ and $\text{Discrete}(p_1, \hdots, p_D)$ (where $p_d \propto (b_d - a_d)$) are well-defined provided that the linear dimension of $\Theta$ is positive. If we start with a box $\Theta$ of positive dimension, then with probability $1$ the cut location $x$ is sampled in an interior point of $[a_d, b_d]$ and then both $\Theta^{<}$ and $\Theta^{>}$ also have positive linear dimension.
\item[•] We are somewhat sloppy about the generated partition as the cutting hyperplane is included in both $\Theta^{<}$ and $\Theta^{>}$. This informality can be excused since any specific point of interest has probability $0$ of being hit by a cut (the cut location $x$ is generated from a continuous distribution). In particular, with probability $1$ no cut will appear in the same location where a cut has already been made.
\end{itemize}
\end{remark}

This generative process translates into the definition of a temporal stochastic process as follows.

\begin{definition} Let $\Theta$ be a box in $\IR^D$ with positive linear dimension. The \textbf{Mondrian process} on $\Theta$, denoted as $\text{MP}(\Theta)$, is a temporal stochastic process $(M_t)_{t \geq 0}$ taking values in guillotine partitions of $\Theta$ and its distribution is specified by the generative process \sc{Mondrian}($\Theta$): the random variable $M_t$ is the guillotine partition of $\Theta$ formed by cuts/nodes with birth time $t_b \leq t$.
\end{definition}

In other words, $M_t$ is the partition generated by \sc{Mondrian}($\Theta$) with all cuts/nodes born after time $t$ ignored. In fact, we can generate the random variable $M_t$ precisely by running the recursive process \sc{Mondrian}($\Theta$) and terminating any recursive call that generates a cut with time $t_0 + T > t$. (This is the way the Mondrian process has first been introduced in \cite{roy2009mondrian}.)

\begin{definition} Let $\Theta$ be a box in $\IR^D$ with positive linear dimension and let $t \geq 0$. The \textbf{Mondrian process} on $\Theta$ \textbf{with lifetime} $t$, denoted as $\text{MP}(t, \Theta)$, is the law of $M_t$ where $M \sim \text{MP}(\Theta)$.
\end{definition}

For fixed $t \geq 0$, $\text{MP}(t, \Theta)$ is simply a probability distribution over guillotine partitions of $\Theta$. Note that existing cuts are never removed from a Mondrian process $M$, so it exhibits the following kind of monotonicity property:
\begin{equation*}
0 \leq t_1 \leq t_2
\hs \Rightarrow \hs
\text{the partition } M_{t_2} \text{ is a refinement of the partition } M_{t_1}
\end{equation*}
Therefore in the family of probability distributions $(\text{MP}(t, \Theta))_{t \geq 0}$ over guillotine partitions of $\Theta$, the lifetime parameter $t$ can be thought of as controlling the complexity of the resulting partition. The generative process of the Mondrian chooses cut locations uniformly at random, so it is in the way the \emph{times} of the cuts are generated where the ingenuity of the Mondrian process construction lies. The resulting elegant mathematical properties, which we explore in subsequent sections, follow from the memoryless property of the exponential distribution and the related concept of competing exponential clocks (Theorem~\ref{CompetingExponentialClocks}).

\begin{figure}[H]
\centering
\begin{minipage}{0.65\textwidth}
\begin{example} Say we sample from a Mondrian process $M$ on a 2D box $\Theta = [0, 1] \times [0, 1]$ with a lifetime cut-off at $t = 1.5$. The figure on the right shows the obtained cuts, together with their birth times.

The process starts at time $t = 0$ with $M_0$ the trivial partition of $\Theta$. The first cut appears after time $T \sim \Exp(\lambda)$ where $\lambda = 1 + 1 = 2$ is the linear dimension of $[0, 1] \times [0, 1]$. At time $T$ the first cut appears at a location chosen uniformly from the linear dimension of $\Theta$. More precisely, first the dimension $d$ of the cut is chosen with probabilities proportional to the lengths of $\Theta$ in each dimension. In our case $\Theta$ has length $1$ in both dimensions, so the cutting dimension $d$ is chosen with equal probability $\frac{1}{2}$ from $\{ 1, 2 \}$. After the dimension $d$ is generated, a point $x$ in $[0, 1]$ is chosen uniformly at random. The cut is then determined by the hyperplane (in our case, a line) lying entirely in dimension $d$ and containing the point $x$ on the $d$-th coordinate axis.
\end{example}
\end{minipage}
\hfill
\begin{minipage}{0.34\textwidth}
\centering
\includegraphics[scale=0.52]{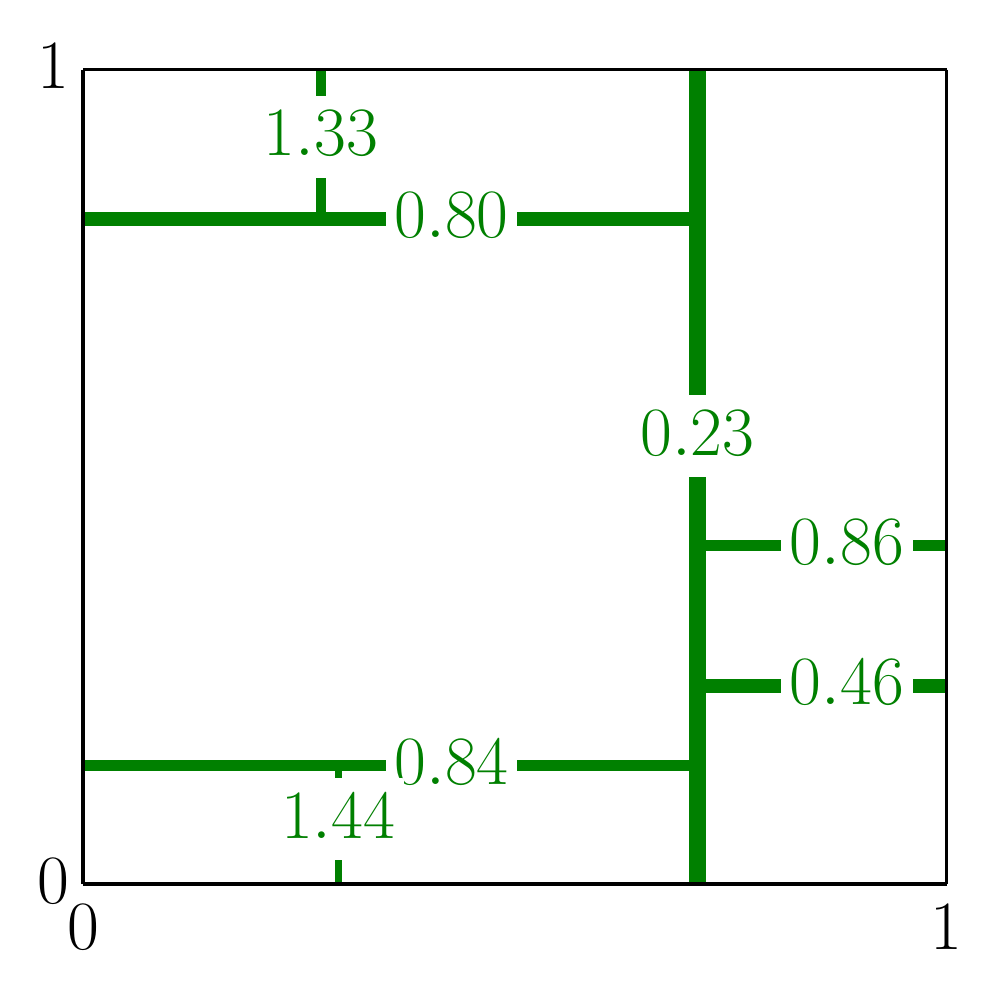}
\label{2DMondrianFigure}
\end{minipage}
\justifying\vspace{0.0001em}

\noindent
For the sample shown in the figure we generated $T \approx 0.23$, $d = 1$ and $x \approx 0.71$. For all $t \in [0, T)$, $M_t$ remains the trivial partition of $\Theta$. The cut made at time $T$ partitions the box $\Theta$ into two sub-boxes $\Theta^{<} = [0, x] \times [0, 1]$ and $\Theta^{>} = [x, 1] \times [0, 1]$. In each of these two sub-boxes the Mondrian process continues to run independently and afresh, started at time $T$.
\end{figure}

\section{Mondrian process in 1D}

As a first illustration of how the choice of exponential distribution yields elegant properties of the Mondrian process, we consider the one-dimensional case, where it turns out that the cut locations follow a Poisson point process. The following definition of a Poisson point process is adapted from \cite{StirzakerPaRP}.

\begin{definition}
\label{PoissonPointProcessDefinition}
Let $\lambda \geq 0$. A random countable subset $\Pi$ of $\IR$ is a \textbf{Poisson point process with (constant) intensity} $\lambda$, if, for all $A \in \mB(\IR)$, the random variables $N(A) := | \Pi \cap A |$ satisfy:
\begin{itemize}
\item[(i)] $N(A) \sim \Poisson( \lambda m(A) )$, where $m(A) \in [0, \infty]$ is the Lebesgue measure of $A$, and
\item[(ii)] if $A_1, \hdots, A_n$ are disjoint sets in $\mB(\IR)$ then $N(A_1), \hdots, N(A_n)$ are independent random variables.
\end{itemize}
Here $\mB(\IR)$ is the Borel $\sigma$-algebra on $\IR$. Also, we allow a Poisson distribution with infinite rate, in which case $N(A) = \infty$ almost surely.
\end{definition}

Suppose we run a Mondrian process $M$ on a one-dimensional axis-aligned box $\Theta$ with positive linear dimension, which is simply an interval $\Theta = [a, b]$ with $a < b$. Up to a finite lifetime $\lambda$, the process generates a hierarchical partition of $[a, b]$, with each cut having a birth time $t_b \in [0, \lambda]$. Let us now only consider the marginal distribution of the cut \emph{locations} (marginalizing out their hierarchy and times). This is a distribution over subsets of $[a, b]$ and in the following lemma we give a simple representation for it.

\begin{lemma}
\label{1DMondrianRepresentation}
Let $a < b$ and $\lambda \geq 0$. The distribution of the cut locations $\{ X_n \}$ of a Mondrian process $M$ run on $[a, b]$ with a finite lifetime $\lambda$ can be represented by the following two-stage generative process:
\begin{IEEEeqnarray*}{rCll+x*}
N & \; \sim \; &
  \Poisson(\lambda (b - a))
\\
X_1, \hdots, X_N \mid N & \stackrel{\text{i.i.d.}}{\sim} &
  \mU([a, b])
\end{IEEEeqnarray*}
In words, the number of cuts is Poisson distributed with rate $\lambda (b - a)$ and the location of each cut is independent and uniformly distributed in the interval.

\begin{proof} Fix a time instant $t \in [0, \lambda]$ and suppose we are conditionally given the evolution of the process $M$ up to time $t$. Let $K - 1$ be the number of generated cuts, so that the interval $[a, b]$ is partitioned into $K$ segments of the form $[x_0, x_1], [x_1, x_2], \hdots, [x_{K - 1}, x_K]$ with $a = x_0 < x_1 < \cdots < x_K = b$. The time until the next cut appears in a segment $[x_k, x_{k - 1}]$ has by memorylessness (Lemma~\ref{LackOfMemoryProperty}) $\Exp(x_k - x_{k - 1})$ distribution and is independent of all the other segments by construction of the Mondrian process.

Thus we are in the setting of $K$ competing exponential clocks and the residual time until the next cut in $[a, b]$ appears has exponential distribution with rate $\sum_{i = 1}^K (x_k - x_{k - 1}) = x_K - x_0 = b - a$. Also, the probability of this cut occurring in a particular segment $[x_k, x_{k - 1}]$ is proportional to its length $(x_k - x_{k - 1})$. Within the chosen segment the location of the cut is generated uniformly, so marginally the location of the next cut in $[a, b]$ is uniformly in $[a, b]$.

Thus we've shown that at any time instant $t$, given the past evolution of the process, the residual time until the next cut appears is $\Exp(b - a)$ distributed and its location is chosen uniformly from $[a, b]$. As these distributions do not depend on the past evolution, the residual time until the next cut and its location are both independent of this past evolution. The times of the cuts form a temporal Poisson process with rate $b - a$, so their number in a time interval of length $\lambda$ is $\Poisson(\lambda (b - a))$ distributed.
\end{proof}
\end{lemma}

\begin{figure}[H]
\begin{center}
\begin{tikzpicture}[scale=1.5]
  \draw[->] (-0.1,0) -- (4.5,0) node[right] {time};
  \draw[-]  (0,-0.1) -- (0,2.2);
  \draw[-]  (-0.1,2) -- (4.5,2);

  \draw[-] (0.3,1.2) -- (4.0,1.2);
  
  \draw[shift={(-0.1,0)}] node[left] {$x_0 = a$};
  \draw[shift={(-0.1,2)}] node[left] {$x_5 = b$};
  
  \foreach \i/\time/\x in 
  {3/0.3/1.2,
   1/0.5/0.4,
   2/1.0/0.9,
   4/0.6/1.7}
  {
    \fill[shift={(\time,\x)}] circle[radius=1.5pt];
    \draw[-] (\time,\x) -- (4.0,\x);
    \draw[-] (0.05,\x) -- (-0.05,\x) node[left] {$x_\i$};
  }
  
  \foreach \i/\j/\x in 
  {0/1/0.20,
   1/2/0.65,
   2/3/1.05,
   3/4/1.45,
   4/5/1.85}
  {
    \draw[shift={(1.4,\x)}] node[right] {$\sim \Exp(x_\j - x_\i)$};
  }  
  
  \draw[-] (1.4,0.05) -- (1.4,-0.05) node[below] {$t$};
  \draw[-] (4.0,0.05) -- (4.0,-0.05) node[below] {$\lambda$};
  
  \draw[dashed, thin, draw=gray] (1.4,0) -- (1.4,2);
  \draw[dashed, thin, draw=gray] (4.0,0) -- (4.0,2);
  
\end{tikzpicture}
\end{center}
\caption{An illustration of the proof of Lemma~\ref{1DMondrianRepresentation}. Here $K = 5$.}
\end{figure}
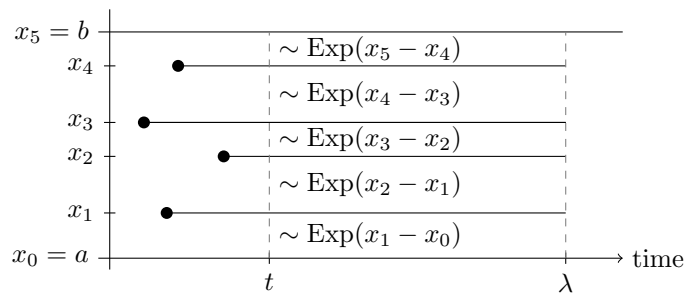

\begin{theorem}
\label{Mondrian1DIsPPP}
Let $a < b$ and $\lambda \geq 0$. The distribution of the cut locations of a Mondrian process $M$ run on $[a, b]$ with a finite lifetime $\lambda$ is a Poisson point process with constant intensity $\lambda$.
\begin{proof} It suffices to show that a Poisson point process with constant intensity $\lambda$ can be generated using the two-stage generative process of Lemma~\ref{1DMondrianRepresentation}. The number of points $N$ generated by a Poisson point process with constant intensity $\lambda$ on $[a, b]$ has $\text{Poisson}(\lambda (b - a))$ distribution by definition, matching the first stage of the generative process. Conditionally given that the Poisson point process generated $N = n$ points, their locations are independent and uniformly distributed, matching the second stage of the generative process. (A proof of the last statement is given as Lemma~\ref{PoissonPointProcessConditional} in the appendix.) As the cut locations of a 1D Mondrian process and of a Poisson point process can be sampled using the same procedure, their distributions must coincide.
\end{proof}
\end{theorem}

\section{Self-consistency of the Mondrian process}

\begin{wrapfigure}{r}{0.33\textwidth}
\vspace{-2.1em}
\begin{center}
\begin{tikzpicture}[scale=4.5]
  \draw[-] (0,0) -- node[below] {$\Theta_1$} (1,0);
  \draw[-] (1,0) -- (1,1);
  \draw[-] (1,1) -- (0,1);
  \draw[-] (0,1) -- node[left] {$\Theta_2$} (0,0);

  \draw[line width=1.5pt, -] (0.4,0.3) -- node[below] {$\Phi_1$} (0.9,0.3);
  \draw[line width=1.5pt, -] (0.9,0.3) -- (0.9,0.9);
  \draw[line width=1.5pt, -] (0.9,0.9) -- (0.4,0.9);
  \draw[line width=1.5pt, -] (0.4,0.9) -- node[left] {$\Phi_2$} (0.4,0.3);

  \draw[dotted, thin, draw=gray]  (0.74,0)   -- (0.74,1);
  \draw[line width=1.5pt, draw=ForestGreen] (0.74,0.3) -- (0.74,0.9);

  \draw[dotted, thin, draw=gray]  (0,0.82)   -- (0.74,0.82);
  \draw[line width=1.5pt, draw=ForestGreen] (0.4,0.82) -- (0.74,0.82);

  \draw[dotted, thin, draw=gray]  (0.74,0.42) -- (1,0.42);
  \draw[line width=1.5pt, draw=ForestGreen] (0.74,0.42) -- (0.9,0.42);

  \draw[dotted, thin, draw=gray]  (0.28,0.82) -- (0.28,1);
  \draw[dotted, thin, draw=gray]  (0,0.15)    -- (0.74,0.15);
  \draw[dotted, thin, draw=gray]  (0.2,0)     -- (0.2,0.15);
  \draw[dotted, thin, draw=gray]  (0.74,0.25) -- (1,0.25);
\end{tikzpicture}
\end{center}
\vspace{-2.5em}
\end{wrapfigure}

This section is concerned with the following natural question: if we run a Mondrian process on a larger box but only look at what happens in a smaller subbox, what distribution of random partitions of the subbox do we obtain? More formally, consider the setup
\begin{IEEEeqnarray*}{rCll+x*}
M & \sim &
  \text{MP}(\Theta_1 \times \cdots \times \Theta_D)
\\
\Phi_1 \times \cdots \times \Phi_D
  & \subseteq &
  \Theta_1 \times \cdots \times \Theta_D
\end{IEEEeqnarray*}
i.e., we run a Mondrian process $M$ on an a box $\Theta := \Theta_1 \times \cdots \times \Theta_D$ and consider some smaller box $\Phi := \Phi_1 \times \cdots \times \Phi_D$ contained within it. Some cuts of $M$ will cross $\Phi$ and thus induce a guillotine-partition-valued stochastic process on $\Phi$. The Mondrian process was conceived precisely so that the distribution of this stochastic process is again a Mondrian process \cite{roy2011thesis}. Here we give an intuitive argument for where this self-consistency property comes from. The choice of the exponential distribution for the times of the cuts turns out to be crucial, as is the notion of competing exponential clocks.

\begin{theorem}[Self-consistency of Mondrian process] The law of the restriction of $M \sim \text{MP}(\Theta)$ to a smaller box $\Phi \subseteq \Theta$ is again a Mondrian process.
\end{theorem}

\begin{wrapfigure}{r}{0.33\textwidth}
\vspace{-2.2em}
\begin{center}
\begin{tikzpicture}[scale=4.5]
  \draw[draw=white] (0,0) -- node[below] {$\Theta_1$} (1,0);
  \draw[-] (1,0) -- (1,1);
  \draw[-] (1,1) -- (0,1);
  \draw[draw=white] (0,1) -- node[left] {$\Theta_2$} (0,0);

  \draw[-] (0.4,0.3) -- node[below] {$\Phi_1$} (0.9,0.3);
  \draw[-] (0.9,0.3) -- (0.9,0.9);
  \draw[-] (0.9,0.9) -- (0.4,0.9);
  \draw[-] (0.4,0.9) -- node[left] {$\Phi_2$} (0.4,0.3);

  \draw[line width=1.5pt, draw=red]         (0,0)   -- (0.4,0);  
  \draw[line width=1.5pt, draw=ForestGreen] (0.4,0) -- (0.9,0);
  \draw[line width=1.5pt, draw=red]         (0.9,0) -- (1,0);
  \draw[line width=1.5pt, draw=red]         (0,0)   -- (0,0.3);  
  \draw[line width=1.5pt, draw=ForestGreen] (0,0.3) -- (0,0.9);
  \draw[line width=1.5pt, draw=red]         (0,0.9) -- (0,1);  

  \draw[dotted, thin, draw=gray] (0.4,0) -- (0.4,0.3);
  \draw[dotted, thin, draw=gray] (0.9,0) -- (0.9,0.3);
  \draw[dotted, thin, draw=gray] (0,0.3) -- (0.4,0.3);
  \draw[dotted, thin, draw=gray] (0,0.9) -- (0.4,0.9);
\end{tikzpicture}
\end{center}
\vspace{-1em}
\caption{Representing the first cut distribution using two competing exponential clocks.}
\label{ConsistencyIntuitionFigure1}
\end{wrapfigure}

We provide intuition for the case $D = 2$, but the ideas generalize to any number of dimensions. To argue that the resulting distribution on $\Phi$ is a Mondrian process, we show that the Mondrian process $M$ running on $\Theta$ generates cuts in $\Phi$ in the same way as a Mondrian process running directly on $\Phi$ would.

The first cut in $M$ occurs at time $T \sim \Exp( \text{LD}(\Theta) )$ and its location is uniformly distributed along the linear dimension of $\Theta$. Employing the notion of competing exponential clocks "backwards", we can represent this distribution of time and location of the first cut using two competing clocks:
\begin{itemize}
\item[(1)] Clock $C_1$ with rate $\lambda_1 = \text{LD}(\Phi)$. If this clock wins, the location of the cut is sampled uniformly from the locations where making a cut splits $\Phi$ (green segments in Figure~\ref{ConsistencyIntuitionFigure1}).
\item[(2)] Clock $C_2$ with rate $\lambda_2 = \text{LD}(\Theta) - \text{LD}(\Phi)$. If this clock wins, the cut location is sampled uniformly form the locations where making a cut doesn't split $\Phi$ (red segments in Figure~\ref{ConsistencyIntuitionFigure1}).
\end{itemize}
Indeed, under this representation the time until the first cut is exponentially distributed with the correct rate $\lambda_1 + \lambda_2 = \text{LD}(\Theta)$ and the cut location is sampled uniformly from the linear dimension of $\Theta$ since the probability that clock $C_1$ wins is proportional to $\lambda_1 = \text{LD}(\Phi)$.

\pagebreak

Note that clock $C_1$ represents the same distribution of the first cut as a Mondrian process running directly on $\Phi$ would. Of course, it may happen that instead clock $C_2$ wins, and a cut is made outside of $\Phi$, as illustrated in Figure~\ref{ConsistencyIntuitionFigure2} below. But observe that when such a cut is made, the measure of the locations where a cut splitting $\Phi$ can be made (the green segments) remains to be $\text{LD}(\Phi)$. Therefore instead of considering two new competing exponential clocks $C_1', C_2'$ as above, for $C_1'$ we can reuse the clock $C_1$ that continues to run as an independent exponential clock of rate $\lambda_1 = \text{LD}(\Phi)$ by Theorem~\ref{CompetingExponentialClocks}.

\begin{figure}[H]
\centering
\begin{minipage}{0.45\textwidth}
\centering
\begin{tikzpicture}[scale=4.5]
  \draw[-] (0,0) -- node[below] {$\Theta_1$} (1,0);
  \draw[-] (1,0) -- (1,1);
  \draw[-] (1,1) -- node[above, text=white] {placeholder} (0,1);
  \draw[-] (0,1) -- node[left] {$\Theta_2$} (0,0);

  \draw[-] (0.4,0.3) -- node[below] {$\Phi_1$} (0.9,0.3);
  \draw[-] (0.9,0.3) -- (0.9,0.9);
  \draw[-] (0.9,0.9) -- (0.4,0.9);
  \draw[-] (0.4,0.9) -- node[left] {$\Phi_2$} (0.4,0.3);

  \draw[line width=1.5pt, draw=red]         (0.17,0)   -- (0.4,0);  
  \draw[line width=1.5pt, draw=ForestGreen] (0.4,0)    -- (0.9,0);
  \draw[line width=1.5pt, draw=red]         (0.9,0)    -- (1,0);
  \draw[line width=1.5pt, draw=red]         (0.17,0)   -- (0.17,0.3);  
  \draw[line width=1.5pt, draw=ForestGreen] (0.17,0.3) -- (0.17,0.9);
  \draw[line width=1.5pt, draw=red]         (0.17,0.9) -- (0.17,1);  

  \draw[dotted, thin, draw=gray] (0.4,0) -- (0.4,0.3);
  \draw[dotted, thin, draw=gray] (0.9,0) -- (0.9,0.3);
  \draw[dotted, thin, draw=gray] (0.17,0.3) -- (0.4,0.3);
  \draw[dotted, thin, draw=gray] (0.17,0.9) -- (0.4,0.9);
\end{tikzpicture}
\caption{Cut outside $\Phi$.}
\label{ConsistencyIntuitionFigure2}
\end{minipage}\hfill
\begin{minipage}{0.45\textwidth}
\centering
\begin{tikzpicture}[scale=4.5]
  \draw[-] (0,0) -- (1,0);
  \draw[-] (1,0) -- (1,1);
  \draw[-] (1,1) -- (0,1);
  \draw[-] (0,1) -- (0,0);

  \draw[-, color=blue] (0.4,0.3) -- node[below] {$\Phi^{<}_1$} (0.9,0.3);
  \draw[-, color=blue]   (0.9,0.3) -- (0.9,0.65);
  \draw[-, color=orange] (0.9,0.65) -- (0.9,0.9);
  \draw[-, color=orange] (0.9,0.9) -- node[below, text=orange] {$\Phi^{>}_1$} (0.4,0.9);
  \draw[-, color=orange] (0.4,0.9) -- node[left] {$\Phi^{>}_2$} (0.4,0.65);
  \draw[-, color=blue] (0.4,0.65) -- node[left] {$\Phi^{<}_2$} (0.4,0.3);

  \draw[-, color=blue] (0.17,0)   -- node[below] {$\Theta^{<}_1$} (1,0);
  \draw[-, color=blue] (1,0)      -- (1,0.65);
  \draw[-, color=blue] (1,0.65)   -- (0.17,0.65);
  \draw[-, color=blue] (0.17,0.65) -- node[left] {$\Theta^{<}_2$} (0.17,0);
  
  \draw[-] (0.4,0.65) -- (0.9,0.65);
  \draw[-, color=black] (0.17,0.65) -- (1,0.65);
  \draw[-, color=orange] (1,0.65) -- (1,1);
  \draw[-, color=orange] (1,1) -- node[above, text=orange] {$\Theta^{>}_1$} (0.17,1);
  \draw[-, color=orange] (0.17,1) -- node[left] {$\Theta^{>}_2$} (0.17,0.65);
  
  \draw[dotted, thin, draw=gray] (0.4,0) -- (0.4,0.3);
  \draw[dotted, thin, draw=gray] (0.9,0) -- (0.9,0.3);
  \draw[dotted, thin, draw=gray] (0.17,0.3) -- (0.4,0.3);
  \draw[dotted, thin, draw=gray] (0.17,0.9) -- (0.4,0.9);
\end{tikzpicture}
\caption{Cut inside $\Phi$.}
\label{ConsistencyIntuitionFigure3}
\end{minipage}
\end{figure}

Hence, cuts made outside of $\Phi$ do not affect the distribution of the first cut within $\Phi$, and this distribution is the same as if a Mondrian process was running on $\Phi$ directly. Now consider the situation when finally a cut is made within $\Phi$, as illustrated in Figure~\ref{ConsistencyIntuitionFigure3}. By definition of the Mondrian process, the processes on the two sides $\Theta^{<}$, $\Theta^{>}$ of this cut continue to run independently, and therefore their restrictions to $\Phi$ are also independent. Thus our argument proceeds by induction, confirming that the generative process for the cuts within $\Phi$ induced by the Mondrian process $M$ run on $\Theta$ is the same as of a Mondrian process running directly on $\Phi$. \qed

\begin{example}[Mondrian slices] An interesting special case of self-consistency is pointed out in \cite{roy2009mondrian}. Suppose that $\Phi = \Phi_1 \times \cdots \times \Phi_D$ lives in lower dimension than $\Theta$, for example $\Phi_1 = \{ x \}$ for some $x \in \Theta_1$. As the probability of making a cut precisely at the point $x$ is zero by continuity of the uniform distribution, a.s. all cuts of $\Theta$ splitting $\Phi$ live in the remaining dimensions $d \not= 1$. Therefore the restriction of $M \sim \text{MP}(\Theta)$ to $\Phi$ can be viewed as a $(d - 1)$-dimensional Mondrian process run on $\Phi_2 \times \cdots \times \Phi_D$.

This observation provides some insight into how partitions generated by a Mondrian process look like. Along any axis-parallel line, the locations of the cuts crossing it follow the distribution of a 1D Mondrian process, which has been shown to coincide with a Poisson point process.
\end{example}

An important corollary of self-consistency is that it provides the Mondrian process with the projectivity property required for extending its definition form bounded boxes to the entire $\IR^D$.

\begin{definition} The \textbf{Mondrian process on $\IR^D$} with lifetime $\lambda \geq 0$, written $\text{MP}(\lambda, \IR^D)$, is the temporal stochastic process taking values in (infinite) partitions of $\IR^D$ with the property that its restriction to any bounded box $\Theta$ has the law $\text{MP}(\lambda, \Theta)$, as defined earlier.
\end{definition}

\pagebreak

\section{Conditional Mondrians}

Conditional Mondrians are a dual notion to consistency. Similarly as before, we have the setup
\begin{IEEEeqnarray*}{rCll}
M & \sim &
  \text{MP}(\lambda, \Theta_1 \times \cdots \times \Theta_D)
  & \hs \lambda \in [0, \infty]
\\
\Phi
:= \Phi_1 \times \cdots \times \Phi_D & \subseteq &
  \Theta_1 \times \cdots \times \Theta_D
\end{IEEEeqnarray*}
but this time we are \emph{conditionally given} the restriction $M^{\Phi} = m^{\Phi}$ of $M$ to the smaller box $\Phi$. (Both the locations and times of cuts in $\Phi$ are given.) The question we want to approach is, what is the conditional distribution $M \mid (M^{\Phi} = m^{\Phi})$ and can we sample from it?

The answer is positive and provides a way of extending an existing sample $M^{\Phi} \sim \text{MP}(\lambda, \Phi)$ on $\Phi$ to a sample $M$ on the larger domain $\Theta$ in such a way that the extended sample has the correct marginal distribution $M \sim \text{MP}(\lambda, \Theta)$. This is because by self-consistency $\text{MP}(\lambda, \Phi)$ can be interpreted both as a Mondrian process running on $\Phi$ or as the restriction to $\Phi$ of a Mondrian process running on $\Theta$.

\begin{wrapfigure}{r}{0.33\textwidth}
\vspace{-2.2em}
\begin{center}
\begin{tikzpicture}[scale=4.5]
  \draw[draw=white] (0,0) -- node[below] {$\Theta_1$} (1,0);
  \draw[-] (1,0) -- (1,1);
  \draw[-] (1,1) -- (0,1);
  \draw[draw=white] (0,1) -- node[left] {$\Theta_2$} (0,0);

  \draw[-] (0.4,0.3) -- node[below] {$\Phi_1$} (0.9,0.3);
  \draw[-] (0.9,0.3) -- (0.9,0.9);
  \draw[-] (0.9,0.9) -- (0.4,0.9);
  \draw[-] (0.4,0.9) -- node[left] {$\Phi_2$} (0.4,0.3);

  \draw[line width=2pt, -] (0.4,0.75) -- node[below] {$C^{\Phi}$} (0.9,0.75);
  \draw[-] (0.8,0.3) -- (0.8,0.75);
  \draw[-] (0.5,0.75) -- (0.5,0.9);

  \draw[line width=1.5pt, draw=red]         (0,0)   -- (0.4,0);  
  \draw[line width=1.5pt, draw=ForestGreen] (0.4,0) -- (0.9,0);
  \draw[line width=1.5pt, draw=red]         (0.9,0) -- (1,0);
  \draw[line width=1.5pt, draw=red]         (0,0)   -- (0,0.3);  
  \draw[line width=1.5pt, draw=ForestGreen] (0,0.3) -- (0,0.9);
  \draw[line width=1.5pt, draw=red]         (0,0.9) -- (0,1);  

  \draw[dotted, thin, draw=gray] (0.4,0) -- (0.4,0.3);
  \draw[dotted, thin, draw=gray] (0.9,0) -- (0.9,0.3);
  \draw[dotted, thin, draw=gray] (0,0.3) -- (0.4,0.3);
  \draw[dotted, thin, draw=gray] (0,0.9) -- (0.4,0.9);
\end{tikzpicture}
\end{center}
\vspace{-1em}
\caption{Conditional Mondrians. $C^{\Phi}$ is the first cut in $\Phi$.}
\label{ConditionalMondriansIntuition}
\end{wrapfigure}
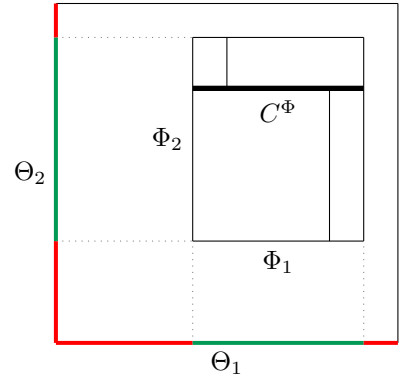

\begin{theorem} Suppose we are conditionally given the restriction $M^{\Phi} = m^{\Phi}$ of a Mondrian process $M \sim \text{MP}(\Theta)$ to a smaller box $\Phi \subseteq \Theta$. Let $C^{\Phi}$ be the first cut in $M^{\Phi}$ and let $t^{\Phi}$ be its time. Then
\begin{itemize}
\item[$\bullet$] with probability $\exp( t^{\Phi} (\text{LD}(\Theta) - \text{LD}(\Phi))$, the cut $C^{\Phi}$ is the first cut in $\Theta$ (it extends throughout $\Theta$)
\item[$\bullet$] with complementary probability $1 - \exp( t^{\Phi} (\text{LD}(\Theta) - \text{LD}(\Phi))$ the first cut in $\Theta$ misses $\Phi$, its time has the truncated exponential distribution with rate $\text{LD}(\Theta) - \text{LD}(\Phi)$ and truncation at $t^{\Phi}$, and the cut location is uniformly distributed along the segments where making a cut doesn't hit $\Phi$.
\end{itemize}
\end{theorem}

Again we only provide an intuition for this result. A calculation using the self-consistency property for the case where $m^{\Phi}$ is the trivial partition of $\Phi$ can be found as Lemma~\ref{ConditionalMondriansCase1Proof} in the appendix.

Observe that the stated probability of $C^{\Phi}$ being the first cut in $\Theta$ is the likelihood of an exponential clock with rate $\text{LD}(\Theta) - \text{LD}(\Phi)$ not to ring at least until time $t^{\Phi}$.

Once again we represent the \emph{unconditional} distribution of the first cut in $\Theta$ by two competing exponential clocks $C_1$, $C_2$ as in the section on self-consistency. Recall that clock $C_1$ has rate $\lambda_1 = \text{LD}(\Phi)$ and is associated with the green segments, where making a cut splits $\Phi$. Clock $C_2$ has rate $\lambda_2 = \text{LD}(\Theta) - \text{LD}(\Phi)$ and is associated with the red segments where making a cut misses $\Phi$. Our conditioning on $M^{\Phi} = m^{\Phi}$ tells us that clock $C_1$ rang at time $t^{\Phi}$, and the two cases in the statement of the theorem correspond respectively to the situation where it was the first and where it was the second clock to ring.

If $C_1$ was the second clock to ring, we know from our representation that the location of the first cut in $\Theta$ is uniformly distributed along the red segments associated with the winning clock $C_2$. Also, in that case the time of this cut has exponential distribution with the rate $\lambda_2$ of clock $C_2$, but truncated at $t^{\Phi}$ since we assume that $C_2$ rang before $t^{\Phi}$. \qed

Hence we obtain a simple algorithm for sampling from the conditional distribution $M \mid (M^{\Phi} = m^{\Phi})$: we sample $T_2 \sim \Exp(\lambda_2)$, the time when clock $C_2$ rings. If $T_2 \leq t^{\Phi}$ we extend the first cut $C^{\Phi}$ in $\Phi$ to the whole of $\Theta$; otherwise we sample the first cut of $\Theta$ uniformly from the locations where it won't hit $\Phi$. In both cases we thus obtain the first cut in $\Theta$. Then by definition of the Mondrian process we may proceed independently and recursively on the two boxes $\Theta^{<}$, $\Theta^{>}$ created by the first cut in $\Theta$. (Note that if this cut is not $C^{\Phi}$ then on one of its sides we are no longer conditioning on anything, i.e. an unconditional Mondrian will be sampled in that recursive call.)

\pagebreak

\section{Remarks}

In this chapter we have defined the Mondrian process and attempted to give intuitive explanations for how the choice of exponential distribution and the notion of competing exponential clocks translate into some of its elegant properties. A rigorous treatment of the Mondrian process requires additional concepts from measure theory and can be found in Dan Roy's PhD thesis \cite{roy2011thesis}. For example, one of the issues we have ignored in our exposition is the possibility of the process exploding, i.e. infinitely many cuts occurring in a bounded box in finite time. Roy \cite{roy2011thesis} confirms that this happens with probability $0$.

Also, the Mondrian process can be defined slightly more generally. The only property of the uniform distribution for sampling cut locations that we have used in our arguments is that it is continuous and therefore with probability $1$ no two cuts occur at the same location. Thus instead we may take $D$ atomless measures $\mu_1, \hdots, \mu_D$ on $\IR$, define the linear dimension of $\Theta = \Theta_1 \times \cdots \times \Theta_D$ as $\text{LD}(\Theta) = \sum_{d = 1}^D \mu_d( \Theta_d )$ and sample cut locations in dimension $d$ from the (possibly unnormalized) measure $\mu_d$. This preserves the self-consistency property and the notion of Conditional Mondrians, while the one dimensional Mondrian becomes a Poisson Point Process with (non-constant) intensity function $\lambda \mu_1$. Our intuitive arguments translate into this more general setting by replacing all interval lengths $y - x$ in dimension $d$ with $\mu_d([x, y])$.

\begin{figure}[H]
  \centering
  \includegraphics{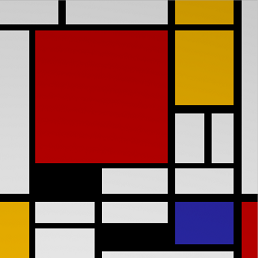}
  \caption{Piet Mondrian: Composition with Large Red Plane, Yellow, Black, Gray, and Blue (1921). The Mondrian process has been named after the French painter Piet Mondrian due to the resemblance of some of his work to partitions generated by a Mondrian process \cite{roy2011thesis}. However, note that unlike the depicted painting, partitions generated by a Mondrian process will not have two cuts crossing each other.}
\end{figure}

\chapter{Mondrian forests}

Apart from exhibiting elegant properties, the Mondrian process turns out to be useful in various machine learning tasks. In this chapter we give a high-level description of Mondrian forests, a concept introduced in \cite{LakshminarayananRoyTeh2014} for online random forest classification. However, in line with the focus of subsequent chapters, we concentrate on regression rather than classification here. The regression problem is defined as follows.

\begin{definition} \textbf{Regression} is the problem of learning a function $f : \IR^D \to \IR$ from a set of \textbf{training} data points $\mD = \{ (\mathsf{x}_1, y_1), \hdots, (\mathsf{x}_N, y_N) \} \subseteq \IR^D \times \IR$, where $y_n$ is a possibly noisy observation of $f(\mathsf{x}_n)$. Given a new \textbf{test} point $\mathsf{x}_{*} \in \IR^D$, the learned function $\hat{f}$ predicts $\hat{y} = \hat{f}(\mathsf{x}_{*})$ for the value of $f(\mathsf{x}_{*})$.
\end{definition}

In particular, we assume the input space to be $\IR^D$. The inputs $\mathsf{x}$ are $D$-dimensional vectors whose components are called \textbf{features} or \textbf{attributes}. Each feature can be thought of as a quantifiable property of the input, and one hopes that these features provide information useful for estimating the target value.

A powerful idea exploiting the assumption that nearby points tend to have similar target values is to partition the input space $\IR^D$ into connected blocks $\IR^D = \bigsqcup_{i \in I} B_i$ and to use a simple regression model in each block. For example, when asked for a prediction at a test point $\mathsf{x}_{*} \in \IR^D$, we might return the average target value across those training points $(\mathsf{x}_n, y_n)$ that fall into the same block $B_i$ as $\mathsf{x}_{*}$ does.

Instead of a single partition, a random forest model obtains $M$ partitions from $M$ independent decision trees that hierarchically partition the input space. At test time, each tree provides a prediction and their average is returned. Using several trees instead of a single one is a bias reduction technique, useful because the partition generated by a single tree is rarely complex enough to match the patterns in training data.

A Mondrian forest algorithm uses $M$ independent samples from a Mondrian process with finite lifetime $\lambda$ to provide the $M$ partitions of $\IR^D$. For Mondrian forest regression, in each cell of each partition we use a constant prediction model with a Gaussian prior $\mN(\mu, \sigma_{\text{prior}}^2)$ and Gaussian observation noise $\mN(0, \sigma_{\text{noise}}^2)$, as in Example~\ref{GaussianPredictionModel}. Apart from acting as a regularizer, the prior ensures that predictions are well-defined in partition cells with no training data.

\begin{algorithm}
  \begin{algorithmic}[1]
  	\Require{training set $\mD = \{ (\mathsf{x}_1, y_1), \hdots, (\mathsf{x}_n, y_n) \}$, lifetime parameter $\lambda \geq 0$, test point $\mathsf{x}_{*}$}
    \Ensure{estimate $\hat{y}$ of $f(\mathsf{x}_{*})$ that utilizes information from the training data $\mD$}
  \end{algorithmic}
  \begin{algorithmic}[1]
  	\Procedure{Train}{$\mD, \lambda$}
    \For{$m = 1$ \textbf{to} $M$}
      	\State $T_m \sim \text{MP}(\lambda, \IR^D)$
      	\State For each partition cell (leaf) $c$ in $T_m$, compute the posterior $\mN( \mu_c, \sigma_c^2 )$
      	\Comment{Example~\ref{GaussianPredictionModel}}
    \EndFor
    \State \Return $(T, \bs{\mu})$
	\EndProcedure
  \end{algorithmic}
  \begin{algorithmic}[1]
	\Procedure{Predict}{$\mathsf{x}_{*}$}
    \State \Return $\frac{1}{M} \sum_{m = 1}^M \mu_{l_m(\mathsf{x}_{*})}$
    \Comment{$l_m(\mathsf{x}_{*})$ is the leaf into which $\mathsf{x}_{*}$ falls in tree $T_m$}
    \EndProcedure
  \end{algorithmic}
\end{algorithm}
The algorithm prescribes sampling Mondrian processes on Euclidean space $\IR^D$, which is strictly speaking impossible as they contain infinitely many cuts with probability $1$. However, we can invoke self-consistency and only sample the Mondrians on a bounded box containing all the datapoints. This is sufficient because the Mondrian samples are only used to partition the datapoints. When new training points arrive in an online learning setting, the notion of Conditional Mondrians allows us to extend the $M$ existing samples to larger regions if necessary.

In the test phase we also need to incorporate test points $\mathsf{x}_{*}$ into the partition. We could again employ Conditional Mondrians if the Mondrian samples have not yet been instantiated at the point $\mathsf{x}_{*}$. However, \cite{LakshminarayananRoyTeh2014} points out that it is easy to consider all possible extensions of the partitions analytically and compute a prediction by integrating over them. Whenever the test point $\mathsf{x}_{*}$ lies outside of the region where a Mondrian sample is instantiated, the notion of Conditional Mondrians tells us exactly the probability with which $\mathsf{x}_{*}$ is separated from the other datapoints by a new cut, in which case the prediction made at $\mathsf{x}_{*}$ is simply the predictive prior.

For a more detailed description of Mondrian random forests we refer the interested reader to \cite{LakshminarayananRoyTeh2014}, where the aforementioned procedures are transparently presented.

\subsection*{Predictive behaviour far from training data}

When a test data point $\mathsf{x}_{*}$ lying far from any training points arrives, the probability that a cut separates it from the training data is high and in that case the predictive distribution is simply the prior. So for a test point $\mathsf{x}_{*}$ far from training data, thanks to integrating over all possible extensions of the $M$ Mondrian samples to incorporate $\mathsf{x}_{*}$, the predictive distribution is (close to) the prior. Hence we do not observe over confident predictions far from training data, as we do in some other random forest models \cite{Criminisi2011TR}.

\subsection*{Classification}

The Mondrian random forest model for classification presented in \cite{LakshminarayananRoyTeh2014} uses the same model for partitioning the input space as outlined above for regression. It only differs in the predictive model used in the leaves, which needs to predict classes rather than a continuous target value. A hierarchical Bayesian modeling approach is taken to achieve a smoothing effect: the hierarchical partitions provided by the Mondrian samples are treated as trees and each node of the tree (not just the leaves) is associated with a predictive distribution. Under the prior, the predictive distribution of a non-root node $n$ is modeled as a normalized stable process (NSP) with base distribution being the predictive distribution of $n$'s parent.

\subsection*{Density estimation}

Density estimation differs from regression and classification in that it is an unsupervised problem, i.e., no labels are observed in training data.

\begin{definition} \textbf{Density estimation} is the problem of learning a probability density $p$ from a set of \textbf{training} samples $\mD = \{ \mathsf{x}_1, \hdots, \mathsf{x}_N \} \subseteq \IR^D$ generated from $p$. Given a new \textbf{test} point $\mathsf{x}_{*}$, the learned density $\hat{p}$ estimates $\hat{p}(\mathsf{x}_{*})$ for the true density at point $\mathsf{x}_{*}$.
\end{definition}

A Mondrian random forest model for density estimation needs to be able to predict density values in its leaves, noting that a probability density must integrate to $1$. To this end, we associate cells of the partitions generated by the Mondrians with probability masses, ensuring that the total mass in one partition is $1$. However, to arrive at the density, the probability mass associated with a box needs to be divided by the volume of that box. This requires us to be able to compute volumes of the partition cells generated by the Mondrians, unlike in regression or classification where it was only the partition induced on the data points that was relevant. As density estimation is not a main theme of this report, a more detailed description of Mondrian random forest density estimation is given in the appendix.

\section{Empirical evaluation}

Note that the partitioning of the input space by $M$ Mondrian samples does not take into account labels (target values) of the training data points (in the case of regression and classification, where these labels are present). It is therefore quite remarkable that an algorithm like this can still achieve as competitive predictive performance as shown in \cite{LakshminarayananRoyTeh2014}.

We have implemented a Mondrian random forest algorithm both for regression and density estimation. More illustrations of empirical results obtained are shown in the chapter on regularization paths, where the models are endowed with an additional functionality that allows them to be evaluated much more efficiently.

\begin{figure}[H]
\centering
\begin{minipage}[t]{0.47\textwidth}
\centering
\begin{figure}[H]
  \centering
  \includegraphics[scale=0.61]{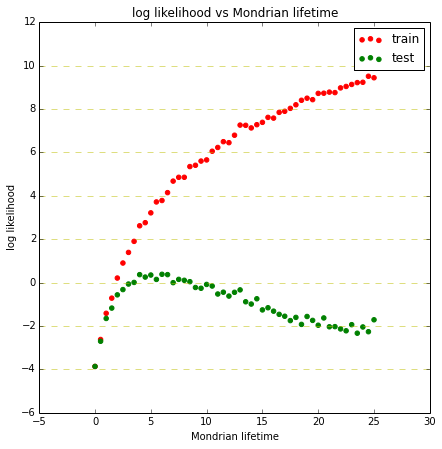}
  \caption{We train a Mondrian random forest density estimator on the Setosa class of the well-known Iris dataset from UCI repository \cite{Lichman:2013}, using several values of the lifetime parameter $\lambda$. Recall that the lifetime controls the complexity of the partitions generated by the Mondrian process. The horizontal axis of the figure shows this lifetime parameter $\lambda$ of the Mondrian process from which $M = 200$ samples were drawn. The vertical axis show the log-likelihood of the learned model on both training (red) and test (green) data. The plot shows how the lifetime parameter affects the complexity of the model: the training likelihood increases as the model gets more complex and is able to fit the training data better, while the test log-likelihood peaks and then starts to decrease as the model overfits to the noise present in training data.
Note that for each value of the lifetime shown, we have trained a new Mondrian random forest density estimation model from scratch. In the chapter on computing entire regularization paths we will present a much more efficient approach, where a single model will be evaluated for all possible lifetime values in a given range.}
\end{figure}
\end{minipage}\hfill
\begin{minipage}[t]{0.47\textwidth}
\centering
\begin{figure}[H]
  \centering
  \includegraphics[scale=0.61]{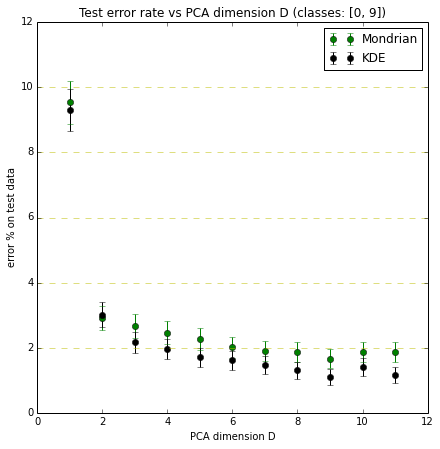}
  \caption{A density estimation procedure can be used indirectly for classification by estimating the probability density of all classes at training time and when a new test point $\mathsf{x}_{*}$ is presented, the density of all classes at $\mathsf{x}_{*}$ can be estimated. Our prediction may then be the class with highest density at $\mathsf{x}_{*}$.
  In the figure we compare two such classifiers: one that uses Mondrian random forest density estimation and another that uses standard kernel density estimation (KDE) with the squared exponential kernel. The dataset consists of images of digits 0 and 9, taken from the popular MNIST dataset \cite{mnisthandwrittendigit}. The horizontal axis shows the number of dimensions into which the inputs were projected using linear PCA. The vertical axis shows the error rate of the classifiers in distinguishing the digits $0$ and $9$. It seems that the Mondrian approach is more competitive in smaller number of dimensions. We have attempted to set the hyperparameters of both models to sensible values and kept them constant throughout these experiments.}
\end{figure}
\end{minipage}
\end{figure}

\chapter{Laplace Kernel Approximation}

This chapter presents a different way of utilizing Mondrian processes in regression problems, by way of approximating the Laplace kernel in a kernel ridge regression setting. We start by reviewing ridge regression as an instance of MAP parameter estimation and show how it can be kernelized.

\section{Ridge regression}

Say we want to model a dataset $\mD = \{ (\mathsf{x}_1, y_1), \hdots, (\mathsf{x}_N, y_N) \} \subseteq \IR^D \times \IR$ using a linear model of the form
\begin{equation*}
y_n = \mathsf{x}_n^T \bs{\theta} + \eps_n
\hs \text{where} \hs
\eps_1, \hdots, \eps_N \stackrel{\text{i.i.d.}}{\sim} \mN(0, \sigma_{\text{noise}}^2)
\end{equation*}
with $\sigma_{\text{noise}} > 0$ fixed and $\bs{\theta}$ to be learned. Suppose we express our belief that parameters are unlikely to take on arbitrarily large values by placing a spherical Gaussian prior $\bs{\theta} \sim \mN(\mathbf{0}, \sigma_{\text{prior}}^2 \bs{I}_D)$ on $\bs{\theta}$. It can be shown (see Theorem~\ref{RidgeRegressionIsMAP} in the appendix for a proof) that the MAP estimate of $\bs{\theta}$ under this prior and likelihood can be found by minimizing the $L_2$-regularized least squares objective function
\begin{equation*}
f(\bs{\theta})
:= \delta^2 \| \bs{\theta} \|_2^2 + \sum_{n = 1}^N (y_n - \mathsf{x}_n^T \bs{\theta})^2
\end{equation*}
where $\delta := \frac{\sigma_{\text{noise}}}{\sigma_{\text{prior}}} > 0$. The $\delta^2$ factor weights the strength of the regularizer: regularization is strong when the ratio of noise and prior variances is large, allowing us to attribute any outliers to noise in the observations. Conversely, regularization is weak when the noise variance is small (in comparison to the prior variance), forcing the model to better match the training observations.

The function $f(\bs{\theta})$ is easily minimized using matrix calculus. To this end, we construct the design matrix $\mathbf{X} \in \IR^{N \times D}$ whose $n$-th row is the data vector $\mathsf{x}_n^T$, and let $\mathbf{y} \in \IR^N$ be a vector with $n$-th entry set to $y_n$. The function $f(\bs{\theta})$ can then be vectorized as
\begin{equation*}
f(\bs{\theta})
= \delta^2 \| \bs{\theta} \|_2^2 + \| \mathbf{y} - \mathbf{X} \bs{\theta} \|_2^2
\end{equation*}
As Theorem~\ref{RidgeRegressionSolutionDD} in the appendix shows, for $\delta > 0$ this function has a unique global minimum at
\begin{equation*}
\bs{\theta}^{\text{MAP}}
= (\mathbf{X}^T \mathbf{X} + \delta^2 \mathbf{I}_D)^{-1} \mathbf{X}^T \mathbf{y}
\end{equation*}

\subsection*{Kernelizing ridge regression}

This formula requires inverting the $D \times D$ matrix $\mathbf{X}^T \mathbf{X} + \delta^2 \mathbf{I}_D$, which involves the \emph{feature} covariance matrix $\mathbf{X}^T \mathbf{X}$. As Theorem~\ref{RidgeRegressionSolutionNN} in the appendix shows, $\bs{\theta}^{\text{MAP}}$ can be alternatively expressed in terms of the $N \times N$ matrix $\mathbf{X} \mathbf{X}^T + \delta^2 \mathbf{I}_N$, where instead the \emph{data} covariance matrix $\mathbf{X} \mathbf{X}^T$ appears:
\begin{equation*}
\bs{\theta}^{\text{MAP}}
= \mathbf{X}^T (\mathbf{X} \mathbf{X}^T + \delta^2 \mathbf{I}_N)^{-1} \mathbf{y}
\end{equation*}
Occasionally we may have $D > N$, in which case this second form leads to more efficient inversion of a smaller matrix. However, we consider it because it allows the prediction $\hat{y}$ at a new test point $\mathsf{x}_{*}$ to be expressed in terms of inner products between data points. More concretely, this prediction is
\begin{equation*}
\hat{y}
= \mathsf{x}_{*}^T \bs{\theta}^{\text{MAP}}
= ( \mathsf{x}_{*}^T \mathbf{X}^T ) ( \mathbf{X} \mathbf{X}^T + \delta^2 \mathbf{I}_N)^{-1} \mathbf{y}
\end{equation*}
Here $\mathsf{x}_{*}^T \mathbf{X}^T \in \IR^N$ is a row vector with $n$-th entry the inner product $\mathsf{x}_{*}^T \mathsf{x}_n$ and $\mathbf{X} \mathbf{X}^T \in \IR^{N \times N}$ is the data covariance matrix with $(i, j)$-entry the inner product $\mathsf{x}_i^T \mathsf{x}_j$. The famous "kernel trick" is to observe that in a model like this where data locations only enter through inner products $\mathsf{x}^T \mathsf{x}'$, we can replace these inner products by a general kernel function $k(\mathsf{x}, \mathsf{x}')$. This kernel function must correspond to inner products in some feature space, but this feature space can be arbitrarily complex, even infinite dimensional. The trick is that we are able to compute inner products in that feature space efficiently via the kernel function $k(\mathsf{x}, \mathsf{x}')$, without the need to map input data to that feature space explicitly. Note that this moves us to the world of non-linear regression, since a linear function in the implied feature space usually does not correspond to a linear function in input space.

A function $k$ is a valid kernel if and only if the resulting Gram matrix is positive semidefinite for any collection of datapoints $\{ \mathsf{x}_1, \hdots, \mathsf{x}_n \}$. This result is known as Mercer's Theorem.

When we replace inner products $\mathsf{x}^T \mathsf{x}'$ by the kernel $k(\mathsf{x}, \mathsf{x}')$, the data covariance matrix $\mathbf{X} \mathbf{X}^T$ is replaced by the \textbf{Gram matrix} $\mathbf{K}$ with $(i, j)$-entry $k(\mathsf{x}_i, \mathsf{x}_j)$ corresponding to the inner product of the $i$-th and $j$-th training data point in the feature space implicitly represented by the kernel function $k$. Similarly, the row vector $\mathsf{x}_{*}^T \mathbf{X}^T \in \IR^N$ is replaced by $\mathbf{k}(\mathsf{x}_{*}, \mathbf{X}) \in \IR^n$, which is a vector with $n$-th entry equal to $k(\mathsf{x}_{*}, \mathsf{x}_n)$. Then the prediction $\hat{y}$ is given by
\begin{equation*}
\hat{y}
= \mathbf{k}(\mathsf{x}_{*}, \mathbf{X}) ( \mathbf{K} + \delta^2 \mathbf{I}_N)^{-1} \mathbf{y}
\end{equation*}

\section{Kernel approximation}

To compute this prediction $\hat{y}$ we need access to the inverse of the $N \times N$ matrix $\mathbf{K} + \delta^2 \mathbf{I}_N$. Even though this inversion only needs to be performed once, the $\Theta(N^3)$ computational cost of inverting such a matrix is prohibitive with large datasets. It is not possible to revert to the earlier formulation of the ridge regression solution involving a $D \times D$ matrix inversion, because that formulation is not in terms of inner products between datapoints. Instead, it has been proposed in \cite{rahimi2007random} to compute a randomized low-dimensional feature map $z : \IR^D \to \IR^C$ (with $C \ll N$) such that
\begin{equation*}
k(\mathsf{x}, \mathsf{x}')
\approx z(\mathsf{x})^T z(\mathsf{x}')
\end{equation*}
i.e., such that inner products in the generated low-dimensional feature space approximate the desired kernel $k$. Using $z$ to map each data point $\mathsf{x}$ to its low-dimensional feature representation $z(\mathsf{x})$, we can then use the first ridge regression solution formulation $\bs{\theta}^{\text{MAP}} = (\mathbf{Z}^T \mathbf{Z} + \delta^2 \mathbf{I}_C)^{-1} \mathbf{Z}^T \mathbf{y}$, where $\mathbf{Z} \in \IR^{N \times C}$ is the feature matrix with $n$-th row equal to $z(\mathsf{x})^T$. This only requires inverting a $C \times C$ matrix, allowing this (approximate) solution to be computed in time $\mO(C^3 + C^2 N)$. As $N$ is expected to dominate $C$, this is essentially $\mO(C^2 N)$.

Many kernels turn out to be effectively approximable this way, with \cite{rahimi2007random} giving two general schemes for constructing the random feature mapping $z$. In this section we show how the Mondrian process can be used to approximate one particular kernel, the (symmetric) Laplace kernel.

\begin{definition} The \textbf{(symmetric) Laplace kernel} is given by
\begin{equation*}
k(\mathsf{x}, \mathsf{x}')
= \exp\left( - \lambda \| \mathsf{x} - \mathsf{x}' \|_1 \right)
= \exp\left( - \lambda \sum_{d = 1}^D |x_d - x'_d| \right)
\end{equation*}
where $\lambda$ is a \textbf{lifetime} parameter of the kernel.
\end{definition}

\begin{remark} The Laplace kernel is usually equivalently defined with a \textbf{length-scale} parameter $\sigma$ that is related to our lifetime parameter $\lambda$ via $\lambda = 1 / 2 \sigma^2$, or $\lambda = 1 / 2 \sigma$ or $\lambda = 1 / \sigma$. Our parametrization and naming of the $\lambda$ parameter as lifetime is non-standard, chosen here because of the connection to the Mondrian process lifetime that will be revealed next.

The term "symmetric" is used here to point out that this kernel has a single lifetime parameter $\lambda$ common to all $D$ dimensions. The last chapter on the Mondrian grid is concerned with approximating the general Laplace kernel, where each dimension can have a different lifetime parameter $\lambda_d$.
\end{remark}

\pagebreak

\subsection*{Symmetric Laplace kernel approximation}

To approximate the Laplace kernel for a collection of datapoints $\{ \mathsf{x}_1, \hdots, \mathsf{x}_N \}$, suppose we sample a Mondrian process on $\IR^D$ with a finite lifetime $\lambda$. As we will only be interested in the partitioning of data points induced by the sample, by self-consistency we can simply sample the Mondrian on minimal bounded boxes containing the data points, as in Mondrian random forest regression. Let $C$ be the number of non-empty partition cells (containing at least one data point) of the sampled Mondrian and label them as $l_1, \hdots, l_C$. Let $l(\mathsf{x})$ be the function that returns the cell into which point $\mathsf{x} \in \IR^D$ falls. We define our random feature mapping $\mathsf{x} \mapsto z(\mathsf{x}) \in \IR^C$ as
\begin{equation*}
z(\mathsf{x})
:= \left( \II( l(\mathsf{x}) = l_1 ), \hdots, \II( l(\mathsf{x}) = l_C ) \right)^T
\end{equation*}
This is simply the indicator vector of the partition cell into which $\mathsf{x}$ falls. In particular, it contains a single non-zero entry. The inner product between two datapoints in the feature space defined by $z$ is
\begin{equation*}
z(\mathsf{x})^T z(\mathsf{x}')
= \II( l(\mathsf{x}) = l(\mathsf{x}') )
= \begin{dcases}
  1 & \text{ if } \mathsf{x}, \mathsf{x}' \text{ are in the same partition cell } \\
  0 & \text{ otherwise }
\end{dcases}
\end{equation*}
Observe that two datapoints $\mathsf{x}$, $\mathsf{x}'$ fall into the same cell if and only if the Mondrian sample has no cut in the minimal axis-aligned box $B( \{ \mathsf{x}, \mathsf{x}' \})$ containing $\mathsf{x}$ and $\mathsf{x}'$. By self-consistency, the probability of this happening is the same as that of running a Mondrian process $\mM$ with the same lifetime $\lambda$ on the box $B( \{ \mathsf{x}, \mathsf{x}' \})$ and not observing any cuts. Thus
\begin{figure}[H]
\vspace{-1.3em}
\begin{minipage}{0.75\textwidth}
\begin{IEEEeqnarray*}{rCl}
\IP(z(\mathsf{x})^T z(\mathsf{x}') = 1) & = &
  \IP(\mM' \text{ contains no cuts})
\\ & = &
  \IP(\text{first cut time in $\mM'$ is $> \lambda$} )
\\ & = &
  \IP\left( \text{Exp}\left( \text{LD}(B( \{ \mathsf{x}, \mathsf{x}' \})) \right) > \lambda \right)
\\ & = &
  \IP\left( \text{Exp}\left( \| \mathsf{x} - \mathsf{x}' \|_1 \right) > \lambda \right)
\\ & = &
  \exp \left( - \lambda \| \mathsf{x} - \mathsf{x}' \|_1 \right)
\end{IEEEeqnarray*}
\end{minipage}\hfill
\begin{minipage}{0.24\textwidth}
\centering
\begin{tikzpicture}[scale=4.0]
  \draw[-] (0.0,0.0) -- node[below] {$|x_1 - x'_1|$} (0.5,0.0);
  \draw[-] (0.5,0.0) -- (0.5,0.6);
  \draw[-] (0.5,0.6) -- (0.0,0.6);
  \draw[-] (0.0,0.6) -- node[left] {$|x_2 - x'_2|$} (0.0,0.0);

  \fill[shift={(0.0,0.0)}] circle[radius=0.5pt] node[below left] {$\mathsf{x}$};
  \fill[shift={(0.5,0.6)}] circle[radius=0.5pt] node[below left] {$\mathsf{x}'$};
  
  \draw[shift={(0.25,0.3)}] node {$B( \{ \mathsf{x}, \mathsf{x}' \})$};  
\end{tikzpicture}
\end{minipage}
\vspace{-0.3em}
\end{figure}
So inner products in our random feature space are Bernoulli random variables and their expectations are precisely the kernel values we want to approximate. To decrease the variance of the approximations, instead of using a single Mondrian, we may sample $M$ independent Mondrians and obtain the randomized feature mapping $z(\mathsf{x})$ by concatenating feature vectors $z_1(\mathsf{x}), \hdots, z_M(\mathsf{x})$ from the $M$ Mondrian samples. We normalize this vector by $M^{-1/2}$, so that
\begin{equation*}
\left( \frac{z(\mathsf{x})}{\sqrt{M}} \right) ^T \left( \frac{z(\mathsf{x}')}{\sqrt{M}} \right)
= \frac{1}{M} z(\mathsf{x})^T z(\mathsf{x}')
= \frac{1}{M} \sum_{m = 1}^M z_m(\mathsf{x})^T z_m(\mathsf{x}')
\;\longrightarrow\; \IE[ z_1(\mathsf{x})^T z_1(\mathsf{x}') ]
= e^{- \lambda \| \mathsf{x} - \mathsf{x}' \|_1}
\end{equation*}
As a Monte Carlo estimate, the convergence to the Laplace kernel as $M \to \infty$ is at the standard rate, i.e., the standard deviation of the estimator decreases as $\mO(M^{-1/2})$.

\pagebreak

\subsection*{Empirical evaluation}

We check experimentally that as the number $M$ of Mondrian samples increases, the performance of the resulting regression model approaches the performance of a model using the exact Laplace kernel. The training data size $N_{\text{train}} = 3000$ is chosen so that we see a computational benefit from using our approximation but an exact $\mO(N_{\text{train}}^3)$ computation is still possible given enough resources. Recall that the asymptotic time complexity of one approximate computation is $\mO(C^2 N)$, where $C$ is the number of dimensions of the random feature space produced by $z$.

We have also implemented another approximation scheme called Random Binning \cite{rahimi2007random} to compare against our Mondrian approximation. Random Binning has a hyperparameter playing a similar role as $M$ that affects the number of random features produced.

\begin{figure}[H]
\centering
\begin{minipage}[t]{0.47\textwidth}
\centering
\begin{figure}[H]
  \centering
  \includegraphics[scale=0.4]{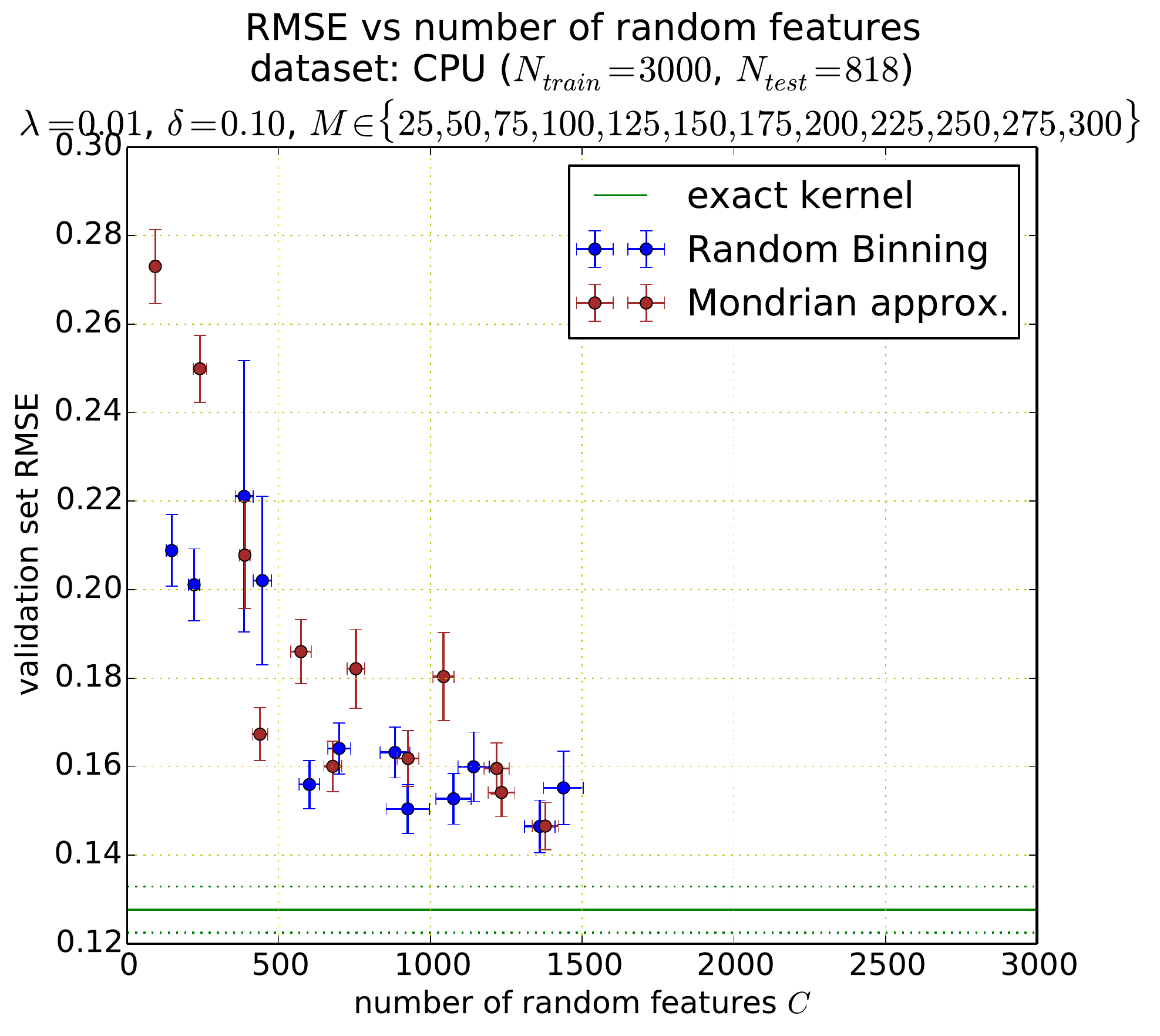}
\end{figure}
\end{minipage}\hfill
\begin{minipage}[t]{0.47\textwidth}
\centering
\begin{figure}[H]
  \centering
  \includegraphics[scale=0.4]{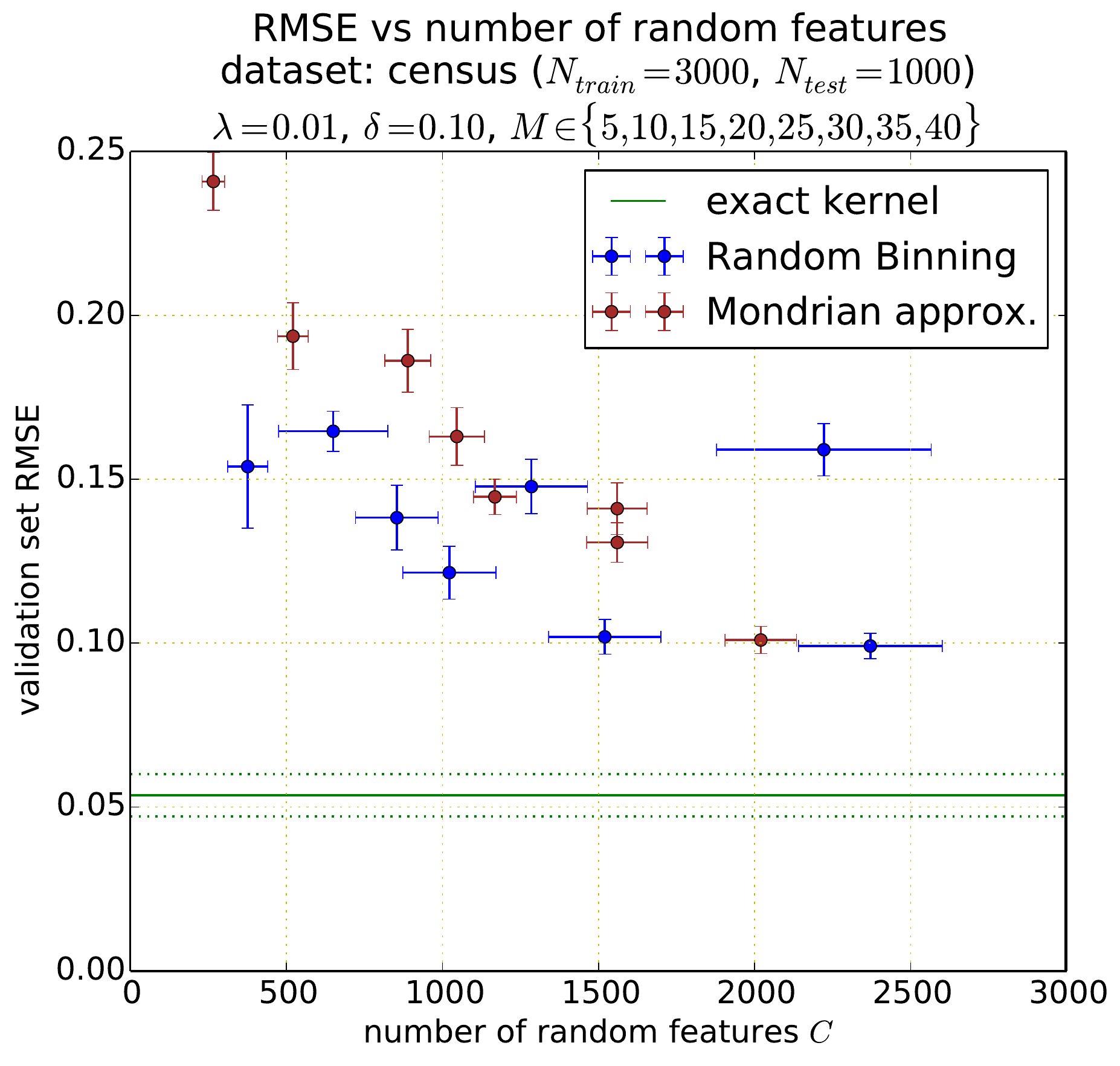}
\end{figure}
\end{minipage}
\caption{Convergence to exact kernel regressor as number $C$ of generated features increases. The left plot uses (a random subset of) the CPU dataset, while on the right (a random subset of) the Census dataset was used. CPU and Census are the two regression datasets used in \cite{rahimi2007random} where Random Binning was introduced. The horizontal axis shows the number of random features $C$ produced by the approximations, while the vertical axis shows the RMSE of the resulting regression model on a validation data set. The horizontal green line (with dotted standard deviation estimates) is the RMSE obtained when using the exact Laplace kernel. We see that the Mondrian approximation and Random Binning need to generate a similar number of random features to achieve the same predictive performance, and that this performance approaches the performance of the exact classifier as the number of random features increases. Random Binning is perhaps slightly more sensitive to randomness, as indicated by occasionally much larger standard deviations for both the number of features produced and the validation set RMSE.}
\end{figure}

As we shall see in the following chapter, the main advantage of the Mondrian approximation (over, say, Random Binning) is that it can be efficiently evaluated for all possible lifetimes $\lambda$ of the approximated kernel in a given range $\lambda \in [0, \Lambda]$. With Random Binning the approximation needs to be reconstructed from scratch for each new lifetime value.

\section{Comparison with Mondrian forest regression}

We have presented two non-linear regression models utilizing Mondrians: the Mondrian random forest and the Mondrian approximation of the Laplace kernel. In both models we independently sample $M$ partitions of the data points at hand, but these partitions are then used differently. In Appendix B on Model interpretation we briefly discuss the theoretical similarities and differences between these two models. We show that they are both linear smoothers, that they coincide in the case $M = 1$ and show that for $M > 1$ they can be interpreted as approximating two different quantities.




\chapter{Regularization paths}

The statistical complexity of many machine learning models can be controlled by adjusting their hyperparameters. In Mondrian process based models this role is played by the lifetime $\lambda$ of the Mondrian, which controls the complexity of the generated partitions.

Suitable hyperparameter values for modeling the dataset at hand are usually found by cross validation, a technique of splitting the dataset into a \textbf{training set} $\mD_{\text{train}} := \{ (\mathsf{x}_1, y_1), \hdots, (\mathsf{x}_{N_{\text{train}}}, y_{N_{\text{train}}} \})$ and a \textbf{validation set} $\mD_{\text{val}} := \{ (\mathsf{x}_{N_{\text{train}} + 1}, y_{N_{\text{train}} + 1}), \hdots, (\mathsf{x}_N, y_N) \}$, training several models on the former and choosing the hyperparameter values giving the best performance on the latter. As this procedure contaminates the validation set, we usually preserve an untouched test set on which the performance of the model with the finally chosen hyperparameters can be more accurately estimated. In the following we implicitly assume that such an independent test set is always preserved.

Training several models with different hyperparameters is often daunting and computationally expensive, especially if for each new hyperparameter configuration the model needs to be trained from scratch. It would be desirable to reuse some parts of the computation with one set of hyperparameters for training and evaluating the model with a new set of hyperparameters. The notion of computing entire regularization paths \cite{Hastie:2004:ERP:1005332.1044706} takes this idea to the extreme: it trains and evaluates the model for all possible values of a regularization hyperparameter at essentially the cost of training and evaluating a single model.

In this chapter we outline how this can be done with the lifetime parameter $\lambda$ in Mondrian process based models. We focus on regression, but the ideas also apply to classification and density estimation.

\subsection*{General setup}

All our Mondrian models start by generating $M$ Mondrian samples to provide $M$ partitions of the data points. Recall that each cut in each Mondrian is associated with a birth time $0 \leq t_b \leq \Lambda$, where $\Lambda$ is some terminal lifetime until which the Mondrians are sampled. Let $K$ be the total number of cuts in all $M$ samples combined, and let $0 < t_1 < \cdots < t_K$ be an ordered list of their times (the values are distinct with probability $1$).

Given the cuts in the Mondrian samples, the model is deterministic. So as time increases from $0$ to $\Lambda$ and new cuts appear in the trees, the model only changes at the $K$ time instants when a cut is added to one of the trees. To be able to efficiently compute the entire regularization path over the lifetime, i.e. to train and validate the model for all lifetimes $\lambda \in [0, \Lambda]$, we need to be able to perform the following operation efficiently:

\begin{tabular}{ p{0.5cm} p{13cm} }
(O1) &
  Given the model trained and evaluated with lifetime $t_i$, compute the model trained and evaluated with lifetime $t_{i + 1}$.
\end{tabular}

Sometimes we will find it easier to traverse the regularization path backwards, which is to say that we train and evaluate the model with the maximal lifetime value $\Lambda$ and then efficiently compute the results for all smaller values of the lifetime in decreasing order. For that, efficient way of performing the following operation is required:

\begin{tabular}{ p{0.5cm} p{13cm} }
(O2) &
  Given the model trained and evaluated with lifetime $t_i$, compute the model trained and evaluated with lifetime $t_{i - 1}$
\end{tabular}

Performing operations (O1) or (O2) in Mondrian random forest models for classification, regression or density estimation turns out to be quite simple, as outlined in the next section. It will be slightly more challenging for the Mondrian approximation of the Laplace kernel since the Mondrians do not directly make predictions, they only provide a randomized feature mapping.

\section{Mondrian random forest}

We traverse the regularization path forwards, starting with lifetime $\lambda = 0$ and performing operation (O1) whenever a new cut appears in any of the $M$ trees. Apart from the regression trees themselves, we maintain two global quantities: the mean squared error on a validation dataset $\text{MSE} \in \IR$ and the vector $\mathbf{\hat{y}} \in \IR^N$ whose $n$-th entry is the forest prediction at the $n$-th data point. We initialize all entries of $\mathbf{\hat{y}}$ to the mean of the predictive prior and compute the resulting $\text{MSE}$ in time $\mO(N)$.

Suppose that at time $t_i$ a leaf $l$ in tree $m$ is split into two new child leaves $l_1$, $l_2$. As predictive distributions in individual leaves are independent, the predictions only change for data points in $l$. The posterior predictive distributions in leaves $l_1$, $l_2$ can be computed analytically in time linear in the number of datapoints that end up in these leaves (see Example~\ref{GaussianPredictionModel}). To see how the model predictions $\bs{\hat{y}}$ and the MSE are updated, suppose that $\mathsf{x}_n$ is a point originally in $l$ that ends up in, say, leaf $l_1$ after the split. If $\mu_l$ is the mean of the predictive distribution in $l$ before the split and $\mu_{l_1}$ is the corresponding quantity in $l_1$ after the split, the global prediction of the forest at point $\mathsf{x}_n$ can be updated as
\begin{equation*}
\hat{y}_n'
\leftarrow \hat{y}_n - \frac{\mu_l}{M} + \frac{\mu_{l_1}}{M}
\end{equation*}
since the prediction is simply the average from the $M$ trees. The corresponding update of the MSE is
\begin{equation*}
\text{MSE}' \leftarrow \text{MSE} - \frac{(\hat{y}_n - y_n)^2}{N} + \frac{(\hat{y}'_n - y_n)^2}{N}
\end{equation*}
Note that these updates take constant time per datapoint in each split, so maintaining the predictive distributions, $\bs{\hat{y}}$ and MSE only multiplies the running time by a constant. Hence the entire regularization path is computed at essentially the same cost as training and evaluating the model at the terminal lifetime $\Lambda$. Finally, note that the RMSE can at any time be easily computed as $\text{RMSE} = \sqrt{\text{MSE}}$.

\subsection*{Examples}

Below we show examples of regularization paths for regression (on the left) and for density estimation (on the right). The validation set RMSE function is computed using the above described procedure and so is the training set RMSE function after simply taking $\mD_{\text{val}} = \mD_{\text{train}}$.

\begin{figure}[H]
\centering
\begin{minipage}[t]{0.47\textwidth}
\centering
  \includegraphics[scale=0.4]{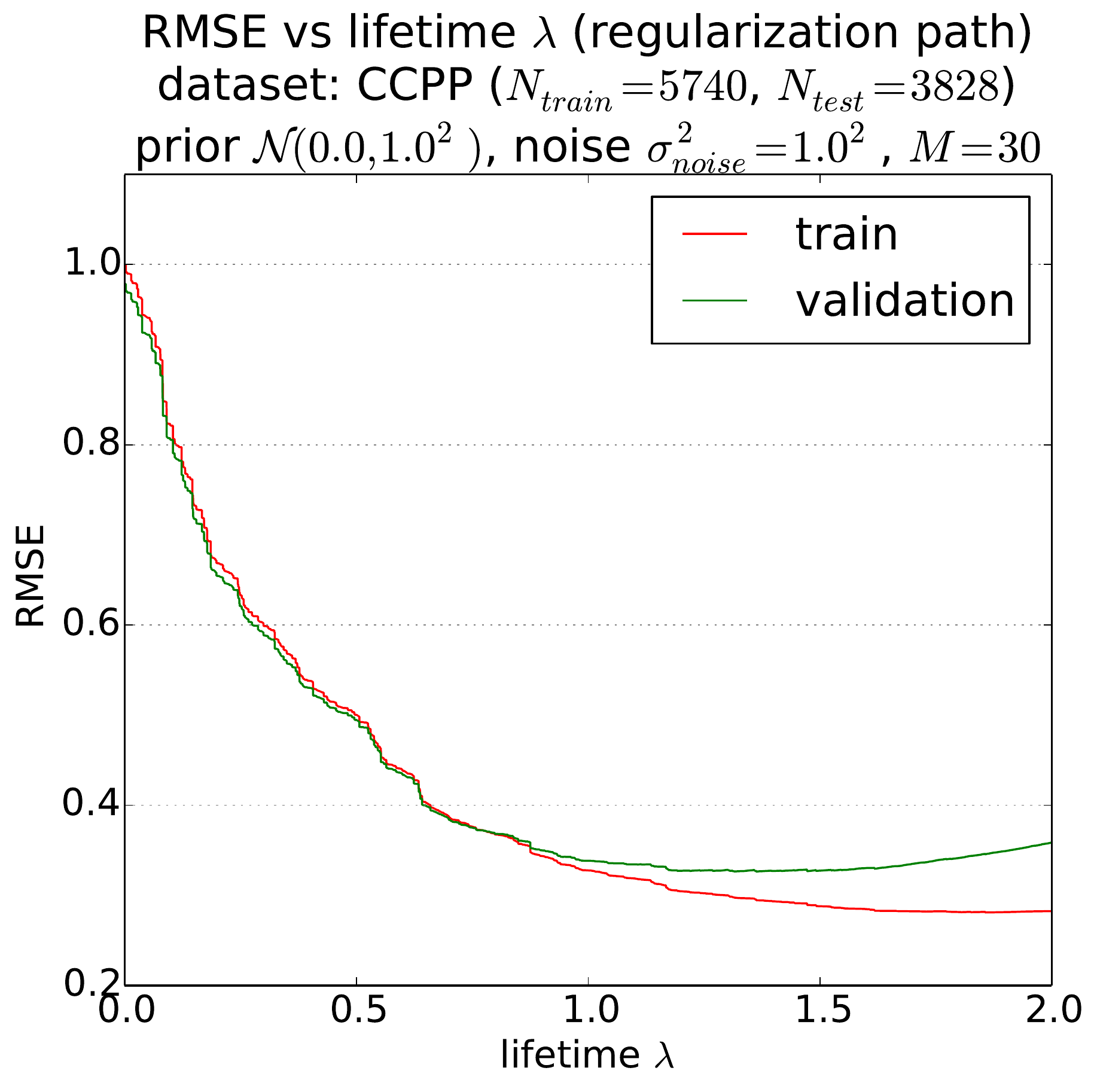}
  \caption{Mondrian random forest regression regularization path. As expected, the red training set RMSE decreases as the flexibility (complexity of partitions) of the model increases, while the green validation set RMSE reaches a minimum and then increases as the model starts to overfit to the noise in training data. Both curves are piecewise constant, with jumps occurring only at times when a cut appears in one of the $M$ trees. This may not be clearly visible only because of the large number of cuts. The regression dataset used here is Combined Cycle Power Plant \cite{Tufekci2014126}.}
\end{minipage}\hfill
\begin{minipage}[t]{0.47\textwidth}
\centering
  \includegraphics[scale=0.4]{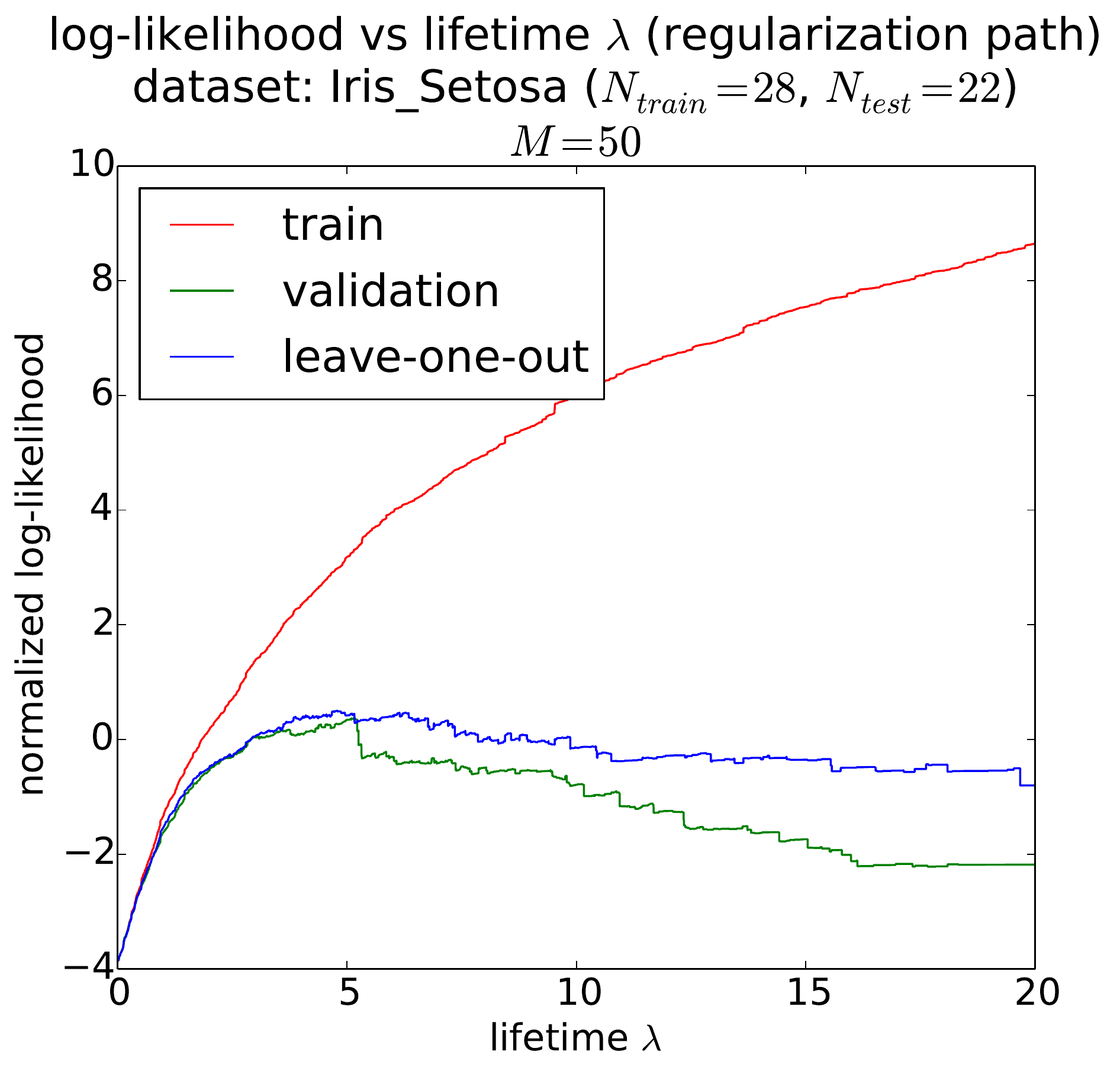}
  \caption{Mondrian random forest density estimation regularization path. The regularization path for Mondrian random forest density estimation can be computed by essentially the same procedure as for regression, with the exception that instead of predictions and mean squared errors (MSE) we maintain the likelihoods of each data point. For density estimation it is also possible to efficiently compute the leave-one-out log-likelihood. All log-likelihoods are normalized for the number of datapoints. As expected, the validation and leave-one-out likelihoods peak and then decrease as the more flexible model starts overfitting the training data. The dataset used here is the Setosa class form the Iris dataset. \cite{Lichman:2013}}
\end{minipage}
\end{figure}

\section{Laplace kernel approximation}

In the Mondrian approximation of the Laplace kernel each datapoint $\mathsf{x}$ is encoded as a (normalized) concatenation $z(\mathsf{x})$ of $M$ indicator vectors $z_1(\mathsf{x}), \hdots, z_M(\mathsf{x})$, where $z_m(\mathsf{x})$ indicates which partition cell of the $m$-th Mondrian the point $\mathsf{x}$ falls into. When a new cut appears in one of the Mondrians, a partition cell is split into two. This corresponds to replacing the feature associated with this cell by two new features, one for each child cell. Conversely, when traversing the regularization path backwards and a cut is removed from one of the Mondrians, the two features corresponding to the merged cells are replaced by their sum. (The sum of indicators of disjoint sets is the indicator of their union.)

Recall that the random feature representations of the datapoints are organized in the feature matrix $\mathbf{Z} \in \IR^{N \times C}$. Each column of $\mathbf{Z}$ corresponds to one feature, i.e. one partition cell in one of the $M$ Mondrian samples. Adding new features amounts to appending new columns to this matrix, while summing two features corresponds to summing the corresponding columns. Both operations can be carried out easily in $\mO(NC)$ time. The challenge lies in the fact that the predictions of the model are a non-trivial function of $\mathbf{Z}$: the prediction $\hat{y}$ at a test point $\mathsf{z}_{*} = z(\mathsf{x}_{*})$ is given by $\hat{y} = \bs{\theta}^T \mathsf{z}_{*}$, where $\bs{\theta}$ is the ridge regression solution
\begin{equation*}
\bs{\theta}
= ( \mathbf{Z}^T \mathbf{Z} + \delta^2 \mathbf{I}_C )^{-1} \mathbf{Z}^T \mathbf{y}
\end{equation*}
When columns of $\mathbf{Z}$ (features) are added, removed or summed  we need to efficiently update the inverse $( \mathbf{Z}^T \mathbf{Z} + \delta^2 \mathbf{I}_C )^{-1}$ in order to compute predictions under the new feature mapping. To this end, the next subsection reviews a set of general tools for efficiently updating matrix inverses under specific perturbations of the matrix that is being inverted.

\subsection*{Matrix inverse updates}

\begin{lemma}[Sherman-Morrison-Woodbury inversion formula]
\label{Woodbury}
For any matrices $\mathbf{A} \in \IR^{n \times n}$, $\mathbf{U} \in \IR^{n \times k}$, $\mathbf{C} \in \IR^{k \times k}$, $\mathbf{V} \in \IR^{k \times n}$ with $\mathbf{A}$ and $\mathbf{C}$ invertible, if the matrix $\mathbf{C}^{-1} + \mathbf{V} \mathbf{A}^{-1} \mathbf{U}$ is invertible then
\begin{equation*}
(\mathbf{A} + \mathbf{U} \mathbf{C} \mathbf{V})^{-1}
= \mathbf{A}^{-1} - \mathbf{A}^{-1} \mathbf{U} (\mathbf{C}^{-1} + \mathbf{V} \mathbf{A}^{-1} \mathbf{U} )^{-1} \mathbf{V} \mathbf{A}^{-1}
\end{equation*}
\begin{proof} Direct computation yields $(\mathbf{A} + \mathbf{U} \mathbf{C} \mathbf{V})(\mathbf{A}^{-1} - \mathbf{A}^{-1} \mathbf{U} (\mathbf{C}^{-1} + \mathbf{V} \mathbf{A}^{-1} \mathbf{U} )^{-1} \mathbf{V} \mathbf{A}^{-1}) = \mathbf{I}_n$.
\end{proof}
\end{lemma}

An important special case of this formula allows us to update a matrix inverse after a rank-one update, i.e. after the addition of a rank-1 matrix $\mathbf{u} \mathbf{v}^T$:

\begin{corollary}[Rank-1 matrix inverse update]
\label{RankOneUpdate}
Let $\mathbf{A} \in \IR^{n \times n}$ be an invertible matrix and let $\mathbf{u}, \mathbf{v} \in \IR^n$ be column vectors. If $1 + \mathbf{v}^T \mathbf{A}^{-1} \mathbf{u} \not= 0$ then
\begin{equation*}
\left( \mathbf{A} + \mathbf{u} \mathbf{v}^T \right)^{-1}
= \mathbf{A}^{-1} - \mathbf{A}^{-1} \mathbf{u} (1 + \mathbf{v}^T \mathbf{A}^{-1} \mathbf{u} )^{-1} \mathbf{v}^T \mathbf{A}^{-1}
= \mathbf{A}^{-1} - \frac{\mathbf{A}^{-1} \mathbf{u} \mathbf{v}^T \mathbf{A}^{-1}}{1 + \mathbf{v}^T \mathbf{A}^{-1} \mathbf{u}}
\end{equation*}
\begin{proof} Apply the Woodbury inversion formula (Lemma~\ref{Woodbury}) with $\mathbf{U} = \mathbf{u}$, $\mathbf{V} = \mathbf{v}^T$ and $\mathbf{C} = 1$.
\end{proof}
\end{corollary}

If $\mathbf{A}^{-1}$ is known, by bracketing the numerator in Corollary~\ref{RankOneUpdate} as $(\mathbf{A}^{-1} \mathbf{u})(\mathbf{v}^T \mathbf{A}^{-1})$ we can compute the inverse of the rank-1 updated matrix $\mathbf{A} + \mathbf{u} \mathbf{v}^T$ in time $\mO(n^2)$, as opposed to the $\mO(n^3)$ running time of matrix inversion from scratch. Note that the following operations are all rank-$1$ updates:
\begin{itemize}
\item[•] Adding the $j$-th row $\mathbf{r}_j$ to the $i$-th row $\mathbf{r}_i$. This amounts to replacing $\mathbf{r}_i$ with $\mathbf{r}_i + \mathbf{r}_j$, which can be expressed as the addition of $\mathbf{u} \mathbf{v}^T$ with $\mathbf{u} = \mathbf{e}_i$ and $\mathbf{v} = \mathbf{r}_j^T$.
\item[•] Adding the $j$-th column $\mathbf{c}_j$ to the $i$-th column $\mathbf{c}_i$. This amounts to replacing $\mathbf{c}_i$ with $\mathbf{c}_i + \mathbf{c}_j$, which can be expressed as the addition of $\mathbf{u} \mathbf{v}^T$ with $\mathbf{u} = \mathbf{c}_j$ and $\mathbf{v} = \mathbf{e}_i$.
\item[•] Add a constant $a \in \IR$ to the $i$-th entry on the main diagonal. This can be expressed as the addition of $\mathbf{u} \mathbf{v}^T$ with $\mathbf{u} = a \mathbf{e}_i$ and $\mathbf{v} = \mathbf{e}_i$.
\end{itemize}

\pagebreak

\begin{lemma}[Inverse of a submatrix]
\label{SubmatrixInverse}
Let $\mathbf{A} \in \IR^{n \times n}$ be invertible, let $1 \leq i \leq n$ and let $\tilde{\mathbf{A}}$ be the matrix obtained from $\mathbf{A}$ by deleting its $i$-th row and $i$-th column. Let $\mathbf{E}$ be the submatrix of $\mathbf{A}^{-1}$ obtained by deleting its $i$-th row and column, let $\mathbf{f}$ be the $i$-th column of $\mathbf{A}^{-1}$ with the $i$-th entry removed, let $\mathbf{g}$ be the $i$-th row of $\mathbf{A}^{-1}$ with the $i$-th entry removed, and finally let $h$ be the $(i, i)$ entry of $\mathbf{A}^{-1}$.
\\
If $h \not= 0$ then $\tilde{\mathbf{A}}$ is invertible and its inverse is $\tilde{\mathbf{A}}^{-1} = \mathbf{E} - \mathbf{f} \mathbf{g}^T / h$.
\begin{proof} The proof appears as Lemma~\ref{SubmatrixInverseProof} in the appendix.
\end{proof}
\end{lemma}

This lemma is sufficiently general for us, since the rows and columns of matrices that we will be inverting correspond to features in the same order and so we won't need to remove the $i$-th row and $j$-th column for $i \not= j$. Also, note that given $\mathbf{A}^{-1}$, the update $\tilde{\mathbf{A}}^{-1} = \mathbf{E} - \mathbf{f} \mathbf{g}^T / h$ can be computed in $\mO(n^2)$ time as $\mathbf{E}$, $\mathbf{f}$, $\mathbf{g}$ and $h$ can be easily extracted from $\mathbf{A}^{-1}$.

\begin{lemma}[Inverse of an extended matrix]
\label{SupermatrixInverse}
Let $\mathbf{A} \in \IR^{n \times n}$ be invertible. For $\mathbf{b} \in \IR^n$, $\mathbf{c} \in \IR^n$ and $d \in \IR$, the extended matrix
\begin{equation*}
  \begin{bmatrix}
  \mathbf{A} & \mathbf{b} \\
  \mathbf{c}^T & d
  \end{bmatrix}
\end{equation*}
is invertible if and only if its Schur complement $s := d - \mathbf{c}^T \mathbf{A}^{-1} \mathbf{b} \not= 0$, in which case the inverse is
\begin{equation*}
  \begin{bmatrix}
  \mathbf{A} & \mathbf{b} \\
  \mathbf{c}^T & d
  \end{bmatrix}^{-1}
= \begin{bmatrix}
  \mathbf{E} & \mathbf{f} \\
  \mathbf{g}^T & h
  \end{bmatrix}
  \hs\text{where}\hs
  \begin{array}{rclrcl}
  \mathbf{E} & = & \mathbf{A}^{-1} + s^{-1} \mathbf{A}^{-1} \mathbf{b} \mathbf{c}^T \mathbf{A}^{-1} &
  \hs \mathbf{f} & = & - s^{-1} \mathbf{A}^{-1} \mathbf{b} \\
  \mathbf{g} & = & - s^{-1} \mathbf{c}^T \mathbf{A}^{-1} &
  h & = & s^{-1}
  \end{array}
\end{equation*}
This inverse can be computed from $\mathbf{A}^{-1}$ in time $\mO(n^2)$.
\begin{proof} Appears as Lemma~\ref{SupermatrixInverseProof} in the appendix.
\end{proof}
\end{lemma}

\subsection*{Removing a cut}

To demonstrate different concepts, we choose to traverse the regularization path backwards for this model. As discussed in the introduction of this chapter, we train and evaluate the model for some terminal lifetime value $\Lambda$ and then seek to efficiently revert each cut one-by-one (operation (O2)), in decreasing order of their birth times.

Suppose we want to revert the effect of cut $c$ with birth time $t_b$ appearing in the $m$-th Mondrian sample. We assume we have access to the feature matrix $\mathbf{Z}$ and the inverse $\mathbf{C}^{-1} = ( \mathbf{Z}^T \mathbf{Z} + \delta^2 \mathbf{I}_C )^{-1}$ corresponding to features generated by the Mondrians with lifetime $t_b$ (i.e., with the cut $c$ present in the $m$-th Mondrian). Let $i$, $j$ be the indices of the two features introduced by the cut $c$. Reverting this cut amounts to merging these two features together and since they are (rescaled) indicators of disjoint sets, this is equivalent to summing the $i$-th and $j$-th columns $\mathbf{z}_i$, $\mathbf{z}_j$ of $\mathbf{Z}$ together. Thus our goal is to obtain the updated inverse
\begin{equation*}
\tilde{\mathbf{C}}^{-1}
= ( \tilde{\mathbf{Z}}^T \tilde{\mathbf{Z}} + \delta^2 \mathbf{I}_{\tilde{C}} )^{-1}
\end{equation*}
where $\tilde{C} = C - 1$ and $\tilde{\mathbf{Z}}$ is the matrix obtained from $\mathbf{Z}$ by replacing its $i$-th and $j$-th column by their sum. (The sum replaces the $i$-th column, and the $j$-th column is removed, say.)

We express the operation of computing $\tilde{\mathbf{C}}$ from $\mathbf{C}$ as a sequence of four operations in such a way that after performing each individual one the resulting matrix is still invertible and the inverse can be computed in time $\mO(C^2)$ using the above introduced tools.

\begin{algorithm}
  \begin{algorithmic}[1]
  	\State Add row $j$ to row $i$
  	\Comment{Rank $1$ update, Corollary~\ref{RankOneUpdate}}
  	\State Add column $j$ to column $i$
	\Comment{Rank $1$ update, Corollary~\ref{RankOneUpdate}}
  	\State Delete the $j$-th row and $j$-th column
  	\Comment{Inverse of a submatrix, Lemma~\ref{SubmatrixInverse}}
    \State Subtract $\delta^2$ from the $(i, i)$ entry
    \Comment{Rank $1$ update, Corollary~\ref{RankOneUpdate}}
  \end{algorithmic}
\end{algorithm}
It is important to carry out steps (3) and (4) in this order, so as to guarantee existence of the inverse after each step. The matrix remains invertible after steps (1) and (2) because adding a row to another row (or a column to another column) is an elementary operation that preserves the rank of the matrix. The inverses after performing steps (3) and (4) are guaranteed to exist because the resulting matrices are in both cases positive definite, as can be easily checked.

Note that we do not in fact require maintaining the matrix $\mathbf{C}$; it suffices to maintain $\mathbf{C}^{-1}$ and $\mathbf{Z}$ as they contain all that is required to perform the updates to $\mathbf{C}^{-1}$.

\subsection*{Implementation}

We start by computing the Mondrian approximation of the Laplace kernel with the terminal lifetime $\Lambda$. This produces $M$ Mondrian trees, each consisting of a hierarchy of cuts with birth times $t_b \in [0, \Lambda]$. We traverse through these cuts in decreasing order of birth time, at each birth time removing the corresponding cut. The previous section describes how the matrix $\mathbf{C}^{-1}$ can be appropriately updated in time $\mO(C^2)$. Having access to this updated inverse, predictions on the validation set can be made via $\hat{y} = \mathsf{z}_{*}^T \bs{\theta}$, where $\bs{\theta} = \mathbf{C}^{-1} (\mathbf{Z}^T \mathbf{y})$. Using the shown bracketing, the parameter vector $\bs{\theta}$ can be computed in time $\mO(C^2 + C N_{\text{train}})$. The predictions on $N_{\text{val}}$ validation points and the resulting RMSE can then be computed in time $\mO(N_{\text{val}} C)$. Hence the total time complexity of removing a single cut is $\mO(C^2 + CN)$. As there are $C - M$ cuts to be removed before lifetime $0$ is reached, the time complexity of traversing the entire regularization path from $\Lambda$ down to $0$ is $\mO(C^3 + C^2 N)$. This is the same as the cost of training and evaluating the initial model with lifetime $\Lambda$.

\begin{figure}[H]
  \centering
  \includegraphics[scale=0.4]{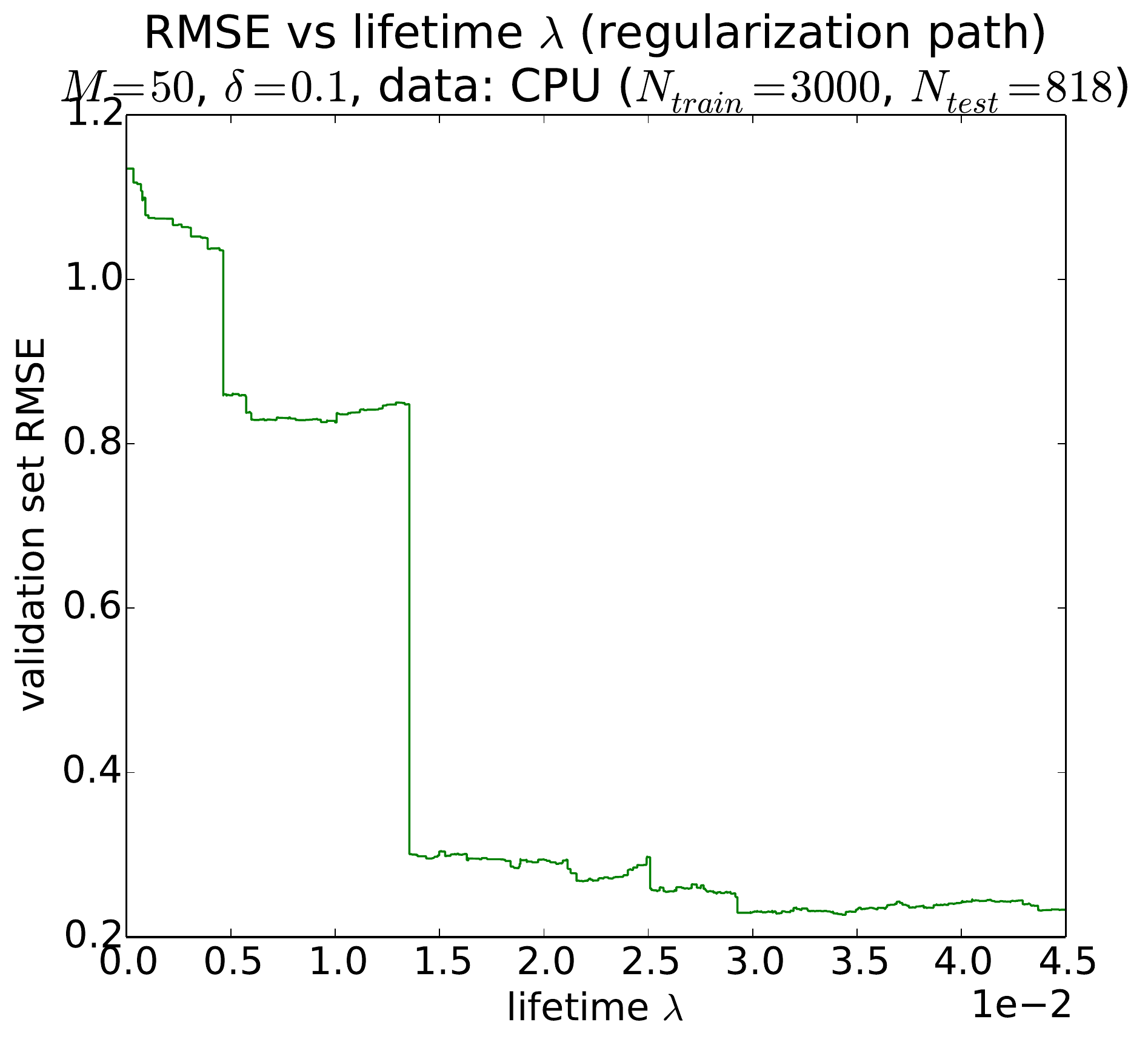}
  \caption{Regularization path of Laplace kernel approximation using $M = 50$ Mondrian trees. The piecewise constant function shows the validation set RMSE of the resulting regression model as a function of the lifetime. In this configuration the regularization path seems to be sensitive to randomness, since large jumps occur at times when particularly well placed cuts appear in one of the trees. The dataset used is (a random subset of) CPU, one of the regression datasets used in \cite{rahimi2007random}.}
\end{figure}

\subsection*{Conclusion}

Initially we have introduced the Mondrian approximation of the symmetric Laplace kernel as a way of avoiding the computationally expensive inversion of an $N \times N$ (regularized) kernel matrix. While this reason still holds, efficient computation of the entire regularization path of this approximation leads to another use case: even if a single $\mO(N^3)$ computation with exact Laplace kernel is feasible, we may want to use the Mondrian approximation to efficiently find a value of the lifetime that performs best on the validation set. As the lifetime can be seen as controlling model complexity, this yields an efficient procedure for determining a suitable model complexity for the dataset at hand. When using the exact Laplace kernel, we would probably need to retrain the model from scratch for several values of the lifetime, each value requiring a new $\mO(N^3)$ matrix inversion.

\chapter{Mondrian Grid}

We have seen how the Mondrian process is useful for approximating the symmetric Laplace kernel, sharing a common lifetime $\lambda$ for all $D$ input dimensions. Moreover, we have seen that the entire regularization path over the lifetime $\lambda$ can be efficiently computed, leading to efficient determination of the right model complexity for a dataset at hand. In this chapter we seek to achieve the same goal with a more general Laplace kernel where different dimensions are allowed to have different lifetimes:

\begin{definition} The (general) \textbf{Laplace kernel} is given by
\begin{equation*}
k(\mathsf{x}, \mathsf{x}')
= \exp\left( - \sum_{d = 1}^D \lambda_d |x_d - x'_d| \right)
\end{equation*}
where $\lambda_1, \hdots, \lambda_D$ are \textbf{lifetime} parameters of the kernel.
\end{definition}
If we were only interested in training a single model with a fixed lifetime configuration $\bs{\lambda} = (\lambda_1, \hdots, \lambda_D)$, we could still use the previous Mondrian approximation after rescaling each input dimension $d$ by $\lambda_d$ and then using a symmetric Laplace kernel of lifetime $1$. However, we are interested in an efficient procedure for the cross validation problem
\begin{equation*}
\argmin_{\lambda_1, \hdots, \lambda_D} \text{error}( (\lambda_1, \hdots, \lambda_D), \mD_{\text{val}})
\end{equation*}
The Mondrian process has the limitation that when stopped at a single lifetime $\lambda$, this lifetime is common to all dimensions. Therefore the earlier presented Mondrian approximation of the Laplace kernel is only useful for cross validation if ratios of lifetimes in different dimensions can be fixed. In this chapter we propose a method that does away with this requirement, allowing us to adjust the approximated lifetime in each dimension independently. This model is no longer based on a $D$-dimensional Mondrian process, but rather on $D$ independent one-dimensional Mondrian processes (which have been shown to coincide with Poisson point processes in Theorem~\ref{Mondrian1DIsPPP}).

\subsubsection{Mondrian grid approximator}

A Mondrian grid is a collection of $D$ independent one-dimensional Mondrian processes $M^{(1)}, \hdots, M^{(D)}$, where $M^{(d)}$ is assumed to run on the $d$-th coordinate axis of $\IR^D$.

\begin{wrapfigure}{r}{0.34\textwidth}
\vspace{-2.2em}
\begin{center}
\begin{tikzpicture}[scale=1.5]
  \draw[->] (-1,0) -- (1.7,0) node[right] {$d = 1$};
  \draw[->] (0,-1.1) -- (0,2) node[above] {$d = 2$};

  \draw[dashed, thin, draw=gray] (-0.6,-1.1) -- (-0.6,2);
  \draw[dashed, thin, draw=gray] (+0.3,-1.1) -- (+0.3,2);
  \draw[dashed, thin, draw=gray] (+0.9,-1.1) -- (+0.9,2);
  \draw[dashed, thin, draw=gray] (+1.3,-1.1) -- (+1.3,2);

  \draw[dashed, thin, draw=gray] (-1,-0.7) -- (1.7,-0.7);
  \draw[dashed, thin, draw=gray] (-1,+1.2) -- (1.7,+1.2);
  \draw[dashed, thin, draw=gray] (-1,+1.7) -- (1.7,+1.7);
  
  \draw node[below left] {$0$};
  \fill[shift={(0.5,0.3)}] circle[radius=1pt] node[below] {$\mathsf{a}$};
  \fill[shift={(0.7,0.8)}] circle[radius=1pt] node[below] {$\mathsf{b}$};
  \fill[shift={(1.15,0.9)}] circle[radius=1pt] node[below] {$\mathsf{c}$};
  
  \foreach \x/\xtext in 
  {-0.6/x^{(1)}_1,
   +0.3/x^{(1)}_2,
   +0.9/x^{(1)}_3,
   +1.3/x^{(1)}_4}
    \fill[shift={(\x,0)}] circle[radius=1.5pt] node[below] {$\xtext$};

  \foreach \y/\ytext in
  {-0.7/x^{(2)}_1,
   +1.2/x^{(2)}_2,
   +1.7/x^{(2)}_3}
    \fill[shift={(0,\y)}] circle[radius=1.5pt] node[left] {$\ytext$};
\end{tikzpicture}
\end{center}
\caption{A sample of a Mondrian grid in 2 dimensions. The point $x^{(d)}_i$ is a cut location in the sample from the Mondrian $M^{(d)}$, the dashed lines show the corresponding cuts of $\IR^2$. Points $\mathsf{a}$ and $\mathsf{b}$ are in the same grid cell, whereas the point $\mathsf{c}$ falls into a different one. Thus here $N = 3$ and $C = 2$.}
\label{MondrianGridIllustration}
\vspace{-5em}
\end{wrapfigure}
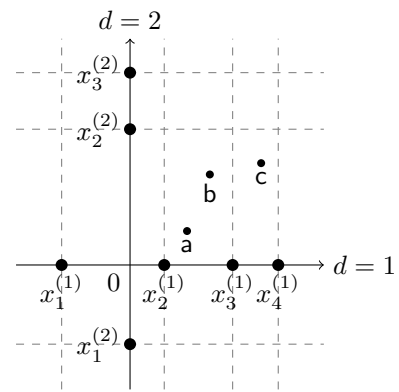

Suppose we sample a Mondrian grid, which is to say that we sample from independent one-dimensional Mondrian processes along each coordinate axis, say until a lifetime $\lambda_d$ in dimension $d$. The cut locations $x^{(d)}_1 < \cdots < x^{(d)}_{C_d}$ of $M^{(d)}$ provide a partitioning of the $d$-th coordinate axis, which in turn yields a partitioning of $\IR^D$ by hyperplanes orthogonal to the $d$-th coordinate axis, crossing it at the cut locations of $M^{(d)}$. (See Figure~\ref{MondrianGridIllustration} for a 2D illustration, where these hyperplanes are dashed lines.)

The cuts induced by all the $D$ Mondrian samples together partition $\IR^D$ into \textbf{cells}, maximal connected subsets of $\IR^D$ not intersecting any cutting hyperplane. Unlike in a $D$-dimensional Mondrian process, these cuts extend all the way through space, uninterrupted by cuts in different dimensions. Hence the name Mondrian grid.

Let $C$ be the number of non-empty grid cells. Similarly as with the Mondrian approximation of the Laplace kernel, our random feature mapping $z : \IR^D \to \IR^C$ maps each datapoint $\mathsf{x}$ to an indicator vector $z(\mathsf{x})$ of the grid cell into which $\mathsf{x}$ falls. Dot products in this feature space are then
\begin{equation*}
z(\mathsf{x})^T z(\mathsf{y})
= \begin{dcases}
  1 & \text{ if } \mathsf{x}, \mathsf{y} \text{ fall into the same grid cell } \\
  0 & \text{ otherwise }
\end{dcases}
\end{equation*}
Using independence of the processes $M^{(1)}, \hdots, M^{(D)}$ on each axis we have that
\begin{IEEEeqnarray*}{rCll}
\IP(z(\mathsf{x})^T z(\mathsf{y}) = 1) & = &
  \IP\left( \bigcap_{d = 1}^D \left\{ \text{no cut between } x_d \text{ and } y_d \text{ in dimension } d \right\} \right)
  \hs &
\\ & = &
  \prod_{d = 1}^D \IP( \text{no cut between } x_d \text{ and } y_d \text{ in dimension } d )
  & \text{[ independence ]}
\\ & = &
  \prod_{d = 1}^D \IP\left( \text{Exp}\left( |x_d - y_d| \right) > \lambda_d \right)
  & \text{[ self-consistency ]}
\\ & = &
  \prod_{d = 1}^D \exp\left( - \lambda_d |x_d - y_d| \right)
\\ & = &
  \exp\left( - \sum_{i = 1}^D \lambda_d |x_d - y_d| \right)
\end{IEEEeqnarray*}
As with the previous Mondrian approximation, inner products are a Bernoulli random variables with expectations equal to the desired kernel values and by concatenating feature vectors $z_1(\mathsf{x}), \hdots, z_M(\mathsf{x})$ from $M$ independent grids into a single feature vector $z(\mathsf{x})$ (and normalizing with $M^{-1/2}$) we obtain the Monte Carlo estimator
\begin{equation*}
\left( \frac{z(\mathsf{x})}{\sqrt{M}} \right)^T \left( \frac{z(\mathsf{y})}{\sqrt{M}} \right)
= \frac{1}{M} z(\mathsf{x})^T z(\mathsf{y})
= \frac{1}{M} \sum_{m = 1}^M z_m(\mathsf{x})^T z_m(\mathsf{y})
\;\longrightarrow\;
\IE\left[ z_1(\mathsf{x})^T z_1(\mathsf{y}) \right]
= e^{- \sum_{i = 1}^d \lambda_d |x_d - y_d|}
\end{equation*}
whose standard deviation decreases as $\mO(M^{-1/2})$ as $M \to \infty$.

\section{Regularization paths}

With the Mondrian grid approximation we can adjust the approximated lifetime in dimension $d$ individually, by changing the lifetime $\lambda_d$ of the one-dimensional Mondrian on the $d$-th coordinate axis. Moreover, it is not necessary to discard the existing grid sample when such a change is made; we only need to add or remove cuts in dimension $d$ according to whether $\lambda_d$ was increased or decreased. In this section we discuss how the predictions of the resulting regression model (using the feature mapping $z$) can be updated when such a change of lifetime in an individual dimension is performed.

\subsection*{Initialization}

For each dimension $d$, let $N_d$ be the number of distinct $d$-coordinates of all $N$ data points (training and validation combined) and let $\tilde{x}^{(d)}_1 < \cdots < \tilde{x}^{(d)}_{N_d}$ be a sorted list of their values.

Observe that the Mondrian grid approximation only depends on how the datapoints are partitioned into cells by the grid. It does not depend on the number of cuts that separate two points, or on the precise location of these cuts. More concretely, the partitioning of the datapoints only depends on whether there is or isn't a cut in the interval $(\tilde{x}^{(d)}_{i - 1}, \tilde{x}^{(d)}_i)$, for each $2 \leq i \leq N_d$ and each $1 \leq d \leq D$. So if for each such interval $(\tilde{x}^{(d)}_{i - 1}, \tilde{x}^{(d)}_i)$ we compute the birth time of the first cut $t^{(d)}_i$ appearing in it, the set of cuts that yield a grid approximating the Laplace kernel with lifetimes $\bs{\lambda} = (\lambda_1, \hdots, \lambda_D)$ is given by taking the cuts with birth times $t^{(d)}_i \leq \lambda_d$ in dimension $d$. Then by including or removing some of these cuts we will be able to easily change the lifetimes of the approximated Laplace kernel.

By self-consistency of the Mondrian process, the distribution of the time of the first cut in an interval $(\tilde{x}^{(d)}_{i - 1}, \tilde{x}^{(d)}_i)$ is $\Exp(\tilde{x}^{(d)}_i - \tilde{x}^{(d)}_{i - 1})$. We sample this quantity for each such interval $M$ times independently, once for each grid. There is no need to sample the exact locations of the cuts, but they would of course be uniformly distributed in the interval.

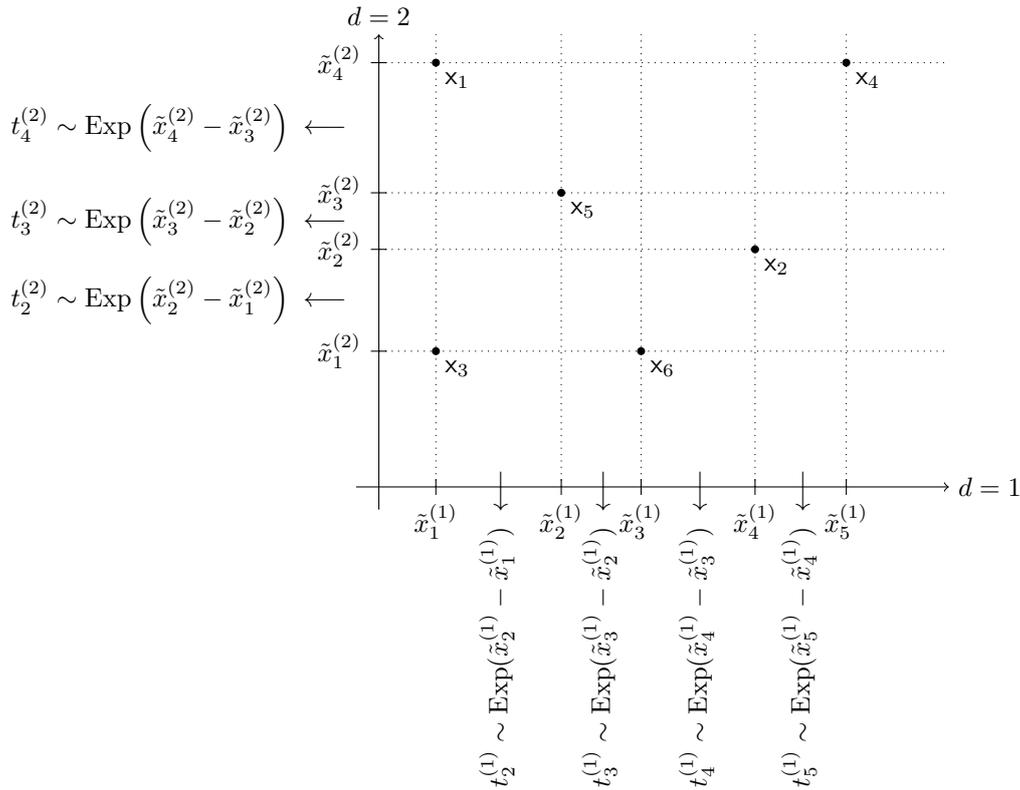
\begin{figure}[H]
\begin{center}
\begin{tikzpicture}[scale=1.5]
  \draw[->] (-0.2,0) -- (5,0) node[right] {$d = 1$};
  \draw[->] (0,-0.2) -- (0,4) node[above] {$d = 2$};

  \foreach \x/\xtext in 
  {0.5/\tilde{x}^{(1)}_1,
   1.6/\tilde{x}^{(1)}_2,
   2.3/\tilde{x}^{(1)}_3,
   3.3/\tilde{x}^{(1)}_4,
   4.1/\tilde{x}^{(1)}_5}
    \draw[shift={(\x,0)}] (0pt,2pt) -- (0pt,-2pt) node[below] {$\xtext$};

  \foreach \i/\j/\x in 
  {1/2/1.05,
   2/3/1.95,
   3/4/2.80,
   4/5/3.70}
  {
    \draw[shift={(\x-0.25,-1.25)}] node[below, rotate=90] {$t^{(1)}_\j \sim \Exp(\tilde{x}^{(1)}_\j - \tilde{x}^{(1)}_\i) \; \longleftarrow$};
  }

  \draw[dotted] (0.5,0) -- (0.5,4);
  \draw[dotted] (1.6,0) -- (1.6,4);
  \draw[dotted] (2.3,0) -- (2.3,4);
  \draw[dotted] (3.3,0) -- (3.3,4);
  \draw[dotted] (4.1,0) -- (4.1,4);

  \foreach \y/\ytext in
  {1.2/\tilde{x}^{(2)}_1,
   2.1/\tilde{x}^{(2)}_2,
   2.6/\tilde{x}^{(2)}_3,
   3.75/\tilde{x}^{(2)}_4}
    \draw[shift={(0,\y)}] (2pt,0pt) -- (-2pt,0pt) node[left] {$\ytext$};

  \foreach \i/\j/\x in 
  {1/2/1.65,
   2/3/2.35,
   3/4/3.175}
  {
    \draw[shift={(-0.2,\x)}] node[left] {$t^{(2)}_\j \sim \Exp\left( \tilde{x}^{(2)}_\j - \tilde{x}^{(2)}_\i \right) \; \longleftarrow$};
  }

  \draw[dotted] (0,1.2)  -- (5,1.2);
  \draw[dotted] (0,2.1)  -- (5,2.1);
  \draw[dotted] (0,2.6)  -- (5,2.6);
  \draw[dotted] (0,3.75) -- (5,3.75);

  \fill (0.5, 3.75) circle[radius=1pt] node[below right] {$\mathsf{x}_1$};
  \fill (0.5, 1.2)  circle[radius=1pt] node[below right] {$\mathsf{x}_3$};
  \fill (1.6, 2.6)  circle[radius=1pt] node[below right] {$\mathsf{x}_5$};
  \fill (2.3, 1.2)  circle[radius=1pt] node[below right] {$\mathsf{x}_6$};
  \fill (3.3, 2.1)  circle[radius=1pt] node[below right] {$\mathsf{x}_2$};
  \fill (4.1, 3.75) circle[radius=1pt] node[below right] {$\mathsf{x}_4$};
\end{tikzpicture}
\end{center}
\caption{Example of unique coordinate values $\tilde{x}^{(d)}_i$ for $D = 2$ dimensions and $N = 6$ data points. Distributions of the time of first cut in each interval are also shown.}
\end{figure}

Before traversing a regularization path along lifetime configurations, we need to pick a starting configuration $\bs{\lambda}_0 = (\lambda_{01}, \hdots, \lambda_{0D})$. For this lifetime configuration we compute the feature matrix $\mathbf{Z} \in \IR^{N \times C}$ in time $\mO(N C)$, where $C$ is the total number of non-empty grid cells in all $M$ grids, where each grid consists of those cuts in each dimension $d$ that have birth time $t_b \leq \lambda_{0d}$. We also compute the inverse $\mathbf{C}^{-1} = (\mathbf{Z}^T \mathbf{Z} + \delta^2 \mathbf{I}_C)^{-1}$ using any standard method in time $\mO(C^3)$.

\begin{example} A natural initialization point might be the lifetime configuration $\bs{\lambda}_0 = \mathbf{0}$, in which case all $M$ grids contain no cuts and so all datapoints fall into the same cell in each grid. Then $\mathbf{Z}$ is an $N \times M$ matrix with all entries equal to $1 / \sqrt{M}$ (each entry $z_{nm}$ indicates that the $n$-th datapoint falls into the only grid cell in the $m$-th grid) and the regularized covariance matrix $\mathbf{C} = \mathbf{Z}^T \mathbf{Z} + \delta^2 \mathbf{I}_M$ has all non-diagonal entries equal to
\begin{equation*}
\sum_{n = 1}^N \frac{1}{\sqrt{M}} \frac{1}{\sqrt{M}} = \frac{N}{M}
\end{equation*}
and all diagonal entries equal to $\frac{N}{M} + \delta^2$. Its inverse $\mathbf{C}^{-1} = (\mathbf{Z}^T \mathbf{Z} + \delta^2 \mathbf{I}_M)^{-1}$ can be computed in time $\Theta(M^3)$ using any standard matrix inversion algorithm. The inverse is guaranteed to exist for $\delta > 0$ as the matrix is positive definite.
\end{example}

\subsection*{Increasing a lifetime}

Say we want to increase the lifetime in dimension $d$ and as a result a new cut is added to the $m$-th grid in an interval $(\tilde{x}^{(d)}_{i - 1}, \tilde{x}^{(d)}_i)$, for some $m$ and $i$. When this cut is added to the grid, \emph{all} cells intersected by this cut are split into two. (Note that in the standard Mondrian approximation, a cut only split one cell.) However, we will only split those cells where after the split both resulting grid cells will contain a datapoint. By carrying out this check we ensure that none of the grids contributes a feature that has value $0$ for all datapoints.

Say we've identified that feature $\mathbf{z}_i = (z_{1i}, \hdots, z_{ni})$ (the $i$-th column of $\mathbf{Z}$) is one of those features that need to be split by the newly added cut. We construct the two sets
\begin{IEEEeqnarray*}{rCll+x*}
S_0 & := &
  \left\{ 1 \leq n \leq N \mid z_{ni} > 0, x_{nd} \leq \tilde{x}^{(d)}_{i - 1} \right\}
\\
S_1 & := &
  \left\{ 1 \leq n \leq N \mid z_{ni} > 0, x_{nd} \geq \tilde{x}^{(d)}_i \right\}
\end{IEEEeqnarray*}
of indices of datapoints to the left and to the right of the newly added cut, respectively. Note that even though the location of the new cut is not determined exactly within $(\tilde{x}^{(d)}_{i - 1}, \tilde{x}^{(d)}_i)$, this is sufficient because no datapoint has $d$-coordinate lying in this open interval by definition. The feature vectors corresponding to the two new grid cells are then
\begin{IEEEeqnarray*}{rCll+x*}
\mathbf{z}'_0 & := &
  \frac{1}{\sqrt{M}} \left( \II( 1 \in S_0 ), \hdots, \II( N \in S_0 ) \right)
\\
\mathbf{z}'_1 & := &
  \frac{1}{\sqrt{M}} \left( \II( 1 \in S_1 ), \hdots, \II( N \in S_1 ) \right)
\end{IEEEeqnarray*}

Now we need to remove feature $\mathbf{z}_i$, add the two new features $\mathbf{z}'_0$, $\mathbf{z}'_1$ to the matrix $\mathbf{Z}$ and update the inverse $\mathbf{C}^{-1} = (\mathbf{Z}^T \mathbf{Z} + \delta^2 \mathbf{I}_C)^{-1}$ accordingly. We also need to compute new predictions $\hat{\mathbf{y}}$ on the validation set and determine the resulting RMSE.
\begin{itemize}
\item[(1)] Deleting the $i$-th column of $\mathbf{Z}$ and appending two new columns $\mathbf{z}'_0$, $\mathbf{z}'_1$ to the end can be performed easily in time $\mO(N C)$. (This allows for reallocating memory for $\mathbf{Z}$ if necessary.)
\item[(2)] The $i$-th row and $i$-th column of the regularized covariance matrix $\mathbf{C}$ correspond to the removed feature, so we would delete this row and column from $\mathbf{C}$. Lemma~\ref{SubmatrixInverse} on the inverse of a submatrix tells us how to update $\mathbf{C}^{-1}$ when the $i$-th row and column of $\mathbf{C}$ are deleted, in time $\mO(C^2)$.

The two new features $\mathbf{z}'_0$, $\mathbf{z}'_1$ appended to $\mathbf{Z}$ manifest themselves as two new columns and two new rows appended to $\mathbf{C}$, where each new entry is a covariance between two features, except for the two new diagonal entries which are the variances of the two new features plus the $\delta^2$ regularization terms. Lemma~\ref{SupermatrixInverse} on the inverse of an extended matrix tells us how to update $\mathbf{C}^{-1}$ when a new row and column are added to the end of $\mathbf{C}$, in time $\mO(C^2)$. We apply this procedure twice, first for feature $\mathbf{z}'_0$ and then for $\mathbf{z}'_1$.

Note that we only need to update $\mathbf{C}^{-1}$, there is no need to maintain the matrix $\mathbf{C}$ itself.
\item[(4)] Given the updated inverse $\mathbf{C}^{-1}$, the ridge regression solution $\bs{\theta}^{\text{MAP}}$ is
\begin{equation*}
\bs{\theta}^{\text{MAP}}
= \mathbf{C}^{-1} \mathbf{Z}_{\text{train}}^T \mathbf{y}_{\text{train}}
\end{equation*}
and can be computed in time $\mO(CN)$ by bracketing the expression as $\bs{\theta}^{\text{MAP}} = \mathbf{C}^{-1} (\mathbf{Z}_{\text{train}}^T \mathbf{y}_{\text{train}})$.
\item[(5)] Given the updated ridge regression solution $\bs{\theta}^{\text{MAP}}$, predictions on the validation set and the resulting RMSE can be easily computed in time $\mO(CN)$.
\end{itemize}
We repeat steps (1)-(5) for each feature that is split into two non-empty features by the newly added cut. If $K$ is the number of such features, adding this cut takes $\mO(K (C^2 + CN))$ time.

\subsection*{Decreasing a lifetime}

Now suppose we want to decrease the lifetime along a dimension $d$ and as a result a cut disappears from the $m$-th grid in an interval $(\tilde{x}^{(d)}_{i - 1}, \tilde{x}^{(d)}_i)$, for some $m$ and $i$. At this stage all pairs of cells that have \emph{only} been separated by this cut need to be merged pairwise together. Thus the problem of decreasing the lifetime decomposes into two parts:
\begin{itemize}
\item[(1)] Determining which pairs of features should be merged.
\item[(2)] Merging (summing) those pairs of features and correspondingly updating the matrices $\mathbf{Z}$, $\mathbf{C}^{-1}$, the model predictions and the resulting validation set RMSE.
\end{itemize}
To solve (1), for each grid cell we maintain a pointer to both its neighbours in each of the $D$ dimensions. When a cut in dimension $d$ is removed, for each cell we check whether it is this cut that separates it from one of its neighbours in dimension $d$. If so, these two features are to be merged.

Once (1) is done, merging a pair of features can be performed simply by removing the two features from $\mathbf{Z}$ and then appending a feature that equals their sum. We have already seen in the previous subsection how $\mathbf{Z}$, $\mathbf{C}^{-1}$ and the predictions can be efficiently updated in time $\mO(C^2 + CN)$ when features are added or removed.

\section{Lifetime configuration exploration}

In the previous section we have described how a regularization path can be traversed by efficiently increasing or decreasing the lifetime in one dimension individually. However, our ultimate goal is to discover a configuration of lifetimes $\bs{\lambda} = (\lambda_1, \hdots, \lambda_D)$ for the $D$ input dimensions that works well for the dataset at hand. As before, we split the dataset into a training set $\mD_{\text{train}} = \{ (\mathsf{x}_1, y_1), \hdots, (\mathsf{x}_{N_{\text{train}}}, y_{N_{\text{train}}} \})$ and a validation set $\mD_{\text{val}} = \{ (\mathsf{x}_{N_{\text{train}} + 1}, y_{N_{\text{train}} + 1}), \hdots, (\mathsf{x}_N, y_N) \}$ and seek a configuration of lifetimes that minimizes the RMSE on the validation set $\mD_{\text{val}}$ of a model trained on $\mD_{\text{train}}$ using this configuration.

We propose to use the Mondrian grid approximation, where evaluation of the validation set RMSE for different lifetime configurations can be performed more efficiently than recomputing it for each configuration individually. This is because having trained the model (computed the matrices $\mathbf{Z}$ and $\mathbf{C}^{-1}$) for one lifetime configuration, moving to a neighbouring lifetime configuration can be done efficiently using the methods described in the previous section.

To decide which lifetime configuration to explore next based on the history of already explored configurations, we need an optimization procedure. The following is a very simple local optimizer that greedily increases the lifetime in the dimension that leads to lowest immediate RMSE on the validation set:

\begin{algorithm}
  \begin{algorithmic}[1]
  	\State Save $\mathbf{Z}$ and $\mathbf{C}^{-1}$
  	\For{$d = 1$ \textbf{to} $D$}
  		\State $c_d \gets $ first cut in dimension $d$ with birth time strictly larger than $\lambda_d$
  		\State $\mathbf{Z}_{(d)}, \mathbf{C}^{-1}_{(d)} \gets$ updated matrices after adding the cut $c_d$
  		\State $e_d \gets$ error on the validation set from $\mathbf{Z}_{(d)}, \mathbf{C}^{-1}_{(d)}$
  	\EndFor
  	\State $d_{\text{move}} \gets \argmin_{1 \leq d \leq D} e_d$
  	\State $\mathbf{Z} \gets \mathbf{Z}_{(d_{\text{move}})}$, $\mathbf{C}^{-1} \gets \mathbf{C}^{-1}_{(d_{\text{move}})}$
  \end{algorithmic}
\end{algorithm}

\begin{figure}[H]
  \begin{minipage}[c]{0.5\textwidth}
    \includegraphics[width=\textwidth]{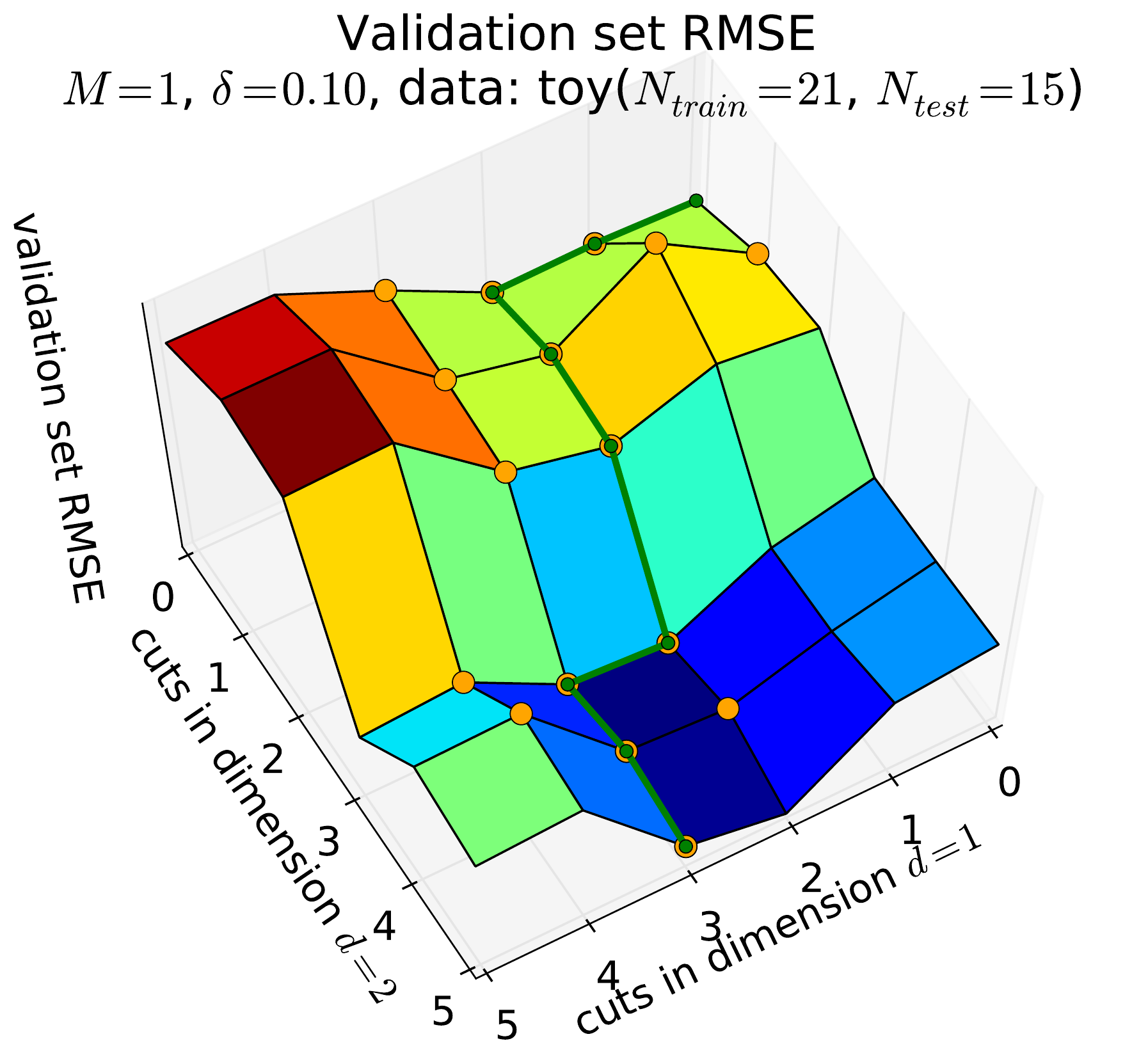}
  \end{minipage}\hfill
  \begin{minipage}[c]{0.49\textwidth}
    \includegraphics[width=\textwidth]{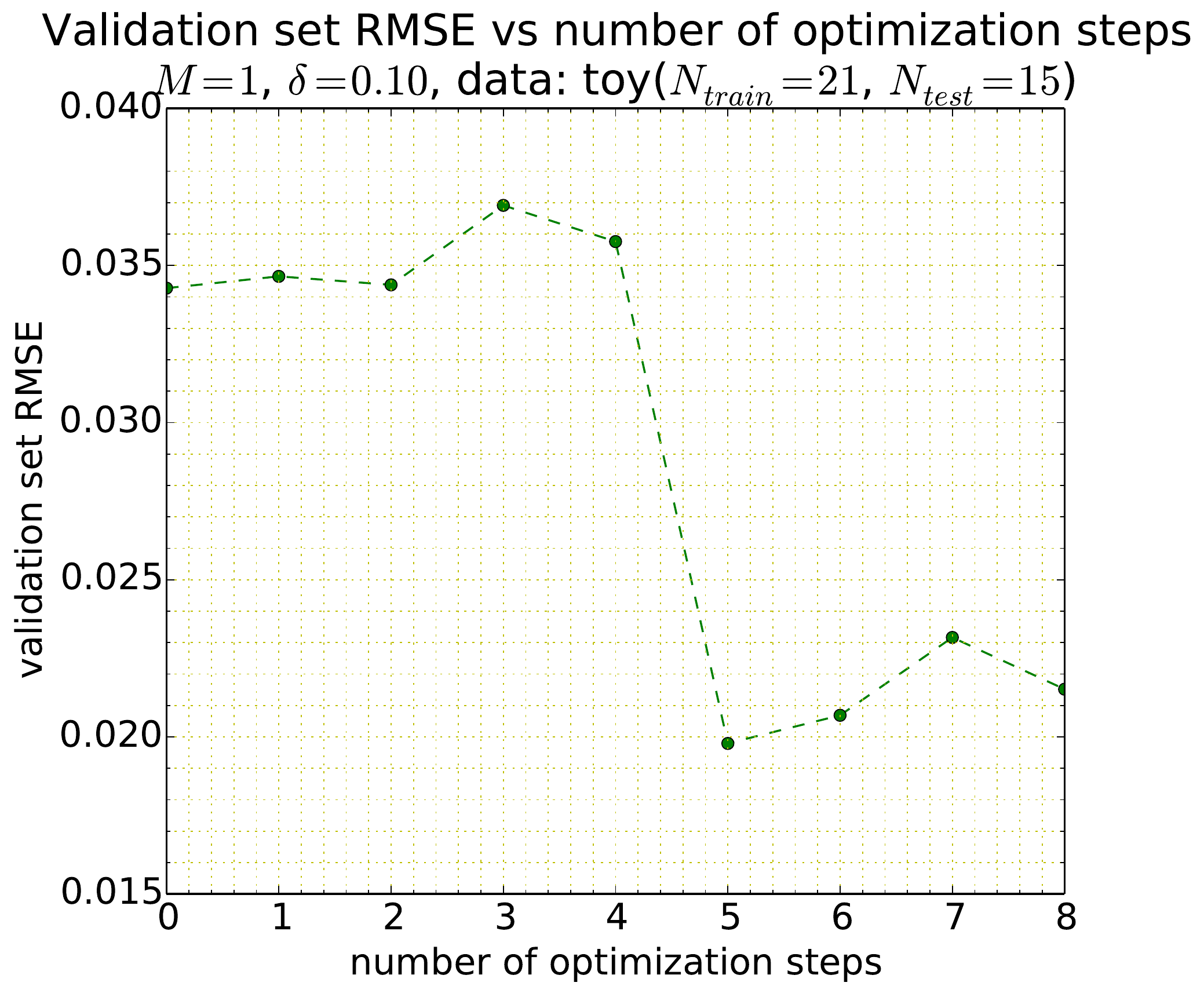}
  \end{minipage}
  \caption{Greedy optimization procedure on a toy 2D dataset, where all lifetime configurations can be evaluated quickly. Nodes of the surface on the left show the validation set RMSE after a given number of cuts has been added in each dimension. The nodes are joined together into a continuous surface for clarity, but note that in fact the RMSE surface is piecewise constant with jumps only when a cut is added/removed. The green line in the left figure shows the path taken by the greedy optimization procedure, started from the origin $\bs{\lambda} = (0, 0)$ (with no cuts added in either dimension) in the top right corner. The orange nodes show the lifetime configurations considered by the optimization procedure for its next step. On this particular toy dataset the greedy optimizer manages to discover the global minimum. The plot on the right plots the height of the green optimization path (validation set RMSE) as a function of the number of optimization steps performed.}
\end{figure}

\begin{figure}[H]
  \begin{minipage}[c]{0.5\textwidth}
    \includegraphics[width=\textwidth]{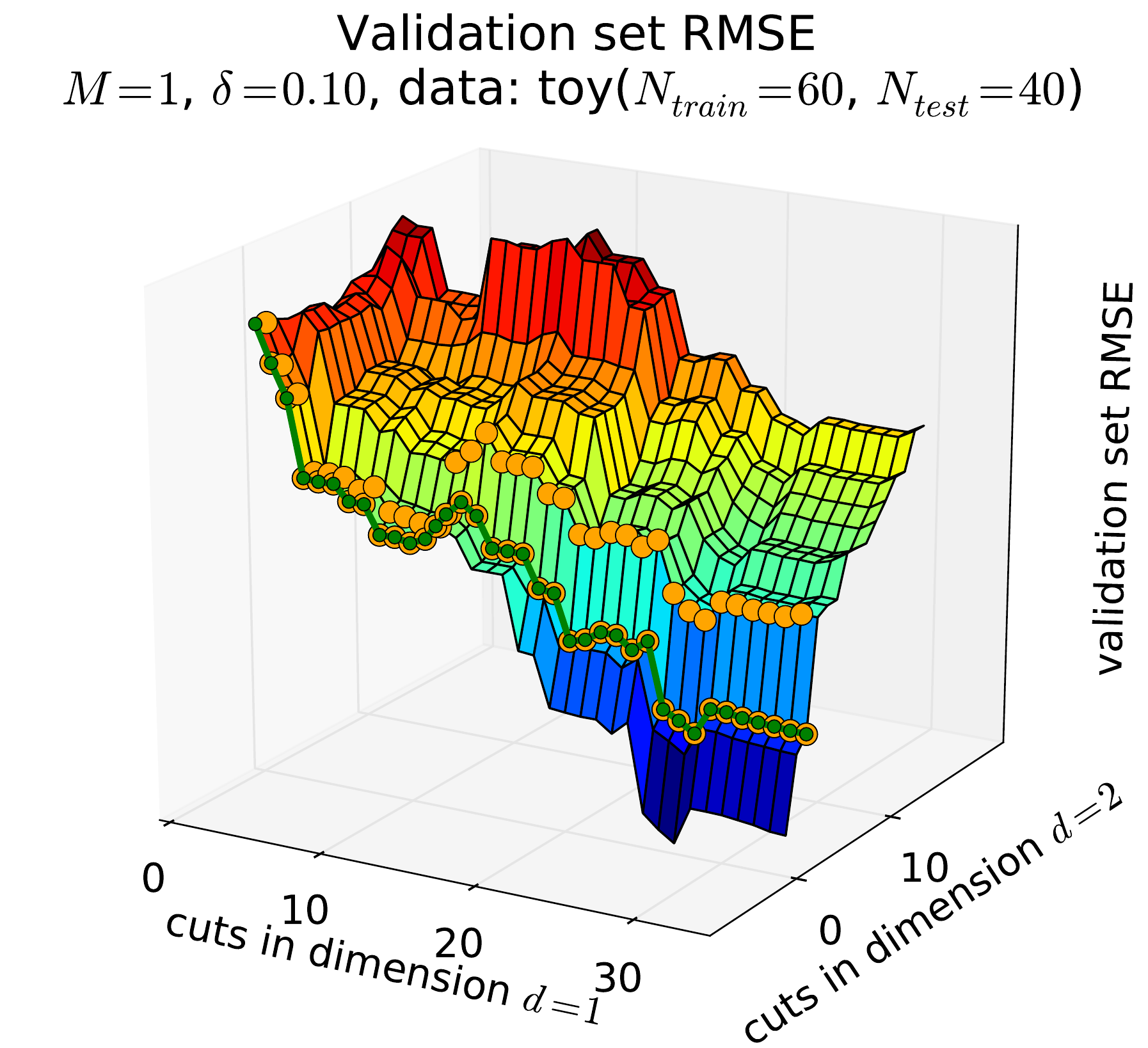}
  \end{minipage}\hfill
  \begin{minipage}[c]{0.49\textwidth}
    \includegraphics[width=\textwidth]{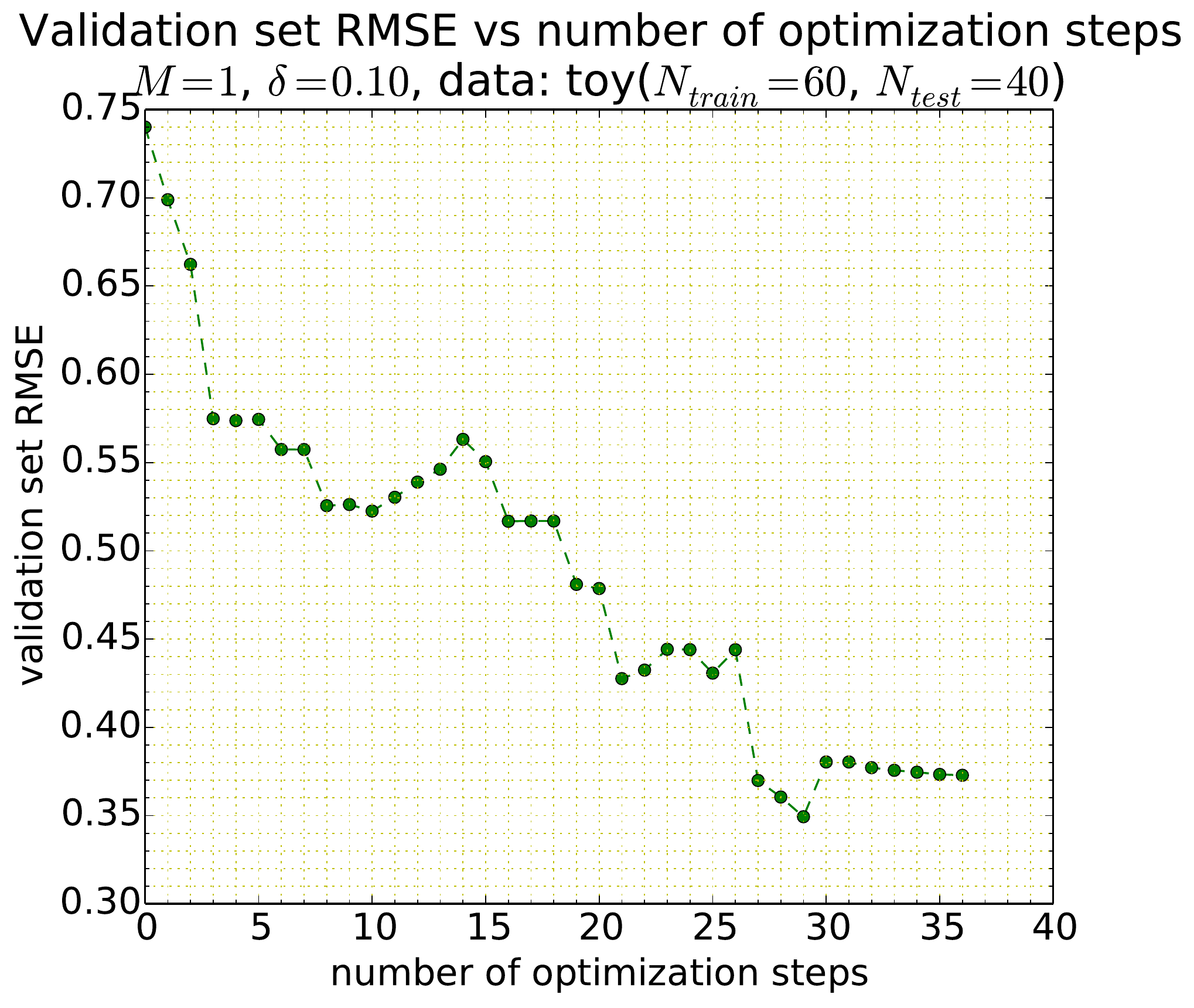}
  \end{minipage}
  \caption{Another toy dataset with two input dimensions, but this time the second dimension $d = 2$ is irrelevant for predicting the target value. The greedy optimization procedure recognizes this and increases the lifetime predominantly in the relevant dimension $d = 1$, as indicated by the green path on the left-hand figure. However, due to randomness in the data, occasionally it also increases the lifetime in the irrelevant dimension. As this greedy optimization procedure only increases lifetimes, such increase can never be reverted in the future even it if would lead to lower validation set RMSE.}
  \label{IrrelevantFeatureExperimentToy}
\end{figure}

The toy experiment shown in Figure~\ref{IrrelevantFeatureExperimentToy} suggests that the Mondrian grid approximator could also be used for basic feature selection. After the optimization procedure discovers a good lifetime configuration $\bs{\lambda} = (\lambda_1, \hdots, \lambda_D)$, a collection of predictive features (input dimensions $d$) can be obtained by selecting those for which $\lambda_d \geq \eps$, where $\eps > 0$ is some small threshold.

\begin{figure}[H]
  \begin{minipage}[c]{0.5\textwidth}
    \includegraphics[width=\textwidth]{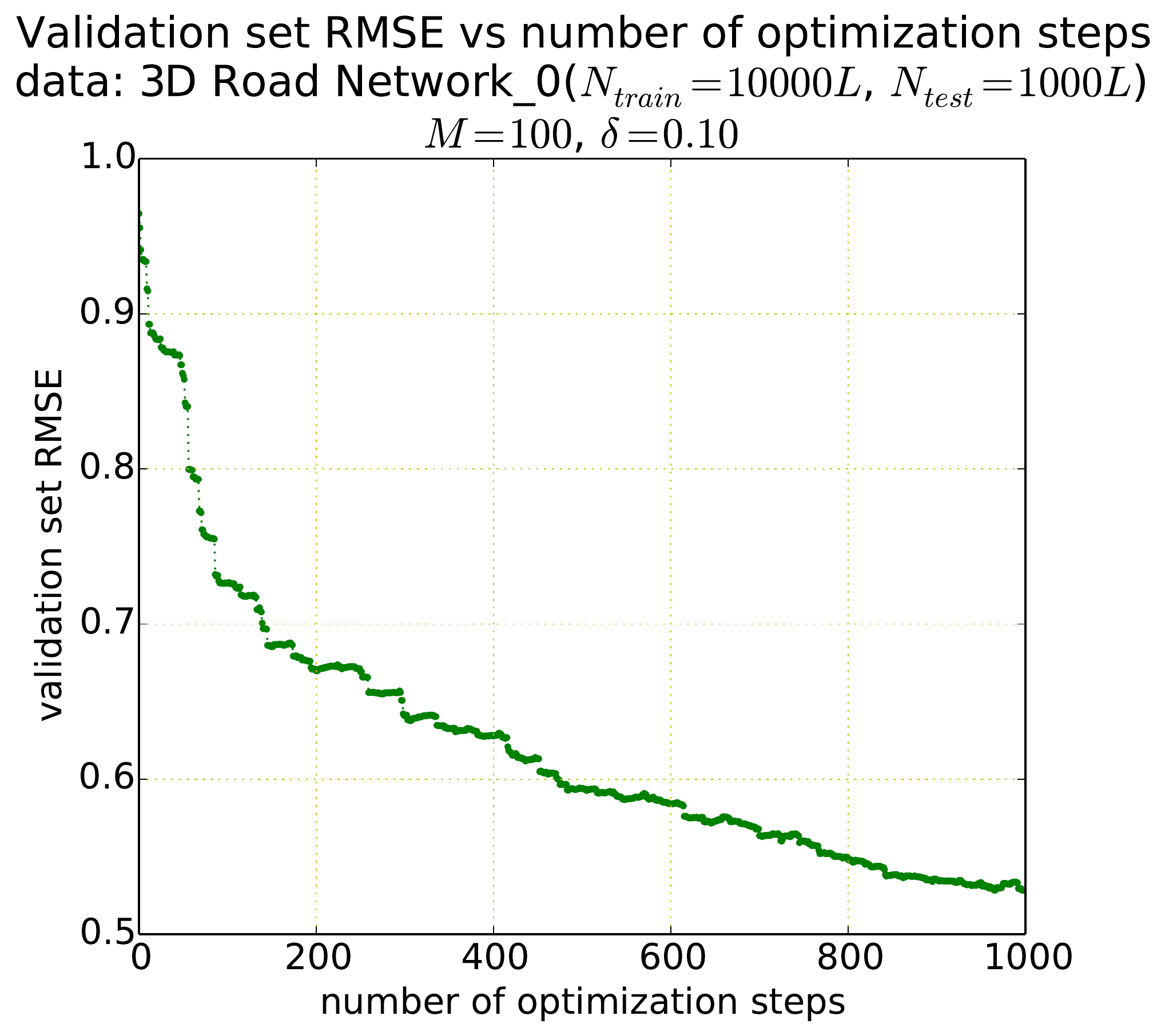}
  \end{minipage}\hfill
  \begin{minipage}[c]{0.49\textwidth}
    \includegraphics[width=\textwidth]{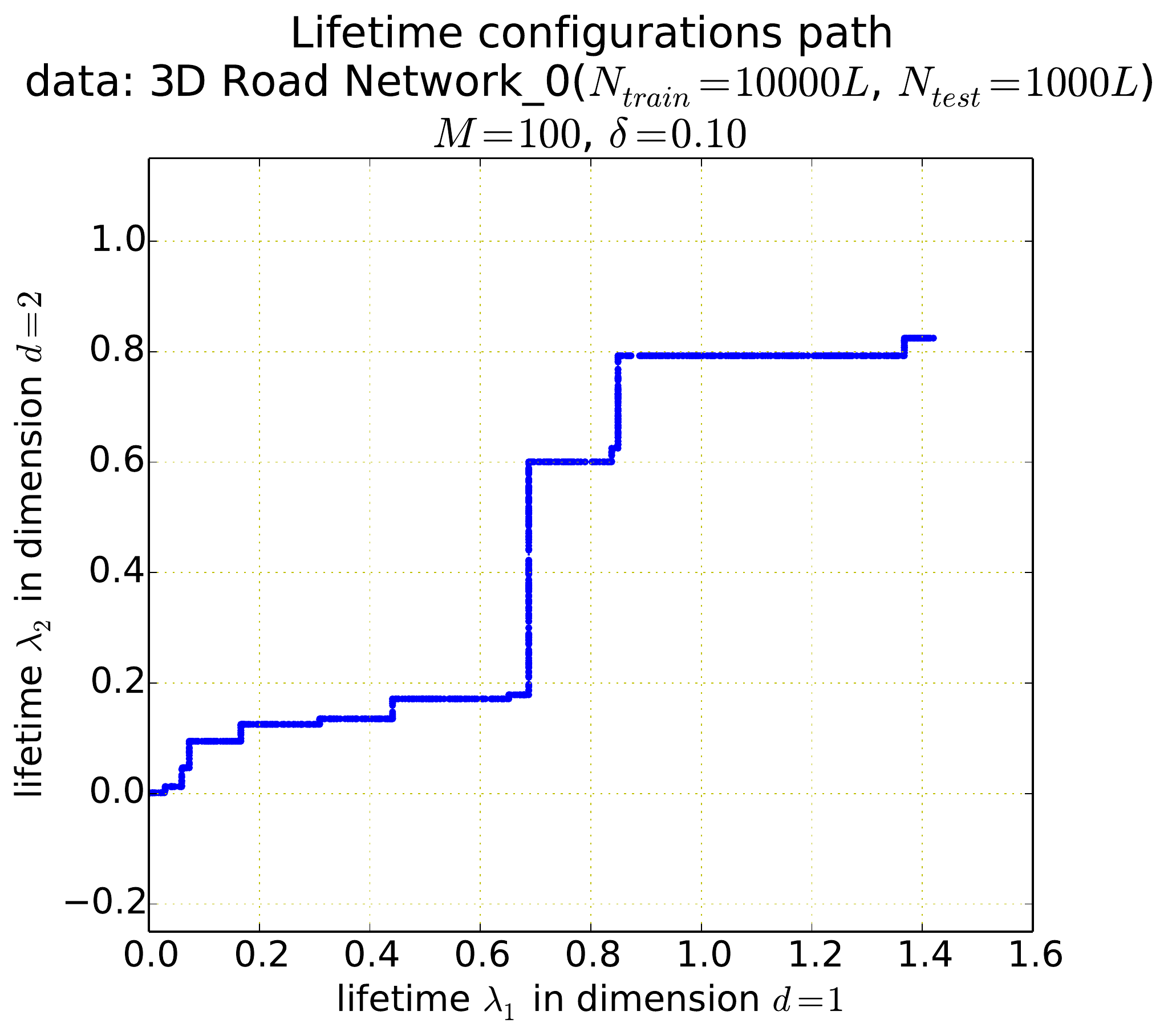}
  \end{minipage}
  \caption{Greedy optimization of Mondrian grid lifetimes on the real-world 3D Road Network dataset \cite{6569130} with $D = 2$ input dimensions. The left-hand plot shows development of validation set RMSE as the optimization progresses and the right-hand plot shows the corresponding evolution of the lifetimes $\lambda_1$, $\lambda_2$ of the two input dimensions. Note that for the training data size $N_{\text{train}} = 10000$ used here the cubic running time of exact computation with the Laplace kernel would be prohibitive, especially if several different lifetime configurations were to be tried out. The  Mondrian grid approximation with our greedy optimization procedure efficiently discovers a good lifetime configuration. (But note there is no guarantee that it is globally optimal among all possible lifetime configurations.)}
\end{figure}

Our greedy optimization procedure described above only considers adding cuts (increasing lifetimes). However, in the previous section we have also explained how the lifetime in one of the dimensions can be efficiently decreased by removing a cut. To see this procedure in action, we have implemented another simple greedy optimization procedure that also considers removing a cut in each dimension individually before deciding which neighboring configuration to explore next.

\begin{figure}[H]
  \begin{minipage}[c]{0.5\textwidth}
    \includegraphics[width=\textwidth]{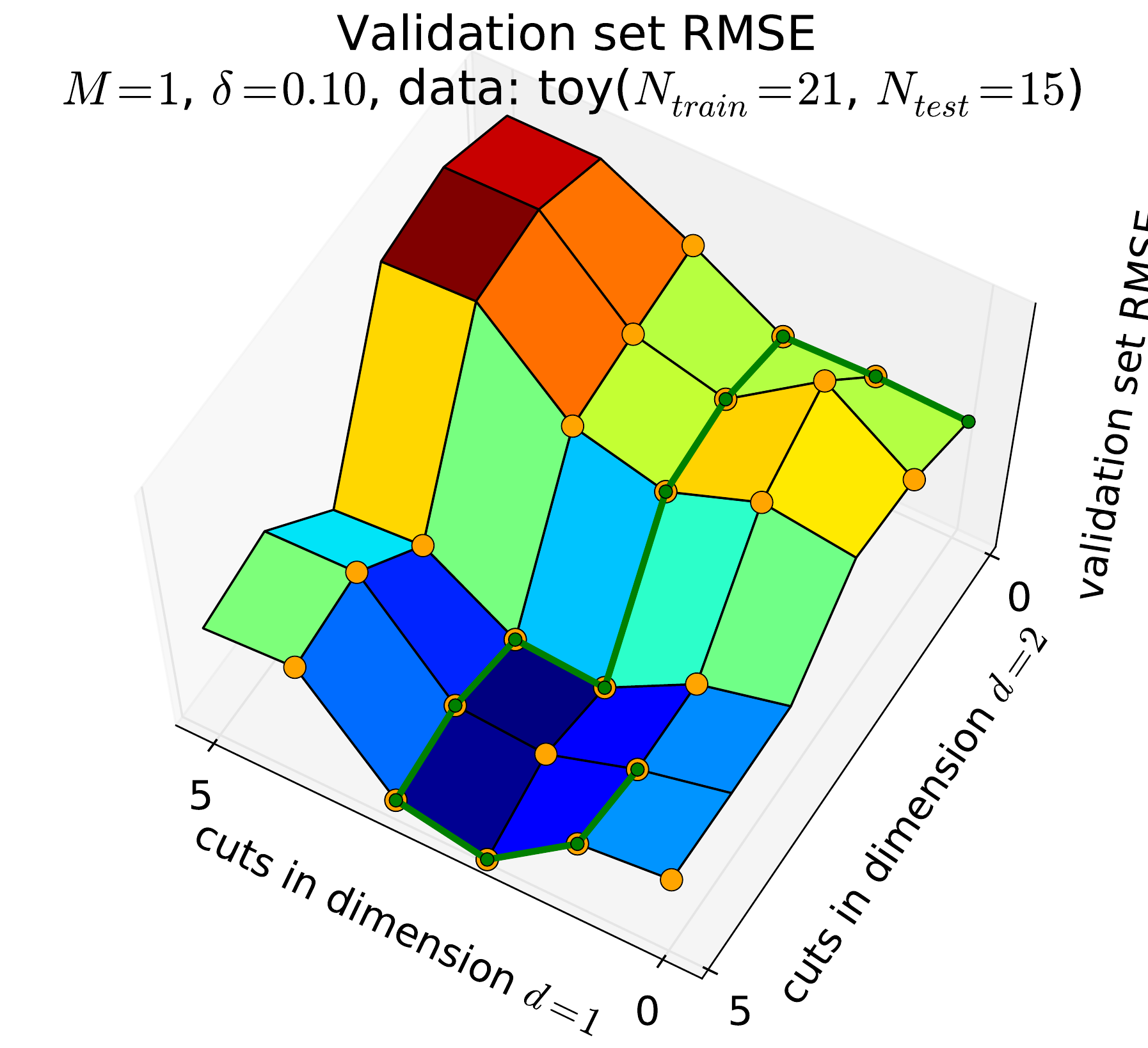}
  \end{minipage}\hfill
  \begin{minipage}[c]{0.49\textwidth}
    \includegraphics[width=\textwidth]{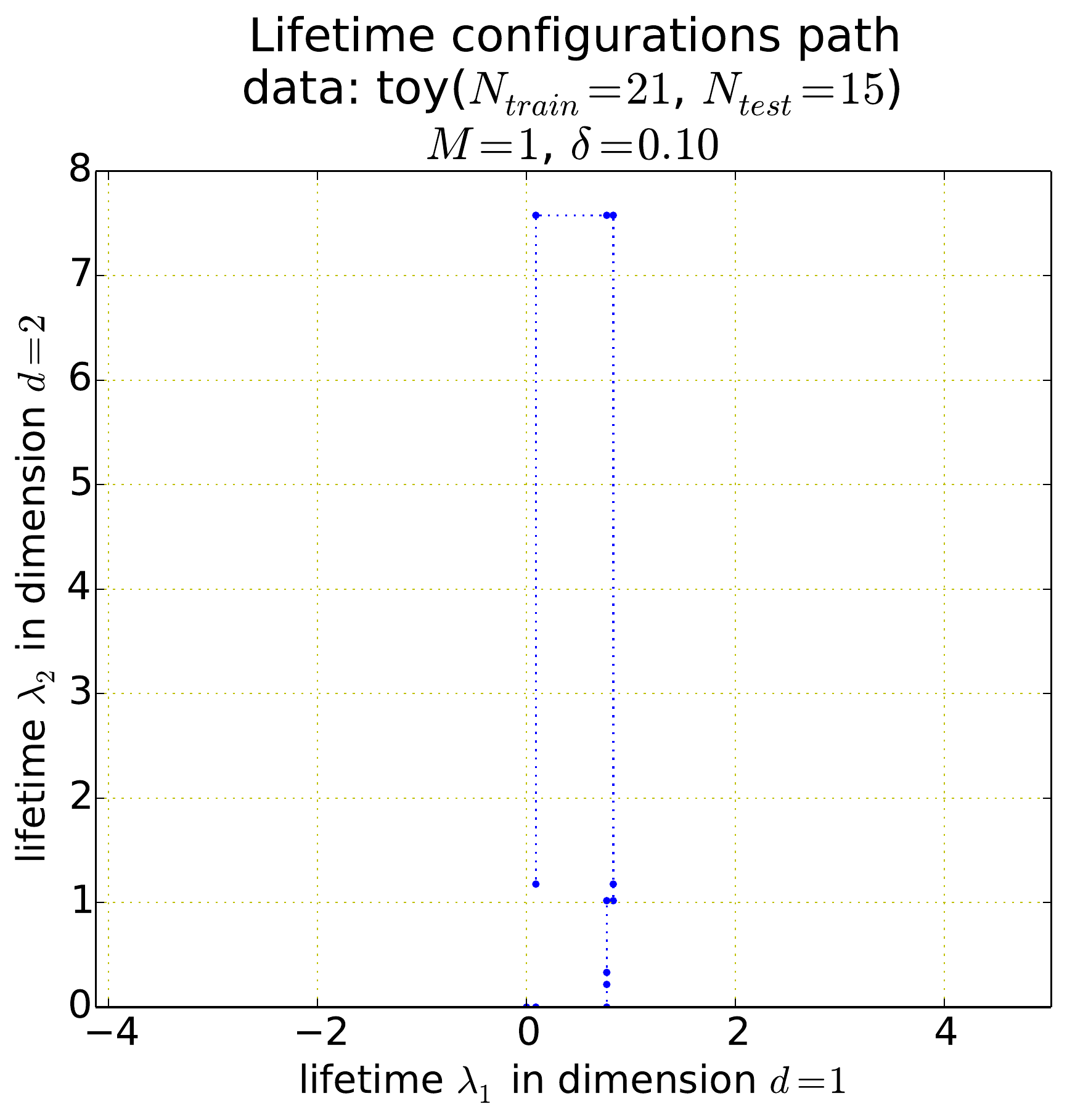}
  \end{minipage}
  \caption{Greedy optimization procedure on a toy 2D dataset, where the procedure is also allowed to remove a cut (decrease the lifetime) if it leads to lower validation RMSE than any cut addition would. A problematic aspect of this greedy optimizer is that it gets easily stuck in local minima.}
\end{figure}

\section{Further work}

There seems to be great room for improvement in the optimization procedure used for deciding which lifetime configuration to explore next. We have implemented two greedy optimizers, one that increases the lifetime in the dimension leading to lowest validation set RMSE and another that also considers decreasing the lifetime. We can easily envision using more sophisticated local optimization algorithms for finding minima of the validation set RMSE as a function of lifetime configuration. For example, instead of looking one step ahead in each dimension, we could compute $K > 1$ steps in each direction before deciding in which dimension to increase the lifetime. Another interesting method to try could be Simultaneous Perturbation Stochastic Approximation (SPSA), which doesn't require access to the gradient of the optimized function and works even in presence of noise in the function measurements. The validation set RMSE as a function of the lifetime configuration is not differentiable and the measurements we get are noisy due to the noise in training and validation datasets.

Another approach we may take is to systematically model the validation RMSE as a random (unknown) function by placing a prior distribution on it (e.g., a Gaussian process), treating the explored lifetime configurations as (noisy) observations of this function and computing the posterior distribution of the function. This posterior could then be used to guide our decision which lifetime configuration to explore next.

To move between different lifetime configurations we have proposed making efficient updates to the matrix $\mathbf{C}^{-1}$, the inverse of the regularized feature covariance matrix $\mathbf{C} = \mathbf{Z}^T \mathbf{Z} + \delta^2 \mathbf{I}_C$. Instead of working with matrix inverses directly it is often suggested for numerical stability reasons to work with the Cholesky decomposition \cite{EPFL-REPORT-161468}. Even though we have not run into numerical issues in our experiments, we outline how the Cholesky decomposition could be used in Appendix D.

\appendix
\part*{Appendix}

\chapter{Selected proofs}
  
\section{Exponential distribution and exponential clocks}

\begin{proposition}
\label{ExpDistributionExpectation}
The expectation of the $\Exp(\lambda)$ distribution is $\frac{1}{\lambda}$.
\begin{proof} Integrating by parts,
\begin{equation*}
\int_{\IR} x p(x | \lambda) \d x
= \int_0^{\infty} \lambda x e^{-\lambda x} \d x
= \left[ \lambda x \frac{1}{- \lambda} e^{-\lambda x} \right]_0^{\infty} - \int_0^{\infty} \lambda \frac{1}{- \lambda} e^{-\lambda x} \d x
= 0 + \left[ \frac{1}{- \lambda} e^{-\lambda x} \right]_0^{\infty}
= \frac{1}{\lambda}
\qedhere
\end{equation*}
\end{proof}
\end{proposition}

\begin{proposition}
\label{ExponentialUniqueMemoryless}
Let $Z$ be any real-valued random variable that is a.s. non-negative and possesses the lack of memory property, i.e.,
\begin{equation*}
\forall{s \geq 0, t \geq 0}
\hs
\IP(Z - t > s \mid Z > t) = \IP(Z > s)
\end{equation*}
Then $Z \sim \Exp(\lambda)$ for some $\lambda > 0$.
\begin{proof} Define $G : [0, \infty) \to [0, 1]$ to be the tail function $G(t) := \IP(Z > t)$. Then $G$ is a decreasing function with $G(0) = 1$ and the assumed lack of memory property gives us the functional equation
\begin{equation*}
G(t + s)
= \IP(Z > t + s)
= \IP(Z - t > s | Z > t) \IP(Z > t)
= \IP(Z > s) \IP(Z > t)
= G(s) G(t)
\end{equation*}
for all $s, t \geq 0$. The rest of the proof is concerned with solving this functional equation.

For $n \in \IN$ we have $G(n t) = G(t + (n - 1) t) = G(t) G( (n - 1) t)$, so by an easy inductive argument we obtain $G(n t) = G(t)^n$. As $t$ is arbitrary here, we can take $t = \frac{1}{n}$ to get $G(1) = G(\frac{1}{n})^n$. Noting that $G$ is a non-negative function, taking the $n$-th root gives $G(\frac{1}{n}) = G(1)^{1/n}$ for all natural $n$.

Suppose $q = \frac{m}{n} \in \IQ$, where $m, n \in \IN$. Then by the already established results
\begin{equation*}
G(q)
= G \left( \frac{m}{n} \right)
= G \left( m \frac{1}{n} \right)
= G \left( \frac{1}{n} \right)^m
= G(1)^{n / m}
= G(1)^q
\end{equation*}
Now suppose $z \in (0, \infty)$. By elementary analysis, we can always find a sequence $(q_n)$ of rational numbers in $(0, z)$ approximating $z$ from below (i.e. $q_n \uparrow z$ as $n \to \infty$). As $G$ is decreasing and limits preserve weak inequalities,
\begin{equation*}
G(z)
\leq \lim_{n \to \infty} G(q_n)
= \lim_{n \to \infty} G(1)^{q_n}
= G(1)^z
\end{equation*}
by continuity of the exponential. Similarly we can consider a sequence of rationals approximating $z$ from above to deduce the opposite inequality $G(z) \geq G(1)^z$. Hence $G(z) = G(1)^z$ for all $z \in (0, \infty)$.

It follows that $G(1) > 0$ (otherwise we'd contradict the assumption $\IP(Z > 0) = 1$) and we can define $\lambda = - \ln G(1) > 0$. Then for all $z \geq 0$ we can write $G(z) = e^{- \mu z}$ and we see that indeed $Z \sim \Exp(\mu)$.
\end{proof}
\end{proposition}

\begin{lemma}[Lack of memory property]
\label{LackOfMemoryPropertyProof}
Let $Z$ be an exponential random variable and $T$ an independent nonnegative random variable. Then $Z$ has the \emph{lack of memory property} at the random time $T$, i.e.
\begin{equation*}
\forall{u}
\hs
\IP(Z - T > u | Z > T) = \IP(Z > u)
\end{equation*}
\begin{proof}
For negative $u$ both sides evaluate to $1$, so in the following we may assume $u \geq 0$.

Let $\lambda$ be the parameter (inverse mean) of the exponential $Z$. For any $a \geq 0$ we have
\begin{IEEEeqnarray*}{rCll}
\IP(Z > R + a) & = &
  \int_0^{\infty} \IP(Z > T + a \mid Z = z) f_Z(z) \d z
  \hs & \text{[ conditioning on $Z$ ]}
\\ & = &
  \int_0^{\infty} \IP(T < z - a) f_Z(z) \d z
  & \text{[ independence of $T$ and $Z$ ]}
\\ & = &
  \int_a^{\infty} \IP(T < z - a) \lambda e^{-\lambda z} \d z
  & \text{[ $T$ is non-negative and $a \geq 0$ ]}
\\ & = &
  \int_0^{\infty} \IP(T < x) \lambda e^{-\lambda (x + a)} \d x
  & \text{[ substitution $x = z - a$]}
\\ & = &
  e^{-\lambda a} \int_0^{\infty} \IP(T < x) f_Z(x) \d x
\end{IEEEeqnarray*}
Using this calculation with $a = u$ and $a = 0$ we get as required,
\begin{equation*}
\IP(Z - T > u \mid Z > T)
= \frac{\IP(Z - T > u, Z > T)}{\IP(Z > T)}
= \frac{\IP(Z - T > u)}{\IP(Z - T > 0)}
= \frac{e^{-\lambda u}}{e^{-\lambda 0}}
= e^{-\lambda u}
= \IP(Z > u)
\qedhere
\end{equation*}
\end{proof}
\end{lemma}

\begin{proposition}
\label{MinExp}
Suppose $X_1, \hdots, X_n$ are independent exponential random variables with rates (inverse means) $\lambda_1, \hdots, \lambda_n$. Then $\min X_i \sim \Exp( \sum \lambda_i )$.
\begin{proof} For any $t \geq 0$ we have by independence of the $X_i$s that
\begin{equation*}
\IP( \min X_i > t )
= \IP \left( \bigcap \{ X_i > t \} \right)
= \prod \IP(X_i > t)
= \prod e^{- \lambda_i t}
= \exp \left( - t \sum \lambda_i \right)
\end{equation*}
and $\IP(\min X_i > t) = 1$ for $t < 0$, so indeed the minimum has the claimed distribution.
\end{proof}
\end{proposition}

The case of two competing exponential clocks is treated by the following theorem. The statement is taken from a problem sheet accompanying my Applied Probability course, the proof is my solution to that question.

\begin{theorem}[Two competing exponential clocks]
Let $X$ and $Y$ be independent exponential random variables (competing exponential alarm clocks) with respective parameters $\lambda$ and $\mu$. Let
\begin{equation*}
W = \min \{ X, Y \}, \hs
Z = \max \{ X, Y \}, \hs
O = Z - W, \hs
M = 1_{ \{X \leq Y \} } =
\begin{dcases}
1 & \text{ if } X \leq Y \\
0 & \text{ if } X > Y
\end{dcases}
\end{equation*}
\begin{itemize}
\item[(a)] Calculate $\IP(W > s)$ and $\IP(M = 1)$. Identify the distributions of $W$ and $M$. Show that the events $\{ W > s \}$ and $\{ M = 1 \}$ are independent.
\item[(b)] Express the event $\{ W \leq w, M = 1, O \leq t \}$ in terms of $X$ and $Y$ and calculate its probability. What is $\IP(W \leq w, M = 0, O \leq t)$? Show that $W$ and $(M, O)$ are independent.
\end{itemize}
\begin{proof}
\textbf{(a)} By Proposition~\ref{MinExp}, $W \sim \Exp(\lambda + \mu)$ and therefore $\IP(W > s) = e^{- (\lambda + \mu) s}$.

Conditioning on the value of $X$ we have
\begin{IEEEeqnarray*}{rCll}
\IP(X \leq Y) & = &
  \int_0^{\infty} \IP(X \leq Y | X = x) f_X(x) \d x
  \hs & \text{[conditioning on the value of $X$]}
\\ & = &
  \int_0^{\infty} \IP(Y \geq x) \lambda e^{-\lambda x} \d x
  & \text{[by independence of $X$ and $Y$]}
\\ & = &
  \frac{\lambda}{\lambda + \mu}
  & \text{[since $\IP(Y \geq x) = e^{-\mu x}$]}
\end{IEEEeqnarray*}
so $\IP(M = 1) = \frac{\lambda}{\lambda + \mu}$ and $\IP(M = 0) = 1 - \IP(M = 1) = \frac{\mu}{\lambda + \mu}$. In other words $M \sim \Ber(\frac{\lambda}{\mu + \lambda})$.

To show independence of the events $\{ W > s \}$ and $\{ M = 1 \}$, we just check that
\begin{IEEEeqnarray*}{rCll}
\IP(W > s, M = 1) & = &
  \int_0^{\infty} \IP(\min \{ X, Y \} > s, X \leq Y | X = x) f_X(x) \d x
  \hs & \text{[conditioning on $X$]}
\\ & = &
  \int_0^s 0 \d x + \int_s^{\infty} \IP(Y \geq x) \lambda e^{-\lambda x} \d x
  & \text{[independence of $X, Y$]}
\\ & = &
  \frac{\lambda}{\lambda + \mu} \left( 1 - e^{-(\lambda + \mu) s} \right)
  & \text{[as $\IP(Y \geq x) = e^{-\mu x}$]}
\\ & = &
  \IP(M = 1) \IP(W > s)
  & \text{[by above]}
\end{IEEEeqnarray*}

\textbf{(b)} By definition of our random variables
\begin{IEEEeqnarray*}{rCll}
E := \IP(W \leq w, M = 1, O \leq t) & = &
  \IP(\min \{ X, Y \} \leq w, X \leq Y, \max \{ X, Y \} - \min \{ X, Y \} \leq t)
\\ & = &
  \IP(X \leq w, X \leq Y, Y - X \leq t)
\end{IEEEeqnarray*}
Conditioning on the value of $X$ (which has to lie in $(0, w)$ if the event $E$ is to occur),
\begin{IEEEeqnarray*}{rCll}
\IP(E) & = &
  \int_0^w \IP(X \leq Y, Y - X \leq t | X = x) f_X(x) \d x
  \hs & \text{[conditioning on the value of $X$]}
\\ & = &
  \int_0^w \IP(x \leq Y \leq x + t) \lambda e^{-\lambda x} \d x
  & \text{[by independence of $X$ and $Y$]}
\\ & = &
  \int_0^w \left( e^{-\mu x} - e^{- \mu (x + t)} \right) \lambda e^{-\lambda x} \d x
  & \text{[as $\IP(Y > a) = e^{-\mu a}$ for $a \geq 0$]}
\\ & = &
  \left( 1 - e^{-\mu t} \right) \frac{\lambda}{\lambda + \mu} \left( 1 - e^{-(\lambda + \mu) w} \right)
\end{IEEEeqnarray*}
Observe that if we repeated the same calculation with the roles of $X$ and $Y$ (and hence of $\lambda$ and $\mu$) swapped, we'd be calculating the probability $\IP( W \leq w, M = 0, O \leq t )$ and the result would be the same except that $\lambda$ and $\mu$ would be swapped). Defining $\alpha(0) = \mu$ and $\alpha(1) = \lambda$, our findings can be expressed compactly as
\begin{equation*}
\IP( W \leq w, M = m, O \leq t )
= \frac{\alpha(m)}{\lambda + \mu} \left( 1 - e^{- \alpha(1 - m) t} \right) \left( 1 - e^{-(\lambda + \mu) w} \right)
\end{equation*}
For any $w, t \geq 0$ and $m \in \{ 0, 1 \}$ we then get (recalling that $W \sim \Exp(\lambda + \mu)$),
\begin{IEEEeqnarray*}{rCl}
\IP(W \in [0, w], (M, O) \in \{ m \} \times [0, t]) & = &
  \IP(W \leq w, M = m, O \leq t)
\\ & = &
  \left[ \frac{\alpha(m)}{\lambda + \mu} \left( 1 - e^{-\alpha(1 - m) t} \right) \right] \left[ 1 - e^{-(\lambda + \mu) w} \right]
\\ & = &
  \IP(W < \infty, M = m, O \leq t) \IP(W \leq w)
\\ & = &
  \IP((M, O) \in \{ m \} \times \{0, t\} ) \IP(W \in [0, w])
\end{IEEEeqnarray*}
As $w, t, m$ were arbitrary, this is sufficient to conclude independence of $W$ and $(M, O)$.
\end{proof}
\end{theorem}

\section{Bayesian Gaussian model}

\begin{proposition}
\label{GaussianPredictionModelProof}
Under the prior $p(\mu) = \mN(\mu | \mu_{\text{prior}}, \sigma_{\text{prior}}^2)$ and likelihood $p(y | \mu) = \mN(y | \mu, \sigma_{\text{noise}}^2)$, the posterior after collecting $N$ independent observations $\mD = \{ y_1, \hdots, y_n \}$ is
\begin{equation*}
p(\mu | \mD)
= \mN\left( \mu \mid \frac{p_{\text{prior}} \mu_{\text{prior}} + p_{\text{noise}} \sum_{n = 1}^N y_n}{p_{\text{prior}} + N p_{\text{noise}}}, ( p_{\text{prior}} + N p_{\text{noise}} )^{-1} \right)
\end{equation*}
where $p_{\text{prior}} = \sigma_{\text{prior}}^{-2}$ and $p_{\text{noise}} = \sigma_{\text{noise}}^{-2}$ are the prior and noise precisions, respectively.
\begin{proof} As the posterior distribution is known to be a probability distribution, it suffices to work up to proportionality ($\propto$) and normalize at the end:
\begin{IEEEeqnarray*}{rCll+x*}
p(\mu | \mD) & = &
  p(\mu) \prod_{n = 1}^N p(y_n | \mu)
\\ & \propto &
  \exp\left( - \frac{(\mu - \mu_{\text{prior}})^2}{2 \sigma_{\text{prior}}^2} - \sum_{n = 1}^N \frac{(y_n - \mu)^2}{2 \sigma_{\text{noise}}^2} \right)
\\ & \propto &
  \exp\left\{ - \frac{1}{2} \left( p_{\text{prior}} \mu^2 - 2 p_{\text{prior}} \mu \mu_{\text{prior}} - 2 p_{\text{noise}} \mu \sum_{n = 1}^N y_n + p_{\text{noise}} N \mu^2 \right) \right\}
\\ & \propto &
  \exp\left\{ - \frac{p_{\text{prior}} + p_{\text{noise}} N}{2} \left(  \mu - \frac{p_{\text{prior}} \mu_{\text{prior}} + p_{\text{noise}} \sum_{n = 1}^N y_n}{p_{\text{prior}} + N p_{\text{noise}}} \right)^2 \right\}
\\ & \propto &
  \mN\left( \mu \mid \frac{p_{\text{prior}} \mu_{\text{prior}} + p_{\text{noise}} \sum_{n = 1}^N y_n}{p_{\text{prior}} + N p_{\text{noise}}}, ( p_{\text{prior}} + N p_{\text{noise}} )^{-1} \right)
  & & \qedhere
\end{IEEEeqnarray*}
\end{proof}
\end{proposition}

\section{Poisson point process}

\begin{lemma}
\label{PoissonPointProcessConditional}
Let $\Pi$ be a Poisson point process on $[a, b]$ with constant intensity $\lambda$. Conditionally given that the process generated $N = n$ points, their locations are i.i.d. uniform in $[a, b]$.
\begin{proof} We follow the proof given in \cite{StirzakerPaRP}. Let $A_1, \hdots, A_K$ be a partition of $[a, b]$ and let $n_1, \hdots, n_K$ be integers with $n_1 + \cdots + n_k = n$. Writing $N(A_k) := | \Pi \cap A_k |$ and $m(A_k)$ for the Lebesgue measure of $A_k$, we have by definition of conditional probability
\begin{IEEEeqnarray*}{rCll+x*}
\IP\left( \bigcap_{k = 1}^K \{ N(A_k) = n_k \} \mid N = n \right)
  & = &
  \frac{\IP\left( \bigcap_{k = 1}^K \{ N(A_k) = n_k \} \right)}{\IP(N = n)}
\\ & = &
  \frac{\prod_{k = 1}^K \IP( N(A_k) = n_k )}{{\IP(N = n)}}
  & \text{[ (ii) in Definition~\ref{PoissonPointProcessDefinition} ]}
\\ & = &
  \frac{\prod_{k = 1}^K e^{-\lambda m(A_k)} \frac{(\lambda m(A_k))^{n_k}}{n_k !}}{e^{- \lambda m([a, b])} \frac{\lambda m([a, b])^n}{n!}}
  & \text{[ (i) in Definition~\ref{PoissonPointProcessDefinition} ]}
\\ & = &
  \binom{n}{n_1, \hdots, n_K} \prod_{k = 1}^K \left( \frac{m(A_k)}{m([a, b])} \right)^{n_k}
  \hs & \text{[ } \sum_{k = 1}^K m(A_k) = m([a, b]) \text{ ]}
\end{IEEEeqnarray*}
We recognize this as the multinomial distribution with unnormalized parameters $m(A_1), \hdots, m(A_K)$. This distribution can be represented as each point being independently assigned to region $A_k$ with probability $\frac{m(A_k)}{m([a, b])}$, so the $n$ points are (conditionally) independent. As the partition $(A_1, \hdots, A_k)$ was arbitrary, the distribution of the point locations is uniform.
\end{proof}
\end{lemma}

\section{Conditional Mondrians}

\begin{lemma}
\label{ConditionalMondriansCase1Proof}
Suppose we are conditionally given that the restriction $M^{\Phi}$ of a Mondrian process $M \sim \text{MP}(\lambda, \Theta)$ with lifetime $\lambda \in [0, \infty)$ to a smaller box $\Phi \subseteq \Theta$ is trivial (contains no cuts). Then
\begin{itemize}
\item[$\bullet$] with probability $\exp( \lambda (\text{LD}(\Theta) - \text{LD}(\Phi))$, $M$ is also trivial
\item[$\bullet$] with complementary probability $1 - \exp( \lambda (\text{LD}(\Theta) - \text{LD}(\Phi))$ the first cut in $\Theta$ misses $\Phi$, its time has the truncated exponential distribution with rate $\text{LD}(\Theta) - \text{LD}(\Phi)$ and truncation at $\lambda$, and the cut location is uniformly distributed along the segments where making a cut doesn't hit $\Phi$.
\end{itemize}
\begin{proof}
Let $T$ be the time of the first cut of $M$. By Bayes' rule, its conditional distribution is
\begin{equation*}
p( T = t \mid M^{\Phi} = \emptyset )
= \frac{p( T = t)}{p(M^{\Phi} = \emptyset)} p( M^{\Phi} = \emptyset \mid T = t )
\end{equation*}
By definition of the Mondrian process $T \sim \Exp( \text{LD}(\Theta) )$ and $p(M^{\Phi} = \emptyset) = p( \text{MP}(\lambda, \Phi) = \emptyset ) = e^{- \lambda \text{LD}(\Phi)}$ using self-consistency. Finally, the probability that $M^{\Phi}$ is empty given that the first cut in $\Theta$ occurs at time $t$ can be obtained as
\begin{IEEEeqnarray*}{rCll+x*}
p( M^{\Phi} = \emptyset \mid T = t ) & = &
  p( \text{first cut misses } \Phi \mid T = t )
  p( M^{\Phi} = \emptyset \mid T = t, \text{first cut misses } \Phi )
\\ & = &
  \frac{\text{LD}(\Theta) - \text{LD}(\Phi)}{\text{LD}(\Theta)}
  p( \text{MP}(\lambda - t, \Phi) = \emptyset )
\\ & = &
  \frac{\text{LD}(\Theta) - \text{LD}(\Phi)}{\text{LD}(\Theta)}
  e^{- (\lambda - t) \text{LD}(\Phi)}
\end{IEEEeqnarray*}
where the second equality again uses self-consistency. Plugging into the Bayes' formula
\begin{IEEEeqnarray*}{rCll+x*}
p( T = t \mid M^{\Phi} = \emptyset ) & = &
  \frac{\text{LD}(\Theta) \exp( - \text{LD}(\Theta) t)}{\exp( - \lambda \text{LD}(\Phi))} \frac{\text{LD}(\Theta) - \text{LD}(\Phi)}{\text{LD}(\Theta)}
  e^{- (\lambda - t) \text{LD}(\Phi)}
\\ & = &
  \left( \text{LD}(\Theta) - \text{LD}(\Phi) \right) \exp\left( -  (\text{LD}(\Theta) - \text{LD}(\Phi)) t \right)
\end{IEEEeqnarray*}
We see that $T \mid (M^{\Phi} = \emptyset) \sim \text{Exp}( \text{LD}(\Theta) - \text{LD}(\Phi) )$. The probability that this time is within the lifetime $\lambda$ of the Mondrian is $1 - \exp( \lambda (\text{LD}(\Theta) - \text{LD}(\Phi))$ and conditionally on being within the lifetime, the distribution becomes truncated at $\lambda$.

For brevity of notation, define the event $A := \{ \text{first cut of } M \text{ occurs outside } \Phi \text{ at time } T = t \}$. The conditional distribution of the location $X$ of the first cut is, using Bayes' formula,
\begin{equation*}
p( X = x \mid M^{\Phi} = \emptyset, A )
= \frac{p( X = x \mid A)}{p(M^{\Phi} = \emptyset \mid A)} p( M^{\Phi} = \emptyset \mid X = x, A)
\end{equation*}
Given that the first cut occurs outside $\Phi$, the density of its location is $p( X = x \mid A) = (\text{LD}(\Theta) - \text{LD}(\Phi))^{-1}$. By self-consistency $p( M^{\Phi} = \emptyset \mid X = x, A)$ does not depend on the value of $x$ and therefore this probability equals the denominator $p(M^{\Phi} = \emptyset \mid A)$. Hence
\begin{equation*}
p( X = x \mid M^{\Phi} = \emptyset, A )
= p( X = x \mid A)
= \frac{1}{\text{LD}(\Theta) - \text{LD}(\Phi)}
\end{equation*}
We see that as advertised, the conditional distribution of the cut location is uniform (among cut locations that don't split $\Phi$).
\end{proof}
\end{lemma}

\section{Ridge regression}

\begin{theorem}
\label{RidgeRegressionIsMAP}
MAP parameter estimation of $\bs{\theta}$ in the linear model
\begin{IEEEeqnarray*}{rCll+x*}
\bs{\theta} & \sim &
  \mN(\bs{0}, \sigma_{\text{prior}}^2 \mathbf{I}_D)
  \hs &
\\
y & = &
  \bs{\theta}^T \mathsf{x} + \eps
  & \hs \text{where } \eps \sim \mN(0, \sigma_{\text{noise}}^2)
\end{IEEEeqnarray*}
is equivalent to $L_2$-regularized least-squares, i.e., to minimizing the function
\begin{equation*}
f(\bs{\theta})
:= \delta^2 \| \bs{\theta} \|_2^2 + \sum_{n = 1}^N (y_n - \mathsf{x}_n^T \bs{\theta})^2
\end{equation*}
where $\delta := \frac{\sigma_{\text{noise}}}{\sigma_{\text{prior}}}$.
\begin{proof} The prior on $\bs{\theta}$ can be written as
\begin{equation*}
p(\bs{\theta})
= \mN(\bs{0}, \sigma_{\text{prior}}^2 \mathbf{I}_D)
= \frac{1}{(2 \pi)^{\frac{D}{2}} | \sigma_{\text{prior}}^2 \mathbf{I}_D |^{\frac{1}{2}}}
  \exp\left( - \frac{1}{2} \bs{\theta}^T (\sigma_{\text{prior}}^2 \mathbf{I}_D)^{-1} \bs{\theta} \right)
= \frac{1}{(2 \pi \sigma_{\text{prior}}^2)^{\frac{D}{2}}} \exp\left( - \frac{\| \bs{\theta} \|_2^2}{2 \sigma_{\text{prior}}^2} \right)
\end{equation*}
By independence of noise in different observations, the likelihood function of the observed data $\mD$ as a function of the parameter $\bs{\theta}$ factorizes as
\begin{equation*}
\mL(\bs{\theta} | \mD)
= p(\mD | \bs{\theta})
= \prod_{n = 1}^N p(y_n | \mathsf{x}_n, \bs{\theta})
= \prod_{n = 1}^N \mN( y_n \mid \mathsf{x}_n^T \bs{\theta}, \sigma_{\text{noise}}^2 )
= \prod_{n = 1}^N \frac{1}{\sqrt{2 \pi \sigma_{\text{noise}}^2}} \exp\left( - \frac{(y_n - \mathsf{x}_n^T \bs{\theta})^2}{2 \sigma_{\text{noise}}^2} \right)
\end{equation*}
The posterior distribution $p( \bs{\theta} | \mD)$ is proportional to the product of the prior $p( \theta )$ and the likelihood $\mL(\bs{\theta} | \mD)$. The MAP estimate of $\bs{\theta}$ is obtained by maximizing this posterior, which is equivalent to minimizing its negative likelihood:
\begin{equation*}
- \ln p( \bs{\theta} | \mD )
= - \ln p( \bs{\theta} ) - \ln \mL(\bs{\theta} | \mD) + \text{const}
= \frac{\| \bs{\theta} \|_2^2}{2 \sigma_{\text{prior}}^2}
  + \sum_{n = 1}^N \frac{(y_n - \mathsf{x}_n^T \bs{\theta})^2}{2 \sigma_{\text{noise}}^2}
  + \text{const}
\end{equation*}
Multiplying this equation by the positive quantity $2 \sigma_{\text{noise}}^2 > 0$ and defining $\delta := \frac{\sigma_{\text{noise}}}{\sigma_{\text{prior}}}$, we can equivalently minimize the following function of $\bs{\theta}$:
\begin{equation*}
f(\bs{\theta})
:= \delta^2 \| \bs{\theta} \|_2^2 + \sum_{n = 1}^N (y_n - \mathsf{x}_n^T \bs{\theta})^2
\end{equation*}
As advertised, this is the standard least squares minimization problem with an $L_2$ regularization term.
\end{proof}
\end{theorem}

\begin{theorem}
\label{RidgeRegressionSolutionDD}
For $\delta > 0$, the $L_2$ regularized least squares objective function
\begin{equation*}
f(\bs{\theta})
= \delta^2 \| \bs{\theta} \|_2^2 + \| \mathbf{y} - \mathbf{X} \bs{\theta} \|_2^2
\end{equation*}
has a unique minimum at $\bs{\theta} = (\mathbf{X}^T \mathbf{X} + \delta^2 \mathbf{I}_D)^{-1} \mathbf{X}^T \mathbf{y}$.
\begin{proof}
Expanding the definition of $f(\bs{\theta})$ we obtain
\begin{IEEEeqnarray*}{rCll+x*}
f(\bs{\theta}) & = &
  \delta^2 \bs{\theta}^T \bs{\theta} + (\mathbf{y} - \mathbf{X} \bs{\theta})^T (\mathbf{y} - \mathbf{X} \bs{\theta})
\\ & = &
  \delta^2 \bs{\theta}^T \bs{\theta} + \mathbf{y}^T \mathbf{y} - \bs{\theta}^T \mathbf{X}^T \mathbf{y} - \mathbf{y}^T \mathbf{X} \bs{\theta} + \bs{\theta}^T \mathbf{X}^T \mathbf{X} \bs{\theta}
  \hs & \text{[ distributivity ]}
\\ & = &
  \text{const} + \delta^2 \bs{\theta}^T \bs{\theta} - 2 \mathbf{y}^T \mathbf{X} \bs{\theta} + \bs{\theta}^T \mathbf{X}^T \mathbf{X} \bs{\theta}
  & \text{[ $\bs{\theta}^T \mathbf{X}^T \mathbf{y}$ and $\mathbf{y}^T \mathbf{X} \bs{\theta}$ are scalars ]}
\end{IEEEeqnarray*}
Having expressed $f(\bs{\theta})$ in terms of matrices, we now recall from matrix calculus that if $\mathbf{A}$ is a matrix constant with respect to $\bs{\theta}$ then $\frac{\partial \mathbf{A} \bs{\theta}}{\partial \bs{\theta}} = \mathbf{A}^T$ and $\frac{\partial \bs{\theta}^T \mathbf{A} \bs{\theta}}{\partial \bs{\theta}} = 2 \mathbf{A}^T \bs{\theta}$. Therefore
\begin{equation*}
\frac{\partial f(\bs{\theta})}{\partial \bs{\theta}}
= 2 \delta^2 \bs{\theta} - 2 \mathbf{X}^T \mathbf{y} + 2 \mathbf{X}^T \mathbf{X} \bs{\theta}
\end{equation*}
Setting these partial derivatives (the gradient of $f$) to $\mathbf{0}$ yields the so-called \textbf{normal equation}
\begin{equation*}
(\mathbf{X}^T \mathbf{X} + \delta^2 \mathbf{I}_D) \bs{\theta} = \mathbf{X}^T \mathbf{y}
\end{equation*}
Observe that for any $\mathbf{v} \in \IR^D \setminus \{ \mathbf{0} \}$ we have $\mathbf{v}^T (\mathbf{X}^T \mathbf{X} + \delta^2 \mathbf{I}_D) \mathbf{v} = \| \mathbf{X} \mathbf{v} \|_2^2 + \delta^2 \| \mathbf{v} \|_2^2 > 0$, so the matrix $(\mathbf{X}^T \mathbf{X} + \delta^2 \mathbf{I}_D)$ is positive definite, therefore all its eigenvalues are positive and hence it is invertible. Thus we can pre-multiply the normal equation by the inverse of this matrix to obtain an explicit form for the location of this stationary point:
\begin{equation*}
\bs{\theta}
= (\mathbf{X}^T \mathbf{X} + \delta^2 \mathbf{I}_D)^{-1} \mathbf{X}^T \mathbf{y}
\end{equation*}
That this value of $\bs{\theta}$ is a minimum of $f$ can be confirmed by computing the Hessian matrix of second derivatives $\mathbf{H} = 2 \delta^2 \mathbf{I}_D + 2 \mathbf{X}^T \mathbf{X}$, which is easily seen to be positive definite. Hence $f$ is strictly convex, implying that the critical point of $f$ found above must indeed be a unique global minimum.
\end{proof}
\end{theorem}

\begin{theorem}
\label{RidgeRegressionSolutionNN}
The MAP estimate $\bs{\theta}^{\text{MAP}}$ of the ridge regression parameter can be also expressed as
\begin{equation*}
\bs{\theta}^{\text{MAP}}
= \mathbf{X}^T (\mathbf{X} \mathbf{X}^T + \delta^2 \mathbf{I}_N)^{-1} \mathbf{y}
\end{equation*}
\begin{proof} For the right-hand expression, we start by rewriting the normal equation from the proof of Theorem~\ref{RidgeRegressionSolutionDD} as $\mathbf{X}^T \mathbf{X} \bs{\theta} + \delta^2 \bs{\theta} = \mathbf{X}^T \mathbf{y}$; rearranging and dividing by the positive scalar $\delta^2 > 0$ then yields
\begin{equation*}
\bs{\theta}
= \delta^{-2} \left( \mathbf{X}^T \mathbf{y} - \mathbf{X}^T \mathbf{X} \bs{\theta} \right)
= \mathbf{X}^T \delta^{-2} \left( \mathbf{y} - \mathbf{X} \bs{\theta} \right)
\end{equation*}
This still involves $\bs{\theta}$ on the right-hand side, but we can obtain an expression for $\mathbf{X} \bs{\theta}$ from the normal equation: pre-multiplying it by $\mathbf{X}$ gives
\begin{equation*}
\mathbf{X} \mathbf{X}^T \mathbf{y}
= \mathbf{X} (\mathbf{X}^T \mathbf{X} + \delta^2 \mathbf{I}_D) \bs{\theta}
= \mathbf{X} \mathbf{X}^T \mathbf{X} \bs{\theta} + \delta^2 \mathbf{X} \bs{\theta}
= ( \mathbf{X} \mathbf{X}^T + \delta^2 \mathbf{I}_N ) \mathbf{X} \bs{\theta}
\end{equation*}
Arguing similarly as before, the matrix $\mathbf{X} \mathbf{X}^T + \delta^2 \mathbf{I}_N$ is positive definite and therefore invertible for any $\delta > 0$, so we may write
\begin{IEEEeqnarray*}{rCl+x*}
\bs{\theta} & = &
  \mathbf{X}^T \delta^{-2} \left( \mathbf{y} - \mathbf{X} \bs{\theta} \right)
\\ & = &
  \mathbf{X}^T \delta^{-2} \left[ \mathbf{y} - ( \mathbf{X} \mathbf{X}^T + \delta^2 \mathbf{I}_N )^{-1} \mathbf{X} \mathbf{X}^T \mathbf{y} \right]
\\ & = &
  \mathbf{X}^T \delta^{-2} \left[ \mathbf{y} - ( \mathbf{X} \mathbf{X}^T + \delta^2 \mathbf{I}_N)^{-1} ( \mathbf{X} \mathbf{X}^T + \delta^2 \mathbf{I}_N) \mathbf{y} + \delta^2 ( \mathbf{X} \mathbf{X}^T + \delta^2 \mathbf{I}_N)^{-1} \mathbf{y} \right]
\\ & = &
  \mathbf{X}^T ( \mathbf{X} \mathbf{X}^T + \delta^2 \mathbf{I}_N)^{-1} \mathbf{y}
\end{IEEEeqnarray*}
Note that we've used an "add and subtract trick" to obtain the third equality.
\end{proof}
\end{theorem}

\section{Updating matrix inverses}

We start with a lemma showing how the inverse of a matrix $\mathbf{A}$ changes when the last row and column of $\mathbf{A}$ are deleted. A more general case is deduced afterwards.

\begin{lemma}
\label{SpecialSubmatrixInverse}
Let $\mathbf{A} \in \IR^{n \times n}$ be invertible with inverse of the block form
\begin{equation*}
\mathbf{A}^{-1}
= \begin{bmatrix}
  \mathbf{E} & \mathbf{f} \\
  \mathbf{g}^T & h
  \end{bmatrix}
\end{equation*}
where $\mathbf{E} \in \IR^{(n - 1) \times (n - 1)}$, $\mathbf{f} \in \IR^{n - 1}$, $\mathbf{g} \in \IR^{n - 1}$ and $h \in \IR$. Let $\tilde{\mathbf{A}}$ be the matrix obtained from $\mathbf{A}$ by deleting its last row and column. If $h \not= 0$ then $\tilde{\mathbf{A}}$ is invertible with inverse $\tilde{\mathbf{A}}^{-1} = \mathbf{E} - \mathbf{f} \mathbf{g}^T / h$.
\begin{proof} Write $\mathbf{A} = \begin{bmatrix} \tilde{\mathbf{A}} & \mathbf{b} \\ \mathbf{c}^T & d \end{bmatrix}$. As $\mathbf{A} \mathbf{A}^{-1} = \mathbf{I}_n$, by properties of matrix multiplication
\begin{IEEEeqnarray*}{rCl}
\tilde{\mathbf{A}} \mathbf{E} + \mathbf{b} \mathbf{g}^T & = &
  \mathbf{I}_{n - 1}
\\
\tilde{\mathbf{A}} \mathbf{f} + \mathbf{b} h & = &
  \mathbf{0}
\end{IEEEeqnarray*}
Dividing the second equality by the scalar $h \not= 0$ and solving for $\mathbf{b}$ gives $\mathbf{b} = - \tilde{\mathbf{A}} \mathbf{f} / h$. Plugging this into the first equality yields
\begin{equation*}
\mathbf{I}_{n - 1}
= \tilde{\mathbf{A}} \mathbf{E} - \tilde{\mathbf{A}} \mathbf{f} \mathbf{g}^T / h
= \tilde{\mathbf{A}} \left( \mathbf{E} - \mathbf{f} \mathbf{g}^T / h \right)
\end{equation*}
This proves that $\tilde{\mathbf{A}}$ is invertible and its inverse has the stated form.
\end{proof}
\end{lemma}

Now we deduce a more general result, where it is the $i$-th row and $i$-th column of $\mathbf{A}$ that are deleted.

\begin{definition} Let $\sigma$ be a permutation of $\{ 1, 2, \hdots, n \}$. The \emph{permutation matrix} $\mathbf{P}^{\sigma} \in \IR^{n \times n}$ is the monomial matrix with $(i, j)$ entry equal to $\II( \sigma(i) = j )$.
\end{definition}

Observe that for $\mathbf{A} \in \IR^{n \times n}$, pre-multiplication $\mathbf{P}^{\sigma} \mathbf{A}$ permutes the rows of $\mathbf{A}$ using $\sigma^{-1}$, while post-multiplication $\mathbf{A} \mathbf{P}^{\sigma}$ permutes the columns of $\mathbf{A}$ using $\sigma$. Note also that $(\mathbf{P}^{\sigma})^{-1} = (\mathbf{P}^{\sigma})^T = \mathbf{P}^{\sigma^{-1}}$.

\begin{lemma}[Inverse of a submatrix]
\label{SubmatrixInverseProof}
Let $\mathbf{A} \in \IR^{n \times n}$ be invertible, let $1 \leq i \leq n$ and let $\tilde{\mathbf{A}}$ be the matrix obtained from $\mathbf{A}$ by deleting its $i$-th row and $i$-th column. Let $\mathbf{E}$ be the submatrix of $\mathbf{A}^{-1}$ obtained by deleting its $i$-th row and column, let $\mathbf{f}$ be the $i$-th column of $\mathbf{A}^{-1}$ with the $i$-th entry removed, let $\mathbf{g}$ be the $i$-th row of $\mathbf{A}^{-1}$ with the $i$-th entry removed, and finally let $h$ be the $(i, i)$ entry of $\mathbf{A}^{-1}$.
\\
If $h \not= 0$ then $\tilde{\mathbf{A}}$ is invertible and its inverse is $\tilde{\mathbf{A}}^{-1} = \mathbf{E} - \mathbf{f} \mathbf{g}^T / h$.
\begin{proof} Using the cycle notation, define the permutation $\sigma = (i \; i + 1 \; \cdots \; n)$. Note that the inverse of this permutation $\sigma^{-1}$ sends $i$ to $n$. Let $\phi : \IR^{n \times n} \to \IR^{(n - 1) \times (n - 1)}$ be the function that deletes the last row and last column of a matrix, and let $\psi$ be the function that updates its inverse accordingly (as dictaded by Lemma~\ref{SpecialSubmatrixInverse}), i.e. $\psi( \mathbf{X}^{-1} ) = \phi(\mathbf{X})^{-1}$. Note that the matrix $\tilde{\mathbf{A}}$ can be equivalently obtained from $\mathbf{A}$ by sending the $i$-th row and column to the last positions and then applying the function $\phi$, so that $\tilde{\mathbf{A}} = \phi(\mathbf{P}^{\sigma} \mathbf{A} \mathbf{P}^{\sigma^{-1}})$. Thus
\begin{equation*}
\tilde{\mathbf{A}}^{-1}
= \phi(\mathbf{P}^{\sigma} \mathbf{A} \mathbf{P}^{\sigma^{-1}})^{-1}
= \psi\left( ( \mathbf{P}^{\sigma} \mathbf{A} \mathbf{P}^{\sigma^{-1}} )^{-1} \right)
= \psi\left( \mathbf{P}^{\sigma} \mathbf{A}^{-1} \mathbf{P}^{\sigma^{-1}} \right)
\end{equation*}
The right-hand side tells us that to compute $\tilde{\mathbf{A}}^{-1}$, we may move the $i$-th row and column of $\mathbf{A}$ to the last positions and then apply the procedure $\psi$ from Lemma~\ref{SpecialSubmatrixInverse}. But that yields precisely that $\tilde{\mathbf{A}}^{-1} = \mathbf{E} - \mathbf{f} \mathbf{g}^T / h$ when $h \not= 0$.
\end{proof}
\end{lemma}

The next lemma goes in the reverse direction, showing how the inverse changes when a new row and column is appended to the original matrix.

\begin{lemma}[Inverse of an extended matrix]
\label{SupermatrixInverseProof}
Let $\mathbf{A} \in \IR^{n \times n}$ be invertible. For $\mathbf{b} \in \IR^n$, $\mathbf{c} \in \IR^n$ and $d \in \IR$, the extended matrix
\begin{equation*}
  \begin{bmatrix}
  \mathbf{A} & \mathbf{b} \\
  \mathbf{c}^T & d
  \end{bmatrix}
\end{equation*}
is invertible if and only if its Schur complement $s := d - \mathbf{c}^T \mathbf{A}^{-1} \mathbf{b} \not= 0$, in which case the inverse is
\begin{equation*}
  \begin{bmatrix}
  \mathbf{A} & \mathbf{b} \\
  \mathbf{c}^T & d
  \end{bmatrix}^{-1}
= \begin{bmatrix}
  \mathbf{E} & \mathbf{f} \\
  \mathbf{g}^T & h
  \end{bmatrix}
  \hs\text{where}\hs
  \begin{array}{rclrcl}
  \mathbf{E} & = & \mathbf{A}^{-1} + s^{-1} \mathbf{A}^{-1} \mathbf{b} \mathbf{c}^T \mathbf{A}^{-1} &
  \hs \mathbf{f} & = & - s^{-1} \mathbf{A}^{-1} \mathbf{b} \\
  \mathbf{g} & = & - s^{-1} \mathbf{c}^T \mathbf{A}^{-1} &
  h & = & s^{-1}
  \end{array}
\end{equation*}
This inverse can be computed from $\mathbf{A}^{-1}$ in time $\mO(n^2)$.
\begin{proof} First suppose that $s \not = 0$. The stated forms for $\mathbf{E}$, $\mathbf{f}$, $\mathbf{g}$, $h$ could be derived from the equations
\begin{equation*}
\begin{bmatrix} \mathbf{A} & \mathbf{b} \\ \mathbf{c}^T & d \end{bmatrix} \begin{bmatrix} \mathbf{E} & \mathbf{f} \\ \mathbf{g}^T & h \end{bmatrix} = \mathbf{I}_{n + 1}
\hs\text{and}\hs
\begin{bmatrix} \mathbf{E} & \mathbf{f} \\ \mathbf{g}^T & h \end{bmatrix} \begin{bmatrix} \mathbf{A} & \mathbf{b} \\ \mathbf{c}^T & d \end{bmatrix} = \mathbf{I}_{n + 1}
\end{equation*}
but given that we have the forms stated, it suffices to verify that they do solve one of these two equations (which then implies that the other is also satisfied). Indeed, for the first equation we have
\begin{IEEEeqnarray*}{rCll+x*}
  & &  
  \begin{bmatrix}
  \mathbf{A} & \mathbf{b} \\
  \mathbf{c}^T & d
  \end{bmatrix}
  \begin{bmatrix}
  \mathbf{A}^{-1} + s^{-1} \mathbf{A}^{-1} \mathbf{b} \mathbf{c}^T \mathbf{A}^{-1} & - s^{-1} \mathbf{A}^{-1} \mathbf{b} \\
  - s^{-1} \mathbf{c}^T \mathbf{A}^{-1} & s^{-1}
  \end{bmatrix}
\\ & = &
  \begin{bmatrix}
  \mathbf{I}_n + s^{-1} \mathbf{b} \mathbf{c}^T \mathbf{A}^{-1} - \mathbf{b} s^{-1} \mathbf{c}^T \mathbf{A}^{-1} &
  - s^{-1} \mathbf{b} + \mathbf{b} s^{-1} \\
  \mathbf{c}^T \mathbf{A}^{-1} + \mathbf{c}^T s^{-1} \mathbf{A}^{-1} \mathbf{b} \mathbf{c}^T \mathbf{A}^{-1} - d s^{-1} \mathbf{c}^T \mathbf{A}^{-1} &
  - \mathbf{c}^T s^{-1} \mathbf{A}^{-1} \mathbf{b} + d s^{-1}
  \end{bmatrix}
\\ & = &
  \begin{bmatrix}
  \mathbf{I}_n & \mathbf{0} \\
  \mathbf{0}^T & 1
  \end{bmatrix}
= \mathbf{I}_{n + 1}
\end{IEEEeqnarray*}
where the definition of the Schur complement $s$ has been used to simplify the two terms in the bottom row to get the last row.

Conversely, if the inverse matrix exists, then $\mathbf{A} \mathbf{f} + \mathbf{b} h = \mathbf{0}$ and $\mathbf{c}^T \mathbf{f} + d h = 1$. From the first equality we obtain $\mathbf{f} = - \mathbf{A}^{-1} \mathbf{b} h$ and plugging this into the second gives $- \mathbf{c}^T \mathbf{A}^{-1} \mathbf{b} h + dh = 1$. Thus $s = - \mathbf{c}^T \mathbf{A}^{-1} \mathbf{b} + d \not= 0$, as required to prove the reverse direction of the claim.

Finally, suppose that $\mathbf{A}^{-1}$, $\mathbf{b}$, $\mathbf{c}$ and $d$ are known. Then $s = d - \mathbf{c}^T (\mathbf{A}^{-1} \mathbf{b})$ can be computed in time $\mO(n^2)$ and then $h = s^{-1}$ is immediate. Also, $\mathbf{f} = - h (\mathbf{A}^{-1} b)$, then $\mathbf{g} = - h (\mathbf{c}^T \mathbf{A}^{-1})$ and finally $\mathbf{E} = \mathbf{A}^{-1} - \mathbf{f} (\mathbf{c}^T \mathbf{A}^{-1})$ can all be computed in $\mO(n^2)$ time.
\end{proof}
\end{lemma}

\chapter{Model interpretation}

In this report we have considered several different models for non-linear regression: Mondrian random forest, kernel ridge regression and a Laplace kernel approximation using the Mondrian process. In this chapter we show that under certain conditions all three models are so-called linear smoothers and hint at what the fundamental difference between these models is.

\begin{definition} Let $\mD = \{ (\mathsf{x}_1, y_1), \hdots, (\mathsf{x}_N, y_N) \}$ be a training dataset. A predictor $\hat{y}$ for a new test point $\mathsf{x}_{*}$ is called a \textbf{linear smoother} if it can be expressed as a linear combination of the training responses, i.e.,
\begin{equation*}
\hat{y}
= \sum_{n = 1}^N k_e(\mathsf{x}_{*}, \mathsf{x}_n, \mathbf{X}) y_n
\end{equation*}
The coefficients $k_e(\mathsf{x}_{*}, \mathsf{x}_n, \mathbf{X})$ are allowed to depend on the locations of all the training datapoints, but not on their response values $y_n$. The function $k_e$ is called the \textbf{smoothing matrix} or the \textbf{equivalent kernel} \cite{Bishop:2006:PRM:1162264}.
\end{definition}

\begin{proposition} Mondrian random forest regression is a linear smoother if and only if the prior predictive distribution in the leaves $\mN( \mu_{\text{prior}}, \sigma_{\text{prior}}^2 )$ has $\mu_{\text{prior}} = 0$ or $\sigma_{\text{prior}}^2 = \infty$ (no prior).
\begin{proof}
Let $p_{\text{prior}} := \sigma_{\text{prior}}^{-2}$ be the prior precision and let $p_{\text{noise}}$ be the observation noise. For $1 \leq m \leq M$, let $l_m(\mathsf{x})$ be the function that returns the leaf of the $m$-th Mondrian tree associated with the partition cell into which points $\mathsf{x}$ falls. The prediction $\hat{y}$ at a point $\mathsf{x}_{*}$ can be expressed as
\begin{equation*}
\hat{y}
= \frac{1}{M} \sum_{m = 1}^M \frac{p_{\text{prior}} \mu_{\text{prior}} + p_{\text{noise}} \sum_{n = 1}^N \II( l_m(\mathsf{x}_{*}) = l_m(\mathsf{x}_n) ) y_n}{p_{\text{prior}} + p_{\text{noise}} \sum_{k = 1}^N \II( l_m(\mathsf{x}_{*}) = l_m(\mathsf{x}_k) )}
\end{equation*}
We see that the constant term not involving $y_n$ disappears if and only if $\mu_{\text{prior}} = 0$ or $p_{\text{prior}} = 0$ (no prior). In that case the prediction is
\begin{equation*}
\hat{y}
= \sum_{n = 1}^N y_n \frac{1}{M} \sum_{m = 1}^M \frac{p_{\text{noise}}  \II( l_m(\mathsf{x}_{*}) = l_m(\mathsf{x}_n) )}{p_{\text{prior}} + p_{\text{noise}} \sum_{k = 1}^N \II( l_m(\mathsf{x}_{*}) = l_m(\mathsf{x}_k) )}
\end{equation*}
revealing a linear smoother with smoothing matrix
\begin{equation*}
k^{\text{MF}}_M(\mathsf{x}_{*}, \mathsf{x}_n, X)
= \frac{1}{M} \sum_{m = 1}^M \frac{\II( l_m(\mathsf{x}_{*}) = l_m(\mathsf{x}_n) )}{\frac{p_{\text{prior}}}{p_{\text{noise}}} + \sum_{k = 1}^N \II( l_m(\mathsf{x}_{*}) = l_m(\mathsf{x}_k) )}
\qedhere
\end{equation*}
\end{proof}
\end{proposition}

\begin{remark}
The $M$ Mondrian trees of a Mondrian forest are sampled independently, so by the law of large numbers, as $M \to \infty$, the obtained smoothing matrix converges at the standard rate to
\begin{equation*}
k^{\text{MF}}(\mathsf{x}_{*}, \mathsf{x}_n, X)
= \IE\left[ \frac{\II( l_m(\mathsf{x}_{*}) = l_m(\mathsf{x}_n) )}{\frac{p_{\text{prior}}}{p_{\text{noise}}} + \sum_{k = 1}^N \II( l_m(\mathsf{x}_{*}) = l_m(\mathsf{x}_k) )} \right]
\end{equation*}
Ignoring the prior (setting $p_{\text{prior}} = 0$), this quantity can be described as the expected proportion of the $n$-th datapoint in the leaf containing the prediction point $\mathsf{x}_{*}$. As we would expect, this is a quantity that
\begin{itemize}
\item[•] is increased when the distance between $\mathsf{x}_{*}$ and $\mathsf{x}_n$ is small
\item[•] is decreased when there is a large number of training point in the vicinity of $\mathsf{x}_{*}$
\end{itemize}
\end{remark}

\pagebreak

\begin{proposition} Kernel ridge regression is a linear smoother for any valid kernel function $k$.
\begin{proof} Expanding the prediction $\hat{y}$ at a test point $\mathsf{x}_{*}$ by writing $\mathbf{k}(\mathsf{x}_{*}, X)$ for the row vector of length $N$ with $n$-th entry equal to $k(\mathsf{x}_{*}, \mathsf{x}_n)$,
\begin{IEEEeqnarray*}{rCll+x*}
\hat{y} & = &
  \mathsf{x}_{*} \bs{\theta}
\\ & = &
  \mathbf{k}(\mathsf{x}_{*}, \mathbf{X}) ( \mathbf{K} + \delta^2 \mathbf{I}_N)^{-1} \mathbf{y}
\\ & = &
  \sum_{n = 1}^N y_n \sum_{i = 1}^N \mathbf{k}(\mathsf{x}_{*}, \mathsf{x}_i) \left(( \mathbf{K} + \delta^2 \mathbf{I}_N)^{-1}\right)_{in}
\end{IEEEeqnarray*}
We see that kernel ridge regression is a linear smoother with smoothing matrix
\begin{equation*}
k_e(\mathsf{x}_{*}, \mathsf{x}_n, \mathbf{X})
= \sum_{i = 1}^N \mathbf{k}(\mathsf{x}_{*}, \mathsf{x}_i) \left( ( \mathbf{K} + \delta^2 \mathbf{I}_N)^{-1}\right)_{in}
\qedhere
\end{equation*}
\end{proof}
\end{proposition}

\begin{remark}
Interpreting the smoothing matrix of kernel ridge regression is made difficult by the presence of the matrix inverse $( \mathbf{K} + \delta^2 \mathbf{I}_N)^{-1}$.
\end{remark}

\section{Mondrian forest vs Laplace kernel approximation}

We have presented two models for non-linear regression that utilize Mondrian trees: the Mondrian forest regression model and the Mondrian approximation of the Laplace kernel. In both models we independently sample $M$ Mondrian trees with finite lifetime $\lambda \geq 0$ and use them to obtain $M$ independent partitions of the datapoints $\mathbf{X}$ at hand.

However, the two models are also clearly different in some way. With Mondrian forest regression, the prediction at a point $\mathsf{x}_{*}$ is made by computing the prediction from each of the $M$ trees independently and then simply averaging them. There is no interaction between the different Mondrian trees. On the other hand, with the Mondrian approximation of the Laplace kernel, we solve a ridge regression problem to find a parameter vector $\bs{\theta}$. There can be non-trivial dependences between all entries of this vector, including those corresponding to features produced by different Mondrian trees. Thus in this model, we can have interaction between the different Mondrian trees.

\subsubsection{The case M=1}

In this subsection we show that under a slight condition, the two regression models coincide in the case $M = 1$. We will write $N(i)$ for the number of training datapoints falling into the $i$-th leaf of the single Mondrian tree that is sampled.

Let us consider the Laplace kernel approximation model first. Recall that the ridge regression solution can be expressed as
\begin{equation*}
\bs{\theta} = (\mathbf{Z}^T \mathbf{Z} + \delta^2 \mathbf{I}_C)^{-1} \mathbf{Z}^T \mathbf{y}
\end{equation*}
where $\mathbf{Z} \in \IR^{N \times C}$ is the training data feature matrix, $\mathbf{y}$ are the training data targets and $C$ is the number of random features created. The prediction at a new test point $\mathsf{z}_{*}$ is given by $\hat{y} = \bs{\theta}^T \mathsf{z}_{*}$. In the case $M = 1$ we have that
\begin{itemize}
\item[•] $\mathbf{Z} \in \IR^{N \times C}$ is a binary matrix, with the $n$-th row containing precisely one non-zero entry $z_{nc}$, indicating that the $n$-th datapoint falls into leaf $c$
\item[•] $\mathbf{C} = \mathbf{Z}^T \mathbf{Z} + \delta^2 \mathbf{I}_C \in \IR^{C \times C}$ has $(i, j)$-entry $c_{ij}$ equal to, for $i \not= j$,
\begin{equation*}
c_{ij}
= \left( \mathbf{Z}^T \mathbf{Z} \right)_{ij} + \delta^2 0
= \sum_{n = 1}^N z_{ni} z_{nj}
= \sum_{n = 1}^N 0
= 0
\end{equation*}
and for $i = j$,
\begin{equation*}
c_{ii}
= \left( \mathbf{Z}^T \mathbf{Z} \right)_{ii} + \delta^2 1
= \sum_{n = 1}^N z_{ni} z_{ni} + \delta^2
= \sum_{n = 1}^N z_{ni} + \delta^2
= N(i) + \delta^2
\end{equation*}
Thus $\mathbf{C}$ is a diagonal matrix, with $i$-th diagonal entry being the number of datapoints in the $i$-th leaf, plus the $\delta^2$ regularization term.
\item[•] $\mathbf{C}^{-1} \in \IR^{C \times C}$ as an inverse of a diagonal matrix is also diagonal, with $i$-th diagonal entry
\begin{equation*}
c_{ii}^{-1}
= \frac{1}{N(i) + \delta^2}
\end{equation*}
\item[•] $\mathbf{Z}^T \mathbf{y} \in \IR^C$ has $i$-th entry $\alpha_i$ equal to
\begin{equation*}
\alpha_i
= \sum_{n = 1}^N z_{in} y_n
\end{equation*}
This is the sum of target values $y_n$ corresponding to datapoints that fall into leaf $i$.
\item[•] $\bs{\theta} = \mathbf{C}^{-1} (\mathbf{Z}^T \mathbf{y}) \in \IR^C$ has $i$-th entry
\begin{equation*}
\theta_i
= c_{ii}^{-1} \alpha_i
= \frac{1}{N(i) + \delta^2} \sum_{n = 1}^N z_{in} y_n
\end{equation*}
With $\delta = 0$ this would be the average target value among datapoints in leaf $i$; with $\delta > 0$ this can be interpreted as having an additional observation of $0$ with weight $\delta^2$.
\item[•] $\hat{y} = \bs{\theta}^T \mathsf{z}_{*} \in \IR$ can, by letting $\tilde{c}$ be the leaf of the Mondrian tree into which $\mathsf{z}_{*}$ falls, be expressed as
\begin{equation*}
\hat{y}
= \bs{\theta}^T \mathsf{z}_{*}
= \sum_{c = C} \theta_c z_{*c}
= \theta_{\tilde{c}}
= \frac{1}{N(\tilde{c}) + \delta^2} \sum_{n = 1}^N z_{(\tilde{c})n} y_n
\end{equation*}
Hence the prediction $\hat{y}$ is simply the training average of targets in the leaf of the prediction location, regularized by a $\delta^2$-weighted prior observation of $0$.
\end{itemize}

This prediction $\hat{y}$ is the same as the one made by the Mondrian forest regression model with a single tree, provided that its hyperparameters are chosen compatibly with $\delta^2$: the prior predictive distribution $\mN(0, \sigma_{\text{prior}}^2)$ must be zero-centered and together with the observation noise $\mN(0, \sigma_{\text{noise}}^2)$ they must satisfy the relation
\begin{equation*}
\frac{\sigma_{\text{noise}}^2}{\sigma_{\text{prior}}^2} = \delta^2
\end{equation*}

This confirms that in the case $M = 1$, the Laplace kernel approximation approach coincides with the Mondrian forest regression model with a zero-centered predictive prior and suitable hyperparameters. Since we've shown that the former computes a random feature space in which the inner products have expected values equal to the Laplace kernel, our model equivalence implies that the same is true of Mondrian forest regression (with $M = 1$), which can therefore also be though of as an approximator for the Laplace kernel.

\subsubsection{The general case}

In Mondrian forest regression, we sample $M$ Mondrians in order to decrease the variance of the predictions directly, by averaging the final predictions from $M$ trees. This increases the probability that the \emph{predictions} will be closer to their expectation.

In Laplace kernel approximation using Mondrian trees, we sample the $M$ Mondrians in order to decrease the variance of the kernel approximation. This increases the probability that \emph{inner products} in the random feature space will better approximate the Laplace kernel.

Another way of thinking about the similarities and differences between these two models can be gained by interpreting them as linear smoothers.

\subsubsection{Linear smoothers}

We've already seen that Mondrian forest regression is a linear smoother for any value of $M$, provided that the predictive prior has mean zero. We've also seen that the kernel ridge regression model (with any valid kernel) is also a linear smoother. Very similarly, it turns out that the Mondrian approximation of the Laplace kernel is also a linear smoother, for any value of $M$.

\pagebreak

\begin{proposition} The Laplace kernel approximation using $M$ Mondrian trees is a linear smoother.
\begin{proof} Expanding the formula for the prediction $\hat{y}$ at a new test point $\mathsf{z}_{*}$,
\begin{IEEEeqnarray*}{rCll+x*}
\hat{y} & = &
  \mathsf{z}_{*}^T \bs{\theta}
\\ & = &
  \mathsf{z}_{*}^T (\mathbf{Z}^T \mathbf{Z} + \delta^2 \mathbf{I}_C)^{-1} \mathbf{Z}^T \mathbf{y}
\\ & = &
  \sum_{n = 1}^N y_n \sum_{c = 1}^C z_{*c} \sum_{k = 1}^C (\mathbf{Z}^T \mathbf{Z} + \delta^2 \mathbf{I}_C)^{-1}_{ck} \mathbf{Z}^T_{kn}
\\ & = &
  \sum_{n = 1}^N y_n \sum_{c = 1}^C z_{*c} \sum_{k = 1}^C (\mathbf{Z}^T \mathbf{Z} + \delta^2 \mathbf{I}_C)^{-1}_{ck} z_{nk}
  & & \qedhere
\end{IEEEeqnarray*}
\end{proof}
\end{proposition}

We can read off the smoothing matrix $k_e(\mathsf{z}_{*}, \mathsf{z}_n, \mathbf{Z}) = \sum_{c = 1}^C z_{*c} \sum_{k = 1}^C (\mathbf{Z}^T \mathbf{Z} + \delta^2 \mathbf{I}_C)^{-1}_{ck} z_{nk}$, but the presence of the matrix inverse makes if difficult to analyse it in the general case (we've seen that for $M = 1$ the inverted matrix is diagonal).

Now we can summarize the difference between Mondrian forest regression and Laplace kernel approximation as follows. In Laplace kernel approximation we use $M$ independent Mondrian samples to approximate the \emph{Laplace} kernel. As the prediction is not a linear function of the kernel (it involves a matrix inverse), the prediction doesn't decompose with the $M$ trees and therefore there can be a non-trivial interaction between the different trees. On the other hand, Mondrian forest regression uses $M$ independent Mondrian samples to approximate the \emph{equivalent} kernel $k^{\text{MF}}(\mathsf{x}_{*}, \mathsf{x}_n, \mathbf{X})$ (the smoothing matrix). The prediction $\hat{y}$ is by definition a linear function of the equivalent kernel:
\begin{equation*}
\hat{y}
= \sum_{n = 1}^N k^{\text{MF}}(\mathsf{x}_{*}, \mathsf{x}_n, \mathbf{X}) y_n
\end{equation*}
Hence for Mondrian forest regression, the prediction necessarily decomposes with the $M$ Mondrian samples (there is no interaction between the different trees).

\chapter{Density estimation}

\begin{definition} The \textbf{Beta distribution} with shape parameters $a, b > 0$ has probability density function
\begin{equation*}
p(x | a, b) = \frac{\Gamma(a + b)}{\Gamma(a) \Gamma(b)} x^{a - 1} (1 - x)^{b - 1}
\hspace{3em} 0 \leq x \leq 1
\end{equation*}
where $\Gamma(t) = \int_0^{\infty} z^{t - 1} e^{-z} \d z$ is the gamma function (a shifted generalization of the factorial function).
\end{definition}

\begin{proposition} The expectation of the $\text{Beta}(a, b)$ distribution is $\frac{a}{a + b}$.
\begin{proof} The expectation can be calculated from definition:
\begin{equation*}
\IE[ \Beta(a, b) ]
= \int_0^1 x \frac{\Gamma(a + b)}{\Gamma(a) \Gamma(b)} x^{a-1} (1 - x)^{b - 1}
= \frac{a}{a + b} \int_0^1 \frac{\Gamma(a + 1 + b)}{\Gamma(a + 1) \Gamma(b)} x^a (1 - x)^{b - 1}
= \frac{a}{a + b}
\end{equation*}
by recognizing the integral over the $\Beta(a + 1, b)$ distribution which must evaluate to $1$.
\end{proof}
\end{proposition}

Density estimation differs from regression and classification in that it is an unsupervised problem, i.e., no labels are observed in training data.

\begin{definition} \textbf{Density estimation} is the problem of learning a probability density $p$ from a set of \textbf{training} samples $\mD = \{ \mathsf{x}_1, \hdots, \mathsf{x}_N \} \subseteq \IR^D$ generated from $p$. Given a new \textbf{test} point $\mathsf{x}_{*}$, the learned density $\hat{p}$ estimates $\hat{p}(\mathsf{x}_{*})$ for the true density at point $\mathsf{x}_{*}$.
\end{definition}

A Mondrian random forest model for density estimation needs to be able to predict density values in its leaves, noting that a probability density needs to integrate to $1$. Following the approach taken for classification, we use a hierarchical Bayesian model. In each sample of the Mondrian process, we associate each node (not just the leaves) with an unknown probability mass, with the constraint that each non-leaf node's mass must equal the sum of masses associated with its two children. The prior distribution over these masses is as follows:
\begin{itemize}
\item[•] With probability $1$, the root (of depth $0$) is associated with a probability mass of $1$.
\item[•] Say $n$ is a non-leaf node of depth $d$, with associated probability mass $P_n$. Let $c_1$, $c_2$ be the children of $n$ and let $V_1$, $V_2$ be the volumes of the boxes in $\IR^D$ associated with $c_1$ and $c_2$. Under the prior, the probability masses $P_{n1}$, $P_{n2}$ associated with the children are then assumed to be generated as follows:
\begin{IEEEeqnarray*}{rCll+x*}
\eps_n & \sim &
  \Beta\left( \gamma \frac{V_1}{V_1 + V_2} (d + 1)^2, \gamma \frac{V_2}{V_1 + V_2} (d + 1)^2 \right)
\\
P_{n1} & = &
  \eps_n P_n
\\
P_{n2} & = &
  (1 - \eps_n) P_n
\end{IEEEeqnarray*}
\end{itemize}
Here $\gamma > 0$ is a hyperparameter of the prior to be chosen. Note that the two parameters of the Beta distribution are proportional to the volumes $V_1$, $V_2$, so that the child of larger volume is more likely to get a larger share of $n$'s associated probability mass $P_n$. Also, the sum of the two Beta parameters is designed to be $\gamma (d + 1)^2$, in line with the Pólya tree prior distribution presented in \cite{RSSB:RSSB190} where this choice leads to an a.s. absolutely continuous function in the limit $d \to \infty$.

As training data is added to the $M$ Mondrian samples, the posterior distributions of $\eps_n$ can be computed analytically thanks to Beta-Binomial conjugacy. Under the posterior distribution, a probability mass $P_n$ of node $n$ is a product of independent Beta distributions, which doesn't have a simple analytic form, but its expectation can be easily computed. If $n$ is in fact a leaf, we assume the mass $P_n$ to be uniformly distributed in the box associated with $n$. Note that this again requires knowledge of this volume, if a density is to be predicted at point lying in $n$.

In regression and classification, the Mondrian samples are used only to partition the datapoints. In our density estimation model we also require knowing the volumes of the partition cells generated by the Mondrians. This proves to be a limiting factor for the computational performance of the model, since self-consistency cannot be invoked as easily as in regression and classification: we need to know how far individual partition cells extend, which might be far from any training data. We have explored two possible solutions to this problem:
\begin{itemize}
\item[(1)] Identify a bounded box around the training data and assume that all probability mass lies in that box. Within this box we instantiate the Mondrian samples completely, i.e. even in regions containing no data points, since some of these cuts may still affect the volume of a box that is non-empty.
\item[(2)] Rather than having an exact value for the volume of each box, we may compute its probability distribution under the randomness stemming from not instantiating the Mondrian samples in regions with no training data. Unfortunately, this distribution is somewhat more complicated than hoped: we believe it to be the product of independent, truncated piecewise exponential distributions.
\end{itemize}

\chapter{Cholesky decomposition}

In the chapters on approximating the Laplace kernel we've proposed making efficient updates to the matrix $\mathbf{C}^{-1}$, the inverse of the regularized covariance matrix $\mathbf{C} = \mathbf{Z}^T \mathbf{Z} + \delta^2 \mathbf{I}_C$, to efficiently explore the space of possible Laplace kernel lifetimes. Instead of working with this inverse directly, it is often suggested for numerical stability reasons to work with the Cholesky decomposition \cite{EPFL-REPORT-161468}. In this section we sketch how this can be achieved in our setting.

\begin{definition} Given a positive definite matrix $\mathbf{C}$ (i.e., $\mathbf{v}^T \mathbf{C} \mathbf{v} > 0$ for all $\mathbf{v} \not= \mathbf{0}$), its \textbf{Cholesky decomposition} is a factorization of the form $\mathbf{C} = \mathbf{L} \mathbf{L}^T$, where $\mathbf{L}$ is a lower-triangular matrix.
\end{definition}

It is a standard linear algebra result that the Cholesky decomposition exists and is unique for positive definite matrices.

Suppose first that the Cholesky decomposition $\mathbf{C} = \mathbf{L} \mathbf{L}^T$ of the regularized covariance matrix $\mathbf{C} = \mathbf{Z}^T \mathbf{Z} + \delta^2 \mathbf{I}_C$ is known. The normal equation of ridge regression then reads $\mathbf{L} \mathbf{L}^T \bs{\theta}^{\text{MAP}} = \mathbf{Z}^T \mathbf{y}$, which can be solved in time $\mO(C^2)$ by performing two back-substitution passes. Indeed, we may define $\mathbf{\gamma} := \mathbf{L}^T \bs{\theta}^{\text{MAP}}$, solve the equation $\mathbf{L} \mathbf{\gamma} = \mathbf{Z}^T \mathbf{y}$ using back-substitution ($\mathbf{L}$ is lower-triangular) and then solve the defining relation of $\mathbf{\gamma}$ for $\bs{\theta}^{\text{MAP}}$ using another back-substitution ($\mathbf{L}^T$ is upper-triangular).

Thus if we can maintain the Cholesky decomposition of the regularized covariance matrix, we will be able to efficiently compute the optimal ridge regression solution and hence the error on the validation set. Now it remains to find a way of efficiently updating the Cholesky decomposition when the regularized covariance matrix changes due to a change in lifetimes of the Laplace kernel being approximated.

First consider how the Cholesky decomposition can be efficiently updated when the matrix $\mathbf{C}$ is extended, which happens when new features are created by adding a new cut into the Mondrian grid. As the ordering of features doesn't matter, we may assume that new features are appended as last columns to the feature matrix $\mathbf{Z}$. This means that we only need to consider extending $\mathbf{C}$ by a new last row and column. The equation
\begin{equation*}
\begin{bmatrix}
\mathbf{C} & \mathbf{a} \\
\mathbf{a}^T & b
\end{bmatrix}
= \begin{bmatrix}
\mathbf{E} & \mathbf{0} \\
\mathbf{c}^T & d
\end{bmatrix}
\begin{bmatrix}
\mathbf{E}^T & \mathbf{c} \\
\mathbf{0} & d
\end{bmatrix}
\end{equation*}
is equivalent to the system $\mathbf{C} = \mathbf{E} \mathbf{E}^T$, $\mathbf{a} = \mathbf{E} \mathbf{c}$ and $\mathbf{c}^T \mathbf{c} + d^2 = b$. Provided that we've maintained the Cholesky decomposition of $\mathbf{C}$, the first equation is solved by taking $\mathbf{E} = \mathbf{L}$. Then as $\mathbf{E}$ is lower-triangular, we can solve for $\mathbf{c}$ in the second equation using back-substitution, and finally obtain $d = \sqrt{b - \mathbf{c}^T \mathbf{c}}$ from the third equation, all in time $\mO(C^2)$.

It is significantly more challenging to update the Cholesky decomposition when a feature is removed, which happens when a feature that has been split into two by a new cut needs to be removed. The reason for the difficulty is that the removed feature need not correspond to the last row and column of $\mathbf{C}$, so updating the Cholesky decomposition is not simply a matter of deleting the corresponding row and column from $\mathbf{L}$. Instead, a two step procedure can be carried out \cite{EPFL-REPORT-161468}:
\begin{itemize}
\item[(1)] Rotate the rows and columns of $\mathbf{C}$ so that those intended for deletion end up as the last row and column. Update the Cholesky decomposition correspondingly, e.g. using the SCHEX \cite{doi:10.1137/1.9781611971811.ch10} subroutine from the LINPACK Fortran linear algebra package.
\item[(2)] Delete the last row and column of $\mathbf{C}$, to which the corresponding update of $\mathbf{L}$ is simply to remove its last row and column, as can be easily checked.
\end{itemize}
Both steps run in $\mO(C^2)$ time. This shows that the maintenance of the Cholesky decomposition $\mathbf{C} = \mathbf{L} \mathbf{L}^T$ can be performed with the same efficiency as the maintenance of the inverse $\mathbf{C}^{-1}$, so indeed we can work with the Cholesky decomposition if so desired.

On the datasets considered we haven't run into numerical stability issues while working with the matrix inverse $\mathbf{C}^{-1}$ directly, so the approach using the Cholesky decomposition has not been implemented. Hence we also omit a technical description of the SCHEX subroutine here.


\renewcommand{\bibname}{References}
\printbibliography
\addcontentsline{toc}{chapter}{References}

\end{document}